\documentclass[a4paper]{cas-sc} 

\usepackage[numbers]{natbib}
\usepackage[utf8]{inputenc}
\usepackage[T1]{fontenc}
\usepackage{hyperref}
\usepackage{url}
\usepackage{booktabs}
\usepackage{adjustbox}
\usepackage{amsfonts}
\usepackage{nicefrac}
\usepackage[expansion=false]{microtype}
\usepackage{xcolor}
\usepackage{graphicx}
\usepackage{tikz}
\usepackage{amsmath}
\usepackage[capitalise]{cleveref}
\usepackage{float}
\usepackage{subcaption}
\usepackage{siunitx}
\usepackage{mdframed}
\usepackage{multirow}
\usepackage[section]{placeins}
\usepackage{etoolbox}
\usetikzlibrary{arrows.meta,calc,positioning}

\newif\ifdisablecolors
\disablecolorstrue   

\usepackage{soul}

\newcommand{\need}[1]{%
  {\sethlcolor{\needbg}\hl{#1}}%
}

\newcommand{\milestone}[1]{%
  {\sethlcolor{\milestonebg}\hl{#1}}%
}

\ifdisablecolors
  \renewcommand{\need}[1]{#1}
  \renewcommand{\milestone}[1]{#1}
\fi

\usepackage{amsmath,amsfonts,bm}

\def\eqref#1{equation~\ref{#1}}

\def\1{\bm{1}}

\def\vx{{\bm{x}}}

\def\vz{{\bm{z}}}
\def\vW{{\bm{W}}}
\def\vX{{\bm{X}}}

\DeclareMathAlphabet{\mathsfit}{\encodingdefault}{\sfdefault}{m}{sl}
\SetMathAlphabet{\mathsfit}{bold}{\encodingdefault}{\sfdefault}{bx}{n}

\def\gX{{\mathcal{X}}}
\def\gY{{\mathcal{Y}}}
\def\gZ{{\mathcal{Z}}}
\def\gZ{{\mathcal{Z}}}

\ExplSyntaxOn
\cs_gset:Npn \__first_footerline:
  {
    \group_begin:
    \small
    \sffamily
    \ifnum\theblind>0\relax
    \else
      \__short_authors:
    \fi
    \group_end:
  }
\ExplSyntaxOff

\usepackage[table]{xcolor}
\usepackage[most]{tcolorbox}
\newtcolorbox{conclusionbox}{
  colback=blue!5,        %
  colframe=blue!75!black, %
  coltitle=black,        %
  fonttitle=\bfseries,   %
  boxrule=1pt,           %
  arc=1mm,               %
  left=2mm,              %
  right=2mm,             %
  top=1mm,               %
  bottom=1mm,            %
}

\begin{document}

\shorttitle{Towards Unified and Data-Efficient PHM}
\shortauthors{R. Theiler, L. Telyatnikov et~al.}

\title[mode=title]{Towards Unified and Data-Efficient Prognostics and Health Management with Tabular Foundation Models}

\author[1]{Raffael Theiler}[type=editor,
                        auid=000,bioid=1,
                        role=Researcher,
                        orcid=0000-0000-0000-0000]
\cormark[1] 
\ead{raffael.theiler@epfl.ch, lev.telyatnikov@epfl.ch}

\author[1]{Lev Telyatnikov} \cormark[1]
\author[1]{Leandro Von Krannichfeldt}
\author[1]{Olga Fink}

\affiliation[1]{organization={IMOS Lab, EPFL},
                city={Lausanne},
                country={Switzerland}}

\cortext[cor1]{Equal contribution and Corresponding author}

\begin{abstract}
  Data-driven Prognostics and Health Management (PHM) uses time-varying condition-monitoring data to diagnose system states and estimate remaining useful life in engineered assets. These tasks are central to maintenance planning, but industrial PHM data are often fragmented, partially observed, and poorly labeled, which hinders supervised learning.
  Foundation models offer a route toward reusable predictive systems, yet most time-series foundation models are designed for forecasting and assume long, coherent, regularly sampled sequences.
  To address this gap, we propose a framework for applying Tabular Foundation Models to industrial time series using in-context learning, and we evaluate them on a variety of PHM tasks.
  By converting raw unit-level signals into tabular rows, we show that these models perform well across multiple tasks - including prognostics, and diagnostics — and are highly data efficient. We compare them directly with sequence models, transformer baselines, and gradient-boosted trees under a common evaluation protocol.
  The results indicate that tabular foundation models achieve the best average ranks across prognostic and diagnostic tasks. Our findings further show that PFN-based models are competitive in low-data regimes, that temporal context can be preserved in the tabular representation, and that performance depends on representative context construction under subsampling.
  These results demonstrate that tabular foundation models provide a practical and general interface for heterogeneous PHM problems.
\end{abstract}


\maketitle

\section{Introduction}

Prognostics and Health Management (PHM) has matured into an established framework for supporting maintenance and operational decisions in complex engineering systems. In industrial practice, PHM solutions are typically based on condition monitoring data collected from heterogeneous sensor networks and are used to support fault detection, diagnostics, and, where feasible, remaining useful life (RUL) prediction \citep{zontaPredictiveMaintenanceIndustry2020,leiMachineryHealthPrognostics2020}. These outputs are then integrated into maintenance planning and asset management processes to reduce downtime and maintenance costs while maintaining high system availability.

A cornerstone of PHM is the availability of degradation trajectories and time-to-failure information, as these data capture the temporal evolution of system health and form the basis for many prognostic approaches \citep{leiMachineryHealthPrognostics2020}. Despite their importance, such trajectories are rarely available in operational environments. Failures are relatively rare events, assets are often maintained or replaced before reaching end of life, and historical data may be fragmented or incomplete. As a consequence, many prognostic approaches developed and evaluated in academic settings cannot be readily deployed in practice \cite{finkPhysicsMachineLearning2026II}. Instead, most industrial PHM implementations today primarily focus on fault detection, where data from normal operation are abundant, and the objective is to identify deviations from healthy behavior rather than to predict precise failure times.

Recent advances in machine learning, particularly deep learning, have aimed to reduce reliance on handcrafted features by learning representations directly from data \citep{lecunDeepLearning2015,zhao2019deep,wangDeepLearningSmartManufacturing2020}. While such approaches have demonstrated strong performance in controlled settings, their success often relies on assumptions that are difficult to satisfy in industrial environments. In particular, training deep models from scratch typically requires large amounts of labeled data that cover the full range of operating conditions and fault types. In industrial PHM, however, labeled failure data are scarce, system configurations evolve over time, and degradation trajectories are rarely observed until the end of life \citep{li2024small}. As a result, models trained under one set of conditions often require costly retraining, adaptation, and validation before they can be deployed across fleets of assets.

In current industrial PHM practice, models are often developed and trained for individual components or specific systems, reflecting differences in design, sensing setups, and operating conditions \citep{hagmeyer2021creation,mauthe2025overview,zhang_pdmbench_2025}. However, in real deployments these models must operate across fleets of assets and multi-component systems within the same organization. Assets frequently differ in age, configuration, and operating regimes, and their behavior may evolve over time due to changes in usage patterns, maintenance strategies, or environmental conditions \cite{finkPhysicsMachineLearning2026II}. Continuously retraining and revalidating separate models for each component, asset, or configuration is costly and difficult to scale, particularly in safety-critical applications. This lack of scalability remains one of the key barriers to broader adoption of advanced PHM methods in industry. Learning approaches that can generalize across assets and adapt to new conditions without repeated retraining are therefore of high practical relevance.

Under these constraints, in-context learning provides an alternative to training models from scratch \citep{dong2024survey,liu2025icl4rul}. Rather than updating model parameters, foundation models adapt their behavior at inference time by conditioning on a small set of task-specific examples \citep{zhou2025comprehensive}. This mechanism is particularly well suited to PHM applications, where labeled data are limited, system behavior varies across assets, and rapid adaptation to changing operating conditions is required. By enabling task adaptation without retraining or parameter updates, in-context learning reduces deployment and maintenance overhead while remaining compatible with established PHM workflows.

While time-series foundation models have recently been proposed and extended to in-context learning \cite{ansari2025chronos2}, their applicability in industrial PHM is constrained by both data assumptions and task formulation \citep{miller2024survey,wang2024tssurvey}. These models are typically designed for forecasting tasks, where the objective is to predict future signal values over a temporal horizon \citep{woo2024unified,das2024timesfm,liu2024timer}, and effective adaptation relies on long, coherent temporal sequences with stable dynamics, consistent sensor availability, and well-aligned sampling across channels. In contrast, prognostics problems are typically framed as regression tasks, such as estimating health indicators or remaining useful life from the current system state, often without access to complete or continuous degradation trajectories. Moreover, in industrial PHM settings, condition monitoring data are frequently irregular, affected by missing values, and disrupted by maintenance actions or changes in operating conditions. These factors limit the suitability of time-series foundation models, particularly for prognostics applications.

At the same time, the absence of continuous run-to-failure trajectories motivates learning paradigms that operate on snapshots of system state rather than complete time-to-failure trajectories. In practice, PHM models often rely on tabular representations composed of aggregated sensor features, condition indicators, and contextual variables \citep{sim2020tutorial,zhao2019deep}. Such representations allow information to be pooled across time and assets, are robust to interruptions caused by maintenance actions, and do not require strict temporal alignment between sensors. This is particularly important in industrial environments, where sensor signals often exhibit different sampling rates, heterogeneous temporal dynamics, and varying levels of reliability, and where data streams are frequently affected by missing values due to sensor faults, communication issues, or changes in instrumentation \citep{hagmeyer2021creation,mauthe2025overview,zhang_pdmbench_2025}. Handling these characteristics in sequential models typically requires additional assumptions, resampling, or imputation strategies, whereas tabular representations and models can more naturally accommodate incomplete, irregular, and heterogeneous observations.

Within this setting, tabular foundation models combined with in-context learning provide a strong alternative to conventional end-to-end training paradigms \citep{hollmanntabpfn,hollmann2025accurate,ma2024tabdpt}. By leveraging a single pre-trained model across tasks and assets, these approaches reduce the need for repeated architecture-specific hyperparameter optimization; the remaining tuning is concentrated mainly on tabular-shape choices such as lookback length, feature construction, and context size. Adaptation to evolving operating conditions, changing sensor availability, or new assets can be achieved through contextual examples at inference time, rather than through parameter updates, thereby avoiding costly retraining cycles and simplifying deployment and lifecycle management.

\section{Related Work}

\subsection{Time-series Foundation Models}
Recently, the broader field of machine learning has witnessed the emergence of \emph{Foundation Models} (FMs) as powerful, general-purpose architectures, leading to a profound impact across multiple disciplines in science \citep{zhou2025comprehensive}.
FMs are large neural networks pre-trained on diverse and heterogeneous datasets; the unprecedented scale of this pretraining equips them with emergent general-purpose capabilities, allowing them to learn general, transferable latent representations and priors that can be reused across tasks and datasets. Especially for Natural Language Processing, this often allows FMs to outperform domain-specialized models with minimal task-specific adaptation \citep{nori2023can}.

There has been substantial interest in reproducing this success in the field of time-series analysis, generalist Time-Series Foundation models. Inference for these models may be seen as generally analogous to that of the large language models: continuous-time measurements are sampled, then embedded as discrete high-dimensional tokens, which are then fed to a deep learning model composed of self/cross-attention layers and multilayer perceptrons. 
Time-series Foundation Models (TSFMs) extend Time-series Transformers architecture beyond task-specific design, to enable zero-shot or few-shot generalization across datasets and domains. In general, the research of TSMs can be divided into three categories \citep{miller2024survey}.
One category concerns itself with the repurposing of language models for time series tasks, such as Time-LLM \citep{jin2023timellm}. Hereby, time series sequence is converted into a word structure, divided into patches, and used as input tokens for the language foundation model.
The second category leverages a pre-train on a large variety of data. The different works either train exclusively on real-world data \citep{woo2024unified}, only on synthetic data \citep{dooley2024forecastpfn} or a combination of real measurements and synthetically generated samples \citep{das2024timesfm, ansari2023chronos}.
The third category creates domain-specific TSFMs by training directly on domain-relevant datasets like BioBert for biomedical literature \citep{lee2019biobert} or PowerPM for Power System tasks \citep{tu2024PowerPM}. Their idea is to improve performance for specific domain tasks, albeit incurring a reduced generalization ability.

The majority of TSFM works focus on univariate forecasting by treating each time series independently \citep{feng2024only}, and consequently do not make effective use of covariance dependencies of a multivariate time series input. Some publications try to mitigate this effect by constructing a new time series out of the multivariate series through a flattening operation \citep{woo2024unified} or subsampling \citep{liu2024timer}. Recent works recognize the potential of in-context learning for covariate information integration, and are slowly shifting towards in-context learning for TSFM's \citep{ansari2025chronos2}.

\subsection{Tabular Foundation Models}
Tabular Foundation Models (TFMs) have recently emerged as  transformer-based FMs for two-dimensional data, including tables and relational databases. TFMs are trained to predict cells in tables by leveraging contextual information from adjacent rows and columns, allowing them to fill missing values, extrapolate rows and columns, and detect anomalies. This is achieved by training on vast amounts of tabular data, coupled with a two-dimensional attention mechanism that captures both row-level and column-level dependencies. These capabilities effectively make TFMs universal pattern recognizers for table-like data, and have achieved state-of-the-art performance on multiple tabular prediction tasks.

TFMs differ on how their attention mechanisms are implemented, as well as on the nature of their training data, leading to multiple implementations in the literature. \emph{Tabular Prior-Fitted Network} (TabPFN) \citep{hollmanntabpfn,hollmann2025accurate}, emulates Bayesian inference on tables by applying attention to raw table data, and is trained on massive causally-generated synthetic datasets; 
\emph{TabDPT} combines synthetic and real datasets to improve generalization and one-shot capacity \citep{ma2024tabdpt};
\emph{TabICL} uses statistical row embeddings to compress tables, enabling scalable extensions to hundreds of thousands of samples without fine-tuning \citep{quTabICL2025};
\emph{Carte} embeds tables with graph-like structure to extract semantic connections between feature columns \citep{kimcarte}.

The success of TFMs has led to these models being used for a variety of two-dimensional signals, including multivariate time series. This typically takes place by representing time series as tables and embedding time markings as additional feature columns. \cite{hooTabularFoundationModel2024} demonstrate on small-scale, scalar time series that a simple tabularization and feature-based representation of time series allows TabPFN to match or outperform specialized forecasting models, highlighting the strength of large-scale synthetic pretraining. 
Similarly, \cite{caiExploreTimeSeries} further adapt TabPFN for time-series forecasting by explicitly encoding time-periodic features as sinusoidal table features, showing consistent improvements over baseline TabPFN and other time series forecasting models for small quasi-periodic time-series.

While the use of tabular foundation models for time-series forecasting has recently gained attention, it remains unclear how to fully exploit the time dependencies within and across time series beyond simple periodic indexing or feature lagging, especially under the computational constraints of tabular foundation models.

\subsection{Foundation Models for PHM}
The success of FMs for sequence modeling has led to substantial interest in using them for industrial time series across multiple applications. Despite numerous FM architectures being proposed to enhance state-of-the-art performance in various time-series-related tasks \citep{wang2024tssurvey}, all-encompassing FM for PHM have seen limited adoption in industry, where classical models such as Particle Filters, LSTM and CNN are still the most widely adopted \cite{salinas-camusComprehensiveReviewEvaluation2025}.
Recent studies in the PHM literature have begun to utilize TSFMs. An example is UniFault \citep{eldele2025unifault}, which proposes large-scale pretraining for rotating machinery fault diagnosis, with the goal of improving cross-domain transfer and few-shot adaptation when only limited labeled data are available for a new machine or condition. 
Another example is \citet{yao2026utilizing}, where the streamlining of prognostics and maintenance of wind turbines with generalization capabilities of TSFMs is conceptually outlined. 
The strong performance of TFMs has also led to exploratory applications in PHM. 
\citet{magadanEarlyFaultClassification2023} use TabPFN for early fault classification in rotating machinery, showing that it maintains strong diagnostic accuracy even with very limited training samples. 
\citet{sunFaultDiagnosisSlewing2025} propose a hybrid TimeGAN--TabPFN framework for slewing bearing fault diagnosis using audible sound signals, where generative data augmentation combined with TabPFN significantly improves classification performance under small-sample conditions. 

Although tabular foundation models are slowly gaining traction in the PHM field with domain-specific pretraining and several diagnostics application studies, a systematic benchmarking study with comprehensive evaluations in both prognostics and diagnostics is still missing.

\subsection{Benchmarks in PHM}
In recent years, several efforts have been conducted in terms of systematic framework proposals and benchmarking evaluations in PHM. An early review study was presented by \citet{ramasso_performance_2014}, who analyzes prognostic methods predating modern deep-learning, developed using the C-MAPSS turbofan degradation datasets \cite{saxena2008damage}. The study examines how the C-MAPSS dataset has been used across prior work, clarified differences among the dataset variants, and reconstructed benchmark results from the PHM 2008 Data Challenge. It also highlights inconsistencies in dataset selection, preprocessing, and metric usage, and provides guidelines for more consistent benchmarking. 
More recently, \citet{qiao2025comparative} carried out a comparative benchmarking evaluation of deep-learning models for bearing fault diagnosis and fault prognosis using the XJTU-SY \citet{yaguo2019xjtu} and CWRU \citep{cwru_bearing_data_center} datasets. The study evaluated CNN, LSTM, ResNet, Transformer, and hybrid architectures, identifying ResCNN-LSTM as a strong model for both diagnosis and prognosis. However, its scope is limited to conventional deep-learning architectures and does not consider TSFMs or TFMs.
PDMBench \citep{zhang_pdmbench_2025} proposes a standardized platform for predictive maintenance research that integrates 14 public datasets covering bearings, motors, gearboxes, and multi-component systems, together with 22 baseline models ranging from traditional machine-learning methods to modern Transformer-based architectures. The framework addresses practical challenges such as irregular sampling rates, imbalanced fault distributions, heterogeneous sensor modalities, and deployment constraints, while focusing on fault classification and remaining useful life prediction. Its experimental analysis shows that Transformer architectures perform strongly on structured bearing datasets but are less robust when applied to noisy motor signals, whereas lightweight models provide attractive accuracy--efficiency trade-offs for deployment. Although PDMBench provides a systematic benchmarking framework, it does not standardize evaluation protocol as well as task semantics and does not evaluate foundation-model approaches.
Beyond dataset-level standardization, recent work has also addressed the reproducible implementation of conventional PHM methods. \citet{telyatnikov2026picid} introduce PICID as a modular PHM evaluation framework for specifying datasets, preprocessing, target construction, model interfaces, and metrics under a shared executable protocol. This framework has been used to implement and evaluate conventional PHM papers as reproducible benchmark components \cite{theiler2026paperbenchmark}. 

Several conceptual as well as experimental benchmarking publications have been carried out with a strong focus on standard deep-learning models and Transformer architectures. Therefore, a rigorous and reproducible investigation of tabular foundation models for prognostics and diagnostics has not yet been addressed.

In this work, we use PICID as the evaluation infrastructure for assessing tabular foundation models under the same task, dataset, preprocessing, and metric interfaces as their conventional PHM competitors. In particular, PICID's fit--predict model interface enables tabular foundation models to be benchmarked against conventional baselines through the same executable protocol, making conventional PHM methods available as standardized competitors rather than as isolated, paper-specific implementations.

\section{Methodology}

In this work, we formalize a unified methodology for transforming raw PHM time-series into representations compatible with both sequence-based models and tabular foundation models.
The central objective is to ensure that all models are evaluated on information-equivalent inputs, thereby isolating the impact of the modeling architecture from differences in preprocessing.

A key challenge arises from the mismatch between the inherently sequential nature of PHM data and the requirements of tabular foundation models, which operate on collections of independent samples and, in the case of in-context learning, rely on well-defined context sets. To bridge this gap, we define a data processing pipeline that converts continuous time-series into supervised samples that can be interpreted consistently across model families.
To this end, we introduce a complete data processing pipeline that spans feature extraction, target alignment, sequence construction, and tabularization. This pipeline allows us to compare fundamentally different model families under controlled and consistent conditions.

\begin{figure}[pos=t]
\centering
\includegraphics[width=0.95\textwidth]{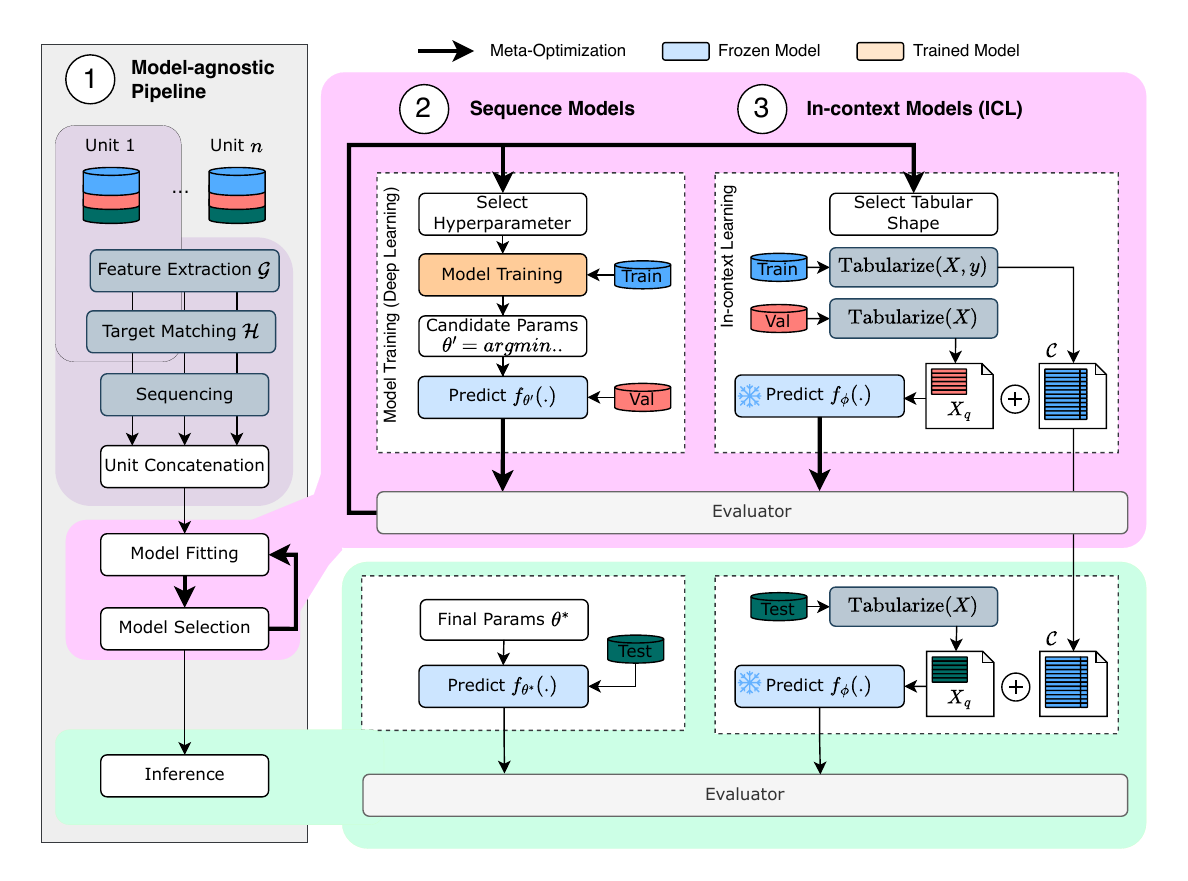}
\caption{Overview of the unified evaluation pipeline. Unit-level PHM signals are transformed into aligned feature--target windows and evaluated either by trained sequence models or by in-context tabular foundation models after tabularization. Validation data are used for model selection or tabular-shape selection before final test evaluation.}
\label{fig:visual_abstract_framework}
\end{figure}

Figure~\ref{fig:visual_abstract_framework} provides a high-level overview of the proposed framework. This section is structured as a step-by-step pipeline. We first define how raw signals are transformed into aligned feature–target sequences. We then describe how supervised samples are constructed through sequence slicing, and finally how these samples are converted into tabular representations via the tabularization operator.

\subsection{Data Pre-Processing Pipeline}

We formalize tabularization as a multi-stage transformation pipeline that maps raw time-series data into supervised learning representations. Specifically, the pipeline consists of three stages: (i) transforming raw signals into a compact feature representation ($\mathcal{G}$),  (ii) slicing the transformed features into temporal sequences ($\mathcal{S}$), and (iii) mapping these sequences into tabular representations via the tabularization operator ($\mathcal{T}$). A key aspect of this formulation is that target alignment is defined explicitly.  This allows us to accommodate feature transformations that modify the temporal grid (e.g., time-frequency representations), while remaining agnostic to the specific alignment convention. 

Importantly, this explicit formulation ensures that the resulting samples are:

\begin{itemize}
    \item well-defined as supervised learning instances, and
    \item directly usable as context elements for tabular foundation models.
\end{itemize}

Table~\ref{tab:notation} summarizes the notation used throughout this pipeline.

\begin{table}[h!]
    \centering
    \caption{Summary of Notation}
    \label{tab:notation}
    \renewcommand{\arraystretch}{1.2}
    \begin{tabular}{@{}ll@{}}
        \toprule
        \textbf{Symbol} & \textbf{Description} \\
        \midrule
        \multicolumn{2}{l}{\textit{Raw Data Space}} \\
        $T$ & Total duration (number of timestamps) of the raw unit lifecycle \\
        $t$ & Discrete raw time index, $t \in \{1, \dots, T\}$ \\
        $\vx(t)$ & Raw sensor vector at time $t$, $\vx(t) \in \mathbb{R}^M$ \\
        $y(t)$ & Raw target value at time $t$ \\
        \midrule
        \multicolumn{2}{l}{\textit{ Data Transformation ($\mathcal{G}, \mathcal{H}$)}} \\
        $T'$ & Total number of steps in the transformed time-series \\
        $w$ & History window size (receptive field) for feature extraction \\
        $s_{\mathcal{G}}$ & Signal processing stride (step size for sliding the transformation window) \\
        $\Psi, \Phi$ & Collections of fitted parameters for feature and target transformations \\
        $\mathcal{G}(\cdot; \Psi)$ & Feature transformation operator for sensor signals \\
        $\vz(j)$ & Transformed feature vector at feature-step $j$, $\vz(j) \in \mathbb{R}^F$ \\
        $z_y(j)$ & Transformed target value at feature-step $j$ \\
        $a(j)$ & Raw-time support associated with feature-step $j$ (timestamp or interval) \\
        $\mathcal{A}(\cdot)$ & Alignment/aggregation operator used within the target pipeline \\
        $\mathcal{H}(\cdot; \Phi)$ & Target transformation operator (e.g., RUL normalization or scaling) \\
        $\widetilde{\mathcal{H}}(\cdot;\Phi)$ & Target pipeline that produces one label per transformed index using $a(j)$ \\
        \midrule
        \multicolumn{2}{l}{\textit{ Dataset Slicing ($\mathcal{S}$)}} \\
        $L_{\mathrm{seq}}$ & History length (number of transformed steps per input window) \\
        $\Delta$ & Stride between consecutive window starts in transformed time \\
        $\rho$ & Warm-start depth (left-padding allowance), $\rho=0$ strict windowing \\
        $\delta$ & Supervision offset (steps after window end), $\delta=0$ end-of-window \\
        $L_{\mathrm{pred}}$ & Supervision segment length (default $L_{\mathrm{pred}}=1$ for PHM window labels) \\
        $\vW_m$ & Temporal feature window \\
        $N_{\mathrm{slices}}$ & Total number of windows extracted from the unit \\
        $\mathcal{K}$ & Admissible set of window start indices in transformed time \\
        \midrule
        \multicolumn{2}{l}{\textit{Tabularization ($\mathcal{T}$)}} \\
        $\vX_m$ & Final tabular input vector for the model \\
        $\mathcal{D}_{\mathrm{seq}}$ & Dataset in sequence format (for time-series models) \\
        $\mathcal{D}_{\mathrm{tab}}$ & Dataset in tabular format (for tabular models) \\
        $D_{\mathrm{tab}}$ & Tabular dimensionality, $D_{\mathrm{tab}}=L_{\mathrm{seq}}F$ \\
        \midrule
        \multicolumn{2}{l}{\textit{Modeling \& Evaluation}} \\
        $f_{\theta}(\cdot)$ & Deep Learning Model (Sequence-based) \\
        $f_{\phi}(\cdot)$ & Foundation Model (Tabular/In-Context) \\
        $\mathcal{C}$ & Context set for In-Context Learning \\
        $\mathcal{L}$ & Loss function \\
        \bottomrule
    \end{tabular}
\end{table}

\subsubsection{Problem Formulation: Single Unit Perspective}

We begin by formalizing the data representation for a single functional unit monitored continuously over its full operational lifecycle. This setting serves as the fundamental building block for the subsequent pipeline,  which transforms raw time-series into supervised learning samples. The raw data consists of a multivariate time-series of sensor measurements $\gX$ and a corresponding  target signal $\gY$:
\begin{equation}
    \gX = \{\vx(t)\}_{t=1}^{T}, \quad \gY = \{y(t)\}_{t=1}^{T}
\end{equation}
where $T$ denotes the total number of recorded timestamps. At each  time step $t$, the system state is represented  by a vector of sensor observations $\vx(t) \in \mathbb{R}^M$, defined as :
\begin{equation}
    \vx(t) = [x_1(t), x_2(t), \dots, x_M(t)]^\top
\end{equation}
where $x_m(t)$ represents the scalar measurement of the $m$-th sensor at time $t$. These channels may correspond to endogenous system variables or exogenous covariates.
The target signal  $y(t)$ encodes  the  system health state  at time $t$. Depending on the PHM task, this may represent a continuous quantity (e.g., remaining useful life for prognostics) or a discrete label (e.g., fault type for diagnostics).
This formulation defines the raw data space on which all subsequent transformations operate. In particular, it establishes the relationship between sensor observations and the target signal at the original temporal resolution, which will serve as the reference for all downstream processing steps.

\begin{conclusionbox}
\textbf{Example.}
Consider a lithium-ion battery cell from the UNIBO21 dataset \citep{10.1145/3462203.3475878} monitored over its operational lifetime. During each discharge cycle, three signals are recorded at regular intervals: voltage, discharge current, and cell temperature, so $M=3$ in this example. The target $y(t)$ is the remaining cumulative discharge throughput, or ah-RUL, which decreases from the first recorded cycle toward zero at end-of-life. This raw pair $(\gX,\gY)$ serves as input to the pipeline.
\end{conclusionbox}

The goal of this stage is to transform the raw time-series into a feature representation that can be used to construct supervised learning samples while ensuring consistency between inputs and targets. In PHM, preprocessing often constitutes a substantial part of the methodological pipeline. It includes operations such as data cleaning, normalization, resampling, windowing, detrending, denoising, and signal-processing transforms. Although these stages typically receive less emphasis than the predictive model, they are at least equally important for final performance and reproducibility. A central challenge is that feature extraction procedures may modify the temporal structure of the data, while the target signal remains defined on the original timeline. As a result, we must explicitly define how features and targets are brought into correspondence. 

\subsubsection{Feature  and Target Transformation Operators}

The first stage pre-processes the raw time-series into a feature representation that can be used to construct supervised learning samples while ensuring consistency between inputs and targets. In PHM, preprocessing is a critical component of the pipeline, often encompassing data cleaning, normalization, resampling, windowing, detrending, denoising, and signal-processing transformations. Despite receiving less attention than the predictive model, these steps are equally important for performance and reproducibility.
To address feature extraction and target processing, we decompose the transformation into two operators:

\begin{itemize}
    \item a feature transformation $\mathcal{G}$
   \item  a target transformation pipeline $\mathcal{H}$ 
\end{itemize}

Target alignment is treated explicitly as part of the pipeline rather than being implicitly induced by the feature representation. This design decouples feature extraction from supervision design, thereby supporting arbitrary feature transformations, including those that alter temporal resolution, while preserving a consistent definition of the target under different alignment conventions.

We define the feature transformation $\mathcal{G}$ as a composition of $P$ processing functions:
\begin{equation}
    \mathcal{G}(\cdot; \Psi) = g_P(\cdot; \psi_P) \circ \dots \circ g_1(\cdot; \psi_1)
\end{equation}
where $\Psi$ denotes the set of fitted parameters (e.g., normalization statistics). Applying the feature transformation $\mathcal{G}$ to the raw sensor series yields a transformed feature series:
\begin{equation}
    \gZ = \{\vz(j)\}_{j=1}^{T'} = \mathcal{G}(\gX;\Psi), \quad \vz(j)\in\mathbb{R}^{F}\ (j=1,\dots,T').
\end{equation}
Similarly, we define the target transformation operator $\mathcal{H}$ as a composition of $Q$ distinct processing functions:
\begin{equation}
    \mathcal{H}(\cdot; \Phi) = h_Q(\cdot; \phi_Q) \circ \dots \circ h_1(\cdot; \phi_1)
\end{equation}
where $\Phi$ denotes the set of fitted parameters used for target preprocessing, estimated exclusively from the training partition (e.g., selection, scaling or clipping parameters). Importantly, $\mathcal{H}$ operates on the target signal itself and does not perform temporal alignment. Its role is to transform the target values into a suitable representation for learning.

We have now introduced the feature and target transformation operators. However, many commonly used feature transformations, such as time–frequency representations (e.g., STFT or wavelets), operate on temporal windows and therefore produce one feature vector per window rather than per original timestamp. This leads to a mismatch between the feature timeline and the raw target timeline. Without a precise alignment strategy, the definition of supervision becomes ambiguous. In addition, it is often necessary to pre-process the target signal before it is used for supervision. For example, in prognostics tasks, the raw target (e.g., remaining useful life) may be clipped, normalized, or rescaled to improve numerical stability and learning behavior. Together, these considerations necessitate an explicit alignment mechanism.

\paragraph{Temporal alignment}
We explicitly define how features and targets are brought into correspondence. To this end, each transformed index \(j\) is associated with a raw-time support \(a(j) \subseteq \{1,\dots,T\}\), which specifies the portion of the original time axis from which the feature \(\vz(j)\) is derived.

The support \(a(j)\) may take one of two forms: it can be a single timestamp \(a(j)\in\{1,\dots,T\}\) (e.g., the end or center of a window), or an interval \(a(j)=[\underline{t}_j,\overline{t}_j]\) (e.g., the window used to compute the feature).

To map this support to a supervision signal, we introduce an alignment operator \(\mathcal{A}\) that extracts a representative target value from the raw target trajectory \(\gY\) over the support \(a(j)\). This yields an aligned target value at index \(j\):
\begin{equation}
z_y(j) = \mathcal{A}(\gY, a(j)).
\end{equation}

Depending on the form of \(a(j)\), \(\mathcal{A}\) may correspond to pointwise sampling (for timestamp supports) or aggregation over a window (e.g., last-value, mean, or max for interval supports). This alignment step is used solely to define supervision and is not provided as input to the model.

\paragraph{Feature-to-raw-time mapping via temporal alignment and temporal indexing.}
When $\mathcal{G}$ modifies the temporal resolution, each transformed index $j$ must be related back to the original time axis. Using the temporal alignment defined above, we associate each feature vector $\vz(j)$ with its corresponding raw-time support $a(j) \subseteq \{1,\dots,T\}$, which specifies the portion of the original time-series from which the feature was derived.

The resulting feature dimension $F$, as seen by the model after applying $\mathcal{G}$, is determined by the final stage operator $g_P$ (e.g., $F = M \times \text{features per channel}$). Each stage $g_p$ operate either pointwise in time (e.g., scaling), or over temporal windows (e.g.,time--frequency transformations such as STFT or wavelets). 
Consequently, the transformation stage may alter both the feature dimension and the temporal resolution.

In many practical implementations, the index set $\{1,\dots,T'\}$ of the transformed sequence $\gZ$ is induced by one or more windowed stages within the pipeline (e.g., STFT or wavelets), which operate with a history length $w$ and stride $s_{\mathcal{G}}$, thereby producing one feature vector per window, yielding
\begin{equation}
    T' = \left\lfloor \frac{T - w}{s_{\mathcal{G}}} \right\rfloor + 1.
\end{equation}
For example, if the windowed stage uses window endpoints $\overline{t}_j = w + (j-1)s_{\mathcal{G}}$, then the raw-time support can be written as $a(j)=[\overline{t}_j-w+1,\overline{t}_j]$ (interval support), or $a(j)=\overline{t}_j$ under an end-of-window convention (timestamp support). Pointwise stages in $\mathcal{G}$ (e.g., scaling) do not affect $T'$ and are applied on whichever grid is current.

\paragraph{Target alignment via raw-time mapping}
Akin to the feature-to-raw-time mapping defined above, target alignment is treated by associating each transformed feature index $j$ with a raw-time support $a(j)$. The target at index $j$ is then constructed from the raw target trajectory $\gY$ using this support, ensuring that each transformed feature is paired with a well-defined supervision signal.
This design decouples feature extraction from supervision design, thereby supporting arbitrary feature transformations, including those that alter temporal resolution, while preserving a consistent definition of the target under different alignment conventions.

Because $\mathcal{G}$ may change the number and meaning of time steps, the target pipeline must produce one label per transformed index $j$. We therefore allow the target pipeline to depend on the raw target series and the associated support $a(j)$. This explicit alignment is particularly critical for prognostics tasks, where targets such as Remaining Useful Life depend on precise temporal positioning (e.g., assigning the RUL at the end of a window) and scaling, whereas in diagnostics tasks labels are often constant or slowly varying over intervals and are therefore less sensitive to the exact alignment choice:

\begin{equation}
    z_y(j) = \widetilde{\mathcal{H}}\big(\gY, a(j); \Phi\big), \quad j=1,\dots,T'.
\end{equation}

Here, $\widetilde{\mathcal{H}}$ is a composition of $Q$ stages (e.g., clipping, scaling, calibration) and may include the previously defined alignment/aggregation operator $\mathcal{A}$ at an arbitrary position in the composition. A common special case is
\begin{equation}
    z_y(j) = \mathcal{A}\Big(\mathcal{H}\big(\gY;\,\Phi\big),a(j)\Big),
\end{equation}
but other transformation sequence orderings (e.g., transforming $y(t)$ pointwise before aggregation over $a(j)$) are equally admissible and task-dependent. Concretely, $\mathcal{A}$ may implement pointwise sampling $\mathcal{A}(\gY;a(j))=y(a(j))$ when $a(j)$ is a timestamp, or window aggregation (e.g., mean/last/max/majority) over $\{y(t)\}_{t\in a(j)}$ when $a(j)$ is an interval. In our experiments, unless otherwise stated, we use an end-of-window convention for window-derived features (i.e., $a(j)=\overline{t}_j$).

Target alignment yields the aligned target series $\gY'=\{z_y(j)\}_{j=1}^{T'}$ paired with $\gZ$. After this stage, the target alignment remains fixed and is no longer modified.

\textbf{Leakage Policy.} All fitted parameters $\Psi$ and $\Phi$ (and any statistics used by $\mathcal{A}$, e.g., thresholds or calibration) are estimated exclusively using the training partition. Once estimated, they are frozen and applied unchanged to validation and test partitions.

\begin{conclusionbox}
\textbf{Example.}
The feature pipeline $\mathcal{G}$ is a two-stage composition. The first stage $g_1$ applies min-max normalization channel-wise using training-partition statistics and preserves the current temporal grid. The second stage $g_2$ computes $K$ time-domain statistics---mean, maximum, RMS, standard deviation, and others---over a sliding window of width $w$ raw samples, expanding the $M$ raw channels into $F=MK$ summary features at each transformed index $j$. The support of $\vz(j)$ is therefore the interval $a(j)=[\overline{t}_j-w+1,\overline{t}_j]$.
For target alignment, we adopt an end-of-window convention, so $a(j)=\overline{t}_j$ for supervision and $z_y(j)=\widetilde{\mathcal{H}}(\gY,\overline{t}_j;\Phi)$. In this example, the target transformation inside $\widetilde{\mathcal{H}}$ scales ah-RUL by dividing by the maximum value observed in the training partition, with this scaling constant included in $\Phi$. Because ah-RUL has a natural lower bound of zero at end-of-life but no fixed upper bound, a cell near end-of-life yields $z_y(j)\approx 0$, while a fresh cell yields $z_y(j)>0$ and may yield $z_y(j)>1$ for test cells with longer lifetimes than any training cell.
\end{conclusionbox}

\subsubsection{Sequence Slicing ($\mathcal{S}$)}

Following the pre-processing stage, the goal of this stage is to construct supervised learning samples for predictive modeling by forming the final input–target pairs from the transformed feature sequence. Rather than predicting from a single time step, each input consists of a finite history window of length $L_{\mathrm{seq}}$ (i.e., how much past context is provided), which is paired with a corresponding target value derived from the aligned target sequence. 
The feature window is created by the slicing operator $\mathcal{S}$ and is applied to the transformed feature sequence $\gZ=\{\vz(j)\}_{j=1}^{T'}$.
Note that this bound on the window length is imposed on the data-level by the input representation and may differ from the effective lookback utilized at the model level due to architectural or optimization constraints.
To formalize the construction of $\mathcal{S}$ , we introduce a set of additional parameters besides $L_{\mathrm{seq}}$ that control how windows are extracted:

\begin{itemize}
    \item  $\Delta\in\mathbb{N}$: the stride between consecutive windows (controls overlap and dataset size),
    \item $\rho\in\mathbb{Z}_{\ge 0}$ (left-padding allowance, with $\rho=0$ strict windowing): an optional warm-start depth that allows windows to start before the first valid index via padding, 
    \item $\delta\in\mathbb{Z}_{\ge 0}$: a supervision offset that determines how far into the future (relative to the window end) the target is read,
    \item $L_{\mathrm{pred}}\in\mathbb{N}$: the length of the supervision segment, where $L_{\mathrm{pred}}=1$ corresponds to a single window-level label. 
\end{itemize}

In the PHM tasks considered here, we use window-level prediction with $L_{\mathrm{pred}}=1$, i.e., a single target value per window and typically $\delta=0$, meaning the target corresponds to the end of the window. However, the more general formulation allows for multi-step prediction or auxiliary objectives without modifying the preprocessing pipeline.

\paragraph{Valid start indices and dataset size.}
To determine the final dataset length, we now characterize the set of valid start indices for window extraction on the transformed timeline, given $(L_{\mathrm{seq}},\Delta,\rho,\delta,L_{\mathrm{pred}})$ and transformed length $T'$.

\begin{equation}
    L_{\mathrm{req}} \triangleq L_{\mathrm{seq}} + \delta + L_{\mathrm{pred}} - 1.
\end{equation}
\begin{equation}
    k_m \triangleq 1-\rho + (m-1)\Delta,\qquad m=1,2,\dots
\end{equation}
\begin{equation}
    k_m + L_{\mathrm{req}} - 1 \le T'.
\end{equation}
\begin{equation}
    N_{\mathrm{slices}} \triangleq \max\!\left\{\,0,\ \left\lfloor \frac{T' - L_{\mathrm{req}} + \rho}{\Delta}\right\rfloor + 1 \right\},\qquad
    \mathcal{K}\triangleq\{k_m\}_{m=1}^{N_{\mathrm{slices}}}.
\end{equation}

Here, $L_{\mathrm{req}}$ is the required right-side coverage in transformed time under the chosen supervision configuration; $k_m$ enumerates candidate window starts with stride $\Delta$ and warm-start depth $\rho$ (so the earliest start is $1-\rho$); the inequality enforces that each window has sufficient remaining trajectory length; $N_{\mathrm{slices}}$ is the resulting number of admissible windows; and $\mathcal{K}$ is the set of admissible start indices.

\paragraph{Left-padding for the feature history (abstract).}
One challenge when comparing models with different context window sizes is that, without additional measures, changing $L_{\mathrm{seq}}$ alters the (testing) dataset size, leading to unfair comparisons between models. To address this, we introduce left padding.
If $\rho>0$, some history indices may satisfy $k_m+i<1$. We therefore define a left-padded feature extension via an abstract padding operator $\mathcal{P}$:
\begin{equation}
\widetilde{\vz}(j)=
\begin{cases}
\vz(j), & j\ge 1,\\
\mathcal{P}(\gZ,j), & j\le 0.
\end{cases}
\end{equation}

For each admissible start $k_m\in\mathcal{K}$, the feature window is
\begin{equation}
    \vW_m = [\widetilde{\vz}(k_m), \widetilde{\vz}(k_m+1), \dots, \widetilde{\vz}(k_m+L_{\mathrm{seq}}-1)]^\top \in \mathbb{R}^{L_{\mathrm{seq}} \times F}.
\end{equation}

For window-level supervision, the supervision index and label are
\begin{equation}
    j_{\mathrm{sup}}(k_m)=k_m+L_{\mathrm{seq}}-1+\delta, \qquad y_m = z_y\!\big(j_{\mathrm{sup}}(k_m)\big).
\end{equation}
The label is guaranteed to come from a real (non-padded) target index for all $m$ provided $\rho \le L_{\mathrm{seq}}-1+\delta$.

The resulting windowed dataset for a single unit is
\begin{equation}
    \mathcal{D}_{\mathrm{seq}} \triangleq \{(\vW_m, y_m)\}_{m=1}^{N_{\mathrm{slices}}}.
\end{equation}

The output of this stage is the set of temporal windows $\mathcal{W}=\{\vW_m\}_{m=1}^{N_{\mathrm{slices}}}$ paired with labels $\{y_m\}_{m=1}^{N_{\mathrm{slices}}}$.

\begin{conclusionbox}
\textbf{Example.}
Each extracted window $\vW_m=[\vz(m),\dots,\vz(m+L_{\mathrm{seq}}-1)]^\top\in\mathbb{R}^{L_{\mathrm{seq}}\times F}$ stacks $L_{\mathrm{seq}}$ consecutive feature vectors and is paired with the ah-RUL label $y_m=z_y(m+L_{\mathrm{seq}}-1)$ at the window end. With stride $\Delta=1$, consecutive windows shift by one step, so labels decrease slowly as the battery ages. A time-series Transformer (or LSTM) consumes $\vW_m$ directly as its input sequence, treating each row as one time step.
\end{conclusionbox}

\subsubsection{ Tabularization Schema ($\mathcal{T}$)}

To support heterogeneous model classes, we define two representations of each sample. Sequence models (e.g., LSTMs and Transformers) operate directly on the temporal window $\vW_m \in \mathbb{R}^{L_{\mathrm{seq}} \times F}$. By contrast, tabular foundation models are pre-trained on row-wise tabular data, where each example is represented by a fixed-dimensional feature vector and inference is performed over sets of such rows. They therefore assume no explicit sequence axis in the input. For compatibility with this input schema, each window $\vW_m$ is mapped to a vector representation that preserves all entries and encodes temporal order through a fixed column ordering.

We define the tabularization operator $\mathcal{T}: \mathbb{R}^{L_{\mathrm{seq}} \times F} \to \mathbb{R}^{D_{\mathrm{tab}}}$ to transform the sequence $\vW_m$ into a flat tabular sample $\vX_m$, where $D_{\mathrm{tab}}=L_{\mathrm{seq}}F$:
\begin{equation}
    \vX_m = \mathcal{T}(\vW_m) = [z_1(k_m), \dots, z_F(k_m), \dots, z_1(k_m+L_{\mathrm{seq}}-1), \dots, z_F(k_m+L_{\mathrm{seq}}-1)]^\top,
\end{equation}
where $z_f(j)$ denotes the $f$-th component of $\vz(j)$ and we use a time-major ordering for reproducibility.
The labels remain paired with the tabularized rows; only the input window is flattened by $\mathcal{T}$.

\begin{conclusionbox}
\textbf{Example.}
The window $\vW_m\in\mathbb{R}^{L_{\mathrm{seq}}\times F}$ is flattened in time-major order to yield the tabular vector $\vX_m=\mathcal{T}(\vW_m)\in\mathbb{R}^{D_{\mathrm{tab}}}$ with $D_{\mathrm{tab}}=L_{\mathrm{seq}}F$. The first $F$ entries encode the windowed statistics at the earliest step in the window, and the last $F$ entries encode those at the most recent step. The pair $(\vX_m,y_m)$ is one element of $\mathcal{D}_{\mathrm{tab}}$. At inference time, TabPFN or TabDPT conditions on an admissible training-only context set $\mathcal{C}\subset\mathcal{D}_{\mathrm{tab}}^{\text{train}}$ to predict the ah-RUL of a query battery window.
\end{conclusionbox}

\subsubsection{Dual Dataset Representation}

The application of the transformation pipeline yields two information-equivalent (up to reshaping) sample representations: the temporal window $\vW_m$ and its tabularized counterpart $\vX_m$. At the dataset level, these induce the sequence dataset $\mathcal{D}_{\mathrm{seq}}$ and the tabular dataset $\mathcal{D}_{\mathrm{tab}}$. The final dimension of the tabular vector is $D_{\mathrm{tab}} = L_{\mathrm{seq}}F$. One may then select the representation that matches the target model class:

\begin{itemize}
    \item \textbf{Sequence Dataset:} $\mathcal{D}_{\mathrm{seq}} = \{(\vW_m, y_m)\}_{m=1}^{N_{\mathrm{slices}}}$, which retains the explicit temporal matrix structure produced by applying $\mathcal{G}$ and slicing $\mathcal{S}$ to the raw data and using the aligned target series $\gY'$.
    \item \textbf{Tabular Dataset:} $\mathcal{D}_{\mathrm{tab}} = \{(\vX_m, y_m)\}_{m=1}^{N_{\mathrm{slices}}}$, which extends the pipeline with the flattening operator $\mathcal{T}$, making the data structurally compatible with tabular learning algorithms.
\end{itemize}

These formats contain identical information content, allowing for flexible model selection without data loss. Both representations inherit the same admissible start set $\mathcal{K}$ and the same split membership for each sample. The total number of samples $N_{\mathrm{slices}}$ in both $\mathcal{D}_{\mathrm{seq}}$ and $\mathcal{D}_{\mathrm{tab}}$ is directly controlled by the slicing stride $\Delta$. By adjusting $\Delta$, we can effectively perform subsampling to manage computational constraints or reduce redundancy without altering the fundamental structure of the samples.

\paragraph{Context sets for tabular foundation models (leakage constraints).}
When using tabular foundation models with in-context learning, predictions may take the form $\hat{y}_m=f_{\phi}(\vX_m;\mathcal{C})$, where $\mathcal{C}=\mathcal{D}_{\mathrm{tab}}^{\text{train}}$ is the set of in-context samples. To prevent leakage and ensure comparability, $\mathcal{C}$ must be drawn exclusively from the training partition; no validation/test samples may appear in $\mathcal{C}$. Under intra-unit temporal splitting, context selection must additionally respect time, i.e., it must not use samples derived from future timestamps relative to the supervision index $j_{\mathrm{sup}}(k_m)$ of the query sample. The context size and sampling/retrieval strategy are fixed in advance, pre-cached for increased performance and applied consistently across methods.

\subsubsection{Generalization to Multiple Units and Partitions}

For a dataset with $U$ heterogeneous units, we apply the operators $\mathcal{G}$, $\mathcal{S}$, and $\mathcal{T}$ independently to each unit $u$. To learn across multiple units, we define the global dataset $\mathcal{D}_{global}$ as the union of these processed unit-specific subsets. Depending on the chosen representation (sequence or tabular), this is defined as:
\begin{equation}
    \mathcal{D}_{global} = \bigcup_{u=1}^{U} \mathcal{D}_{unit}^{(u)}
\end{equation}
where $\mathcal{D}_{unit}^{(u)}$ is either $\mathcal{D}_{\mathrm{seq}}^{(u)}$ or $\mathcal{D}_{\mathrm{tab}}^{(u)}$.

We formally define the partitioning of the global dataset into disjoint training, validation, and testing subsets:
\begin{equation}
    \mathcal{D}_{global} = \mathcal{D}^{\text{train}}_{\text{global}} \cup \mathcal{D}^{\text{val}}_{\text{global}} \cup \mathcal{D}^{\text{test}}_{\text{global}}
\end{equation}
This partitioning can be performed using two distinct strategies depending on the application constraints:
\begin{itemize}
    \item \textbf{Inter-Unit Splitting:} The set of units $\{1, \dots, U\}$ is disjointly divided. All samples derived from a specific unit belong exclusively to one partition. This tests generalization to unseen machines.
    \item \textbf{Intra-Unit Splitting:} The splitting occurs along the temporal dimension within each unit. This evaluates generalization across time on known machines under a causal protocol (no future information in training-time preprocessing).
\end{itemize}

\textbf{Prevention of Data Leakage:} Crucially, the transformation parameters $\Psi = \{\psi_1, \dots, \psi_P\}$ (introduced in Stage 1) are estimated \textit{solely} from the training partition $\mathcal{D}^{\text{train}}_{\text{global}}$. Likewise, any fitted target parameters $\Phi$ and any choices inside $\widetilde{\mathcal{H}}$ (including alignment/aggregation design and any statistics it requires) are determined using only the training partition. These fixed parameters are then applied to transform the validation and test partitions, ensuring no information from future or unseen data influences the feature extraction process. For intra-unit splitting, ``training partition'' refers to the training time range within each unit, so that no future timestamps contribute to fitted preprocessing statistics.

Due to the frameworks abstraction layers, models receive individual windows $\vW_m$ identically regardless of which unit they originate from. However, the framework is unit-sensitive: unit identities are tracked throughout the pipeline to support per-unit evaluation (e.g., per-unit MAE on battery and bearing datasets; Appendix~\ref{app:prog_per_unit}). For clarity and without loss of generality, throughout the remainder of this work we refer to the datasets as $\mathcal{D}_{\mathrm{seq}}$ or $\mathcal{D}_{\mathrm{tab}}$ without explicit mention of the global split superscripts, unless the context requires distinguishing between partitions.

\subsubsection{Task-Specific Target Definitions}

The goal of the learning task is to map each input window $\vW_m$ (or its tabular representation $\vX_m$) to an associated target variable. The precise definition of this target depends on the specific PHM task and how supervision is aligned with the input window. In particular, targets may represent either future system evolution or the current system state, and can be defined at a specific time index or aggregated over the window.

\begin{itemize}
    \item \textbf{Prognostics:} The target is a scalar $y_m \in \mathbb{R}_{\geq 0}$, representing a monotonically decreasing value such as the Remaining Useful Life (RUL) associated with the end of the window $\vW_m$. Unless otherwise stated, supervision is read at the window-level supervision index $j_{\mathrm{sup}}(k_m)=k_m+L_{\mathrm{seq}}-1+\delta$, and the label is defined as $y_m=z_y\!\big(j_{\mathrm{sup}}(k_m)\big)$. The mapping from $j_{\mathrm{sup}}(k_m)$ to raw time (and any windowing/aggregation) is governed by $a\!\big(j_{\mathrm{sup}}(k_m)\big)$ and the target pipeline $\widetilde{\mathcal{H}}$.
    
    \item \textbf{Diagnostics:} The target is a discrete class label $y_m \in \{0, 1, \dots, K-1\}$, indicating the specific fault type present within the window $\vW_m$. If the diagnostic label is defined over the window (rather than pointwise), this is implemented within $\widetilde{\mathcal{H}}$ by choosing $\mathcal{A}$ to aggregate labels over the supports corresponding to the input window (e.g., last-value, max-over-window, or majority vote), rather than implicitly redefining $\mathcal{H}$.
\end{itemize}

\section{Training and Implementation}

\subsection{Supervised Learning with Sequential Models}

Conventional sequence models, such as 1D-CNNs, LSTMs, and Time-Series Transformers, are trained to minimize a task-specific loss over the entire sequence dataset $\mathcal{D}_{\mathrm{seq}}^{\text{train}}$. These models do not operate on the tabularized data $\vX_k$; instead, they ingest the temporal sequence $\vW_k$ directly.

Let $f_\theta$ denote a model parameterized by $\theta$. The model produces a prediction $\hat{y}_k = f_\theta(\vW_k)$. The optimal parameters $\theta^*$ are obtained by minimizing the empirical risk over the training sequence dataset:
\begin{equation}
    \theta^* = \operatorname*{arg\,min}_{\theta} \frac{1}{|\mathcal{D}_{\mathrm{seq}}^{\text{train}}|} \sum_{(\vW_k, y_k) \in \mathcal{D}_{\mathrm{seq}}^{\text{train}}} \ell\left(f_\theta(\vW_k), y_k\right)
\end{equation}
where $\ell(\cdot, \cdot)$ is a task-dependent loss function (e.g., Mean Squared Error for prognostics or Cross-Entropy for diagnostics). We typically employ stochastic gradient descent variants (e.g., AdamW) to solve this optimization problem.

\subsection{Tabular Foundation Models: The In-Context Learning Paradigm}

Tabular Foundation Models (e.g., TabPFN) fundamentally operate via In-Context Learning (ICL). In this paradigm, the model $f_\phi$, parameterized by weights $\phi$, makes predictions for a query instance $\vX_q$ by conditioning on a context set $\mathcal{C}$ of labeled support examples. The prediction is given by:
\begin{equation}
    \hat{y}_q = f_\phi(\mathcal{C}, \vX_q)
\end{equation}
Crucially, to ensure a valid evaluation and prevent data leakage, the context set $\mathcal{C}$ must always be drawn strictly from the training partition of the tabular dataset:
\begin{equation}
    \mathcal{C} = \{(\vX_{j}, y_{j})\}_{j=1}^{C} \subset \mathcal{D}_{\mathrm{tab}}^{\text{train}}
\end{equation}
This mechanism allows the model to adapt its posterior distribution to the specific task at inference time. We utilize this general paradigm in two distinct modes:

\subsubsection{Pre-trained In-Context Learning (Zero-Shot)}
\label{sec:pre-trained-in-context-learning}
We use the foundation model with its original, pre-trained weights $\phi$ (frozen). The model adapts to the specific PHM task solely through the information provided in the context set $\mathcal{C}$ (drawn from $\mathcal{D}_{\mathrm{tab}}^{\text{train}}$) at inference time.
\begin{equation}
    \hat{y}_q = f_{\phi}(\mathcal{C}, \vX_q)
\end{equation}
The context size is typically restricted (e.g., $|\mathcal{C}| \leq 4 \times 10^{4}$). If the available training data in $\mathcal{D}_{\mathrm{tab}}^{\text{train}}$ exceeds this limit, we construct $\mathcal{C}$ by subsampling the tabularized training windows.

For the data-efficiency experiments, we vary this subsampling explicitly through a subset ratio $r_{\mathrm{sub}} \in (0,1]$, applied after sequence construction and tabularization. With aggregate subsampling, $\max(1,\lfloor r_{\mathrm{sub}} N \rfloor)$ windows are sampled uniformly without replacement from the $N$ eligible training windows, and the selected indices are sorted before constructing the context or training set. With blockwise subsampling, the same total number of windows is distributed over $B$ contiguous index blocks with randomly sampled block lengths and gaps. The selected windows therefore remain locally contiguous while covering only parts of the training trajectories. This operation reduces the number of rows supplied to the model, but it does not alter the feature dimension, sequence length, or contents of any selected window. Test windows are not subsampled in these experiments.

\subsection{Evaluation Protocol}
\label{sec:evaluation_protocol}

To ensure a fair and standardized comparison across disparate modeling architectures, we employ a unified evaluation protocol. Regardless of whether a model consumes sequential data ($\vW_k \in \mathcal{D}_{\mathrm{seq}}$) or tabular data ($\vX_k \in \mathcal{D}_{\mathrm{tab}}$), all metrics are calculated on the same set of test instances derived from the held-out partition $\mathcal{D}^{\text{test}}$.

We define a task-specific loss function $\ell(\hat{y}, y)$ (e.g., squared error for regression or 0-1 loss for classification). The aggregate test loss $\mathcal{L}_{test}$ is computed over the test set, ensuring a consistent evaluation logic for both model families:
\begin{equation}
    \mathcal{L}_{test} = \frac{1}{|\mathcal{D}^{\text{test}}|} \sum_{k \in \mathcal{D}^{\text{test}}} \ell(\hat{y}_k, y_k), \quad \text{where } \hat{y}_k = 
    \begin{cases} 
        f_\theta(\vW_k) & \text{if } f \text{ is Sequential} \\
        f_\phi(\mathcal{C}, \vX_k) & \text{if } f \text{ is Tabular}
    \end{cases}
\end{equation}
For Prognostics, we employ standard regression metrics such as Root Mean Squared Error (RMSE) and Mean Absolute Error (MAE). For Diagnostics, we utilize classification metrics including Accuracy and Macro-F1 Score to account for class imbalance. For cross-dataset comparisons, we report the average rank obtained by ranking models within each dataset and averaging the resulting ranks across datasets, with rank 1 denoting the best model for a given dataset.

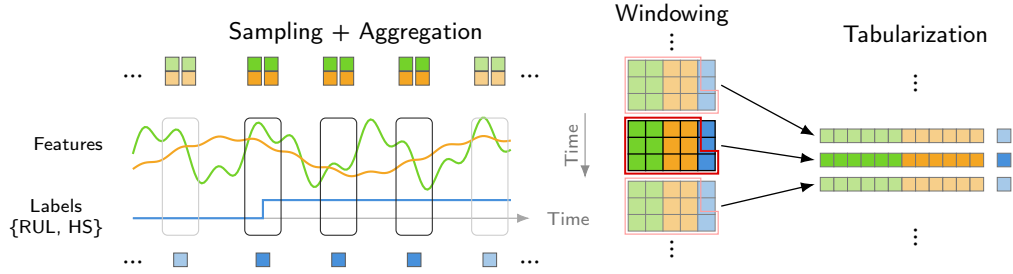
\begin{figure}[pos=t]
\centering
\begin{tikzpicture}[
    x=1cm,
    y=1cm,
    >=Latex,
    signal/.style={line width=0.8pt},
    window/.style={draw, rounded corners=2pt, minimum width=0.48cm, minimum height=1.55cm},
    cell/.style={draw=black!65, line width=0.25pt, minimum width=0.18cm, minimum height=0.18cm, inner sep=0pt},
    rowcell/.style={draw=black!55, line width=0.2pt, minimum width=0.18cm, minimum height=0.18cm, inner sep=0pt},
    smalllabel/.style={font=\scriptsize},
    titlelabel/.style={font=\small},
]
\definecolor{tabphmgreen}{RGB}{116,210,42}
\definecolor{tabphmlightgreen}{RGB}{190,228,145}
\definecolor{tabphmorange}{RGB}{244,166,35}
\definecolor{tabphmlightorange}{RGB}{247,207,138}
\definecolor{tabphmblue}{RGB}{72,145,220}
\definecolor{tabphmlightblue}{RGB}{166,201,233}

\node[titlelabel] at (2.95,3.18) {Sampling + Aggregation};
\node[titlelabel] at (7.13,3.42) {Windowing};
\node[titlelabel] at (10.35,3.18) {Tabularization};

\node[smalllabel, anchor=east] at (-0.28,1.72) {Features};
\node[smalllabel, anchor=east, align=center] at (-0.28,0.73) {Labels\\\{RUL, HS\}};

\draw[->, black!35] (0,0.73) -- (5.25,0.73);
\node[smalllabel, text=black!55, anchor=west] at (5.36,0.73) {Time};
\draw[signal, tabphmgreen] plot[domain=0:5.0, samples=160]
    (\x,{1.58 + 0.33*sin(250*\x) + 0.16*sin(720*\x)});
\draw[signal, tabphmorange] plot[domain=0:5.0, samples=160]
    (\x,{1.55 + 0.24*sin(105*\x - 45) + 0.03*sin(720*\x)});
\draw[signal, tabphmblue] (0,0.73) -- (1.72,0.73) -- (1.72,0.97) -- (5.0,0.97);
\foreach \x/\shade in {0.63/20,1.72/85,2.72/85,3.72/85,4.72/20} {
    \node[window, draw=black!\shade] at (\x,1.28) {};
}

\foreach \x/\g/\o/\b in {
    0.63/tabphmlightgreen/tabphmlightorange/tabphmlightblue,
    1.72/tabphmgreen/tabphmorange/tabphmblue,
    2.72/tabphmgreen/tabphmorange/tabphmblue,
    3.72/tabphmgreen/tabphmorange/tabphmblue,
    4.72/tabphmlightgreen/tabphmlightorange/tabphmlightblue} {
    \foreach \dx in {-0.11,0.11} {
        \node[cell, fill=\g] at (\x+\dx,2.77) {};
        \node[cell, fill=\o] at (\x+\dx,2.58) {};
    }
    \node[cell, fill=\b] at (\x,0.18) {};
}
\node[font=\small] at (0.0,2.58) {$\cdots$};
\node[font=\small] at (5.25,2.58) {$\cdots$};
\node[font=\small] at (0.0,0.12) {$\cdots$};
\node[font=\small] at (5.25,0.12) {$\cdots$};

\draw[->, black!45] (5.98,2.12) -- (5.98,1.26);
\node[smalllabel, rotate=90, text=black!55] at (5.80,1.69) {Time};
\foreach \yy/\gcol/\ocol/\bcol/\gridcol/\gridwidth/\framecol/\framewidth in {
    2.16/tabphmlightgreen/tabphmlightorange/tabphmlightblue/black!50/0.30pt/red!38/0.45pt,
    1.36/tabphmgreen/tabphmorange/tabphmblue/black/0.50pt/red!85!black/0.75pt,
    0.56/tabphmlightgreen/tabphmlightorange/tabphmlightblue/black!50/0.30pt/red!38/0.45pt} {
    \begin{scope}[shift={(6.55,\yy)}, line join=miter, line cap=rect]
        \def\cw{0.23}
        \def\ch{0.22}
        \path[fill=\gcol] (0,0) rectangle ({2*\cw},{3*\ch});
        \path[fill=\ocol] ({2*\cw},0) rectangle ({4*\cw},{3*\ch});
        \path[fill=\bcol] ({4*\cw},0) rectangle ({5*\cw},{3*\ch});
        \draw[\gridcol, line width=\gridwidth] (0,0) rectangle ({5*\cw},{3*\ch});
        \foreach \col in {1,...,4} {
            \draw[\gridcol, line width=\gridwidth] ({\col*\cw},0) -- ({\col*\cw},{3*\ch});
        }
        \foreach \row in {1,2} {
            \draw[\gridcol, line width=\gridwidth] (0,{\row*\ch}) -- ({5*\cw},{\row*\ch});
        }
        \draw[draw=\framecol, line width=\framewidth]
            (-0.04,-0.04) -- ({5*\cw+0.04},-0.04)
            -- ({5*\cw+0.04},{\ch+0.04})
            -- ({4*\cw+0.04},{\ch+0.04})
            -- ({4*\cw+0.04},{3*\ch+0.04})
            -- (-0.04,{3*\ch+0.04}) -- cycle;
    \end{scope}
}
\node[font=\small] at (7.13,3.06) {$\vdots$};
\node[font=\small] at (7.13,0.33) {$\vdots$};
\foreach \yy/\g/\o/\b in {1.82/tabphmlightgreen/tabphmlightorange/tabphmlightblue,1.50/tabphmgreen/tabphmorange/tabphmblue,1.18/tabphmlightgreen/tabphmlightorange/tabphmlightblue} {
    \foreach \i in {0,...,5} {
        \node[rowcell, fill=\g] at (9.18+0.18*\i,\yy) {};
    }
    \foreach \i in {0,...,5} {
        \node[rowcell, fill=\o] at (10.26+0.18*\i,\yy) {};
    }
    \node[rowcell, fill=\b] at (11.52,\yy) {};
}
\draw[->, line width=0.5pt] (7.78,2.49) -- (9.02,1.82);
\draw[->, line width=0.5pt] (7.78,1.69) -- (9.02,1.50);
\draw[->, line width=0.5pt] (7.78,0.89) -- (9.02,1.18);
\node[font=\small] at (10.35,2.52) {$\vdots$};
\node[font=\small] at (10.35,0.48) {$\vdots$};
\end{tikzpicture}
\caption{Our tabularization scheme. Time-series features are aggregated into their statistics on a per-window basis, along with the corresponding labels. These are then composed into a row in our table, providing richer context.}
\label{fig:phm_tabularization_tikz}
\end{figure}

\section{Practical Implementation Details}

\subsection{Unified Evaluation Pipeline Details}

All baselines are trained from scratch and evaluated alongside all foundation models using the evaluation infrastructure of \citet{telyatnikov2026picid}, under identical preprocessing, splitting, windowing, and metric computation (Section~\ref{sec:evaluation_protocol}), ensuring that performance differences reflect modeling choices rather than data preparation. Model training is performed using \texttt{PyTorch} with its implementation of the \texttt{AdamW} optimizer. The framework uses an adapted version of the \texttt{Trainer} provided by \texttt{Lightning}\footnote{\url{https://lightning.ai/docs/pytorch/stable/}}, extended to accommodate models that use a Scikit-learn-like \texttt{fit()} and \texttt{predict()} API.
Our training regimen for transformer models incorporates a custom learning rate scheduling that includes a warm-up phase and reduces the learning rate upon reaching a plateau, as this step has improved performance in the original forecasting tasks. We use min-max or z-score normalization, depending on the dataset. All preprocessing, feature extraction, and sequence slicing, represented by the operators $\mathcal{G}$ and $\mathcal{S}$, are shared among the models. The tabular foundation models and XGBoost share the same tabularization schema $\mathcal{T}$.
Hyper-parameter optimization was performed using grid search,
with initial parameter ranges selected from the original works.
Hyper-parameter tuning, which includes the number of columns and rows for tabular in-context learning, was performed on the validation set.
Dataset split and target-construction metadata are summarized in Section~\ref{sec:datasets} (e.g., Tables~\ref{tab:nasa_split}, \ref{tab:unibo_split}, \ref{tab:xjtu_lifetimes}, and \ref{tab:cmapss_split}).
We report means and standard deviations computed over five random seeds.

\subsection{Benchmark Models}
\label{subsec:model_families}
This section introduces the model families considered in our study and formalizes the data-processing pipeline used to compare them fairly. Since our objective is to benchmark \emph{tabular foundation models} against established \emph{time-series models} on PHM tasks, the key challenge is to construct two representations of the same underlying problem: a sequence representation for temporal models and a tabular representation for foundation models. For completeness, we also include strong non-transformer baselines commonly used in PHM.

\subsubsection{Tabular Foundation Models}
\label{subsubsec:tfms}

Recent progress in Prior-Fitted Networks has led to a growing family of tabular foundation models (TFMs), including several alternatives to the original TabPFN framework (e.g.,  \cite{quTabICL2025, arbelEquiTabPFN2025}).  In this work, we evaluate the following two representatives:

\begin{itemize}
    \item \textbf{TabPFN:} Tabular Prior-Fitted Networks \citep{hollmanntabpfn, hollmann2025accurate}, are  designed to approximate Bayesian inference on tabular datasets. Trained on large-scale causally-generated synthetic data, TabPFN supports both classification and regression through in-context learning and often outperforms strong tree-based methods. More details are given in Appendix~\ref{app:model_tabpfn_tabdpt}.

    \item \textbf{TabDPT:} A transformer-based tabular foundation model trained on real-world datasets, that uses retrieval-based self-supervised pre-training and in-context learning to generalize to unseen tabular datasets, including both classification and regression, with no task-specific training or hyper-parameter tuning \citep{ma2024tabdpt}. More details are given in Appendix~\ref{app:model_tabpfn_tabdpt}.
\end{itemize}

\subsubsection{Transformer models}
For our experiments, we benchmark tabular foundation models against  
state-of-the-art (SOTA) models in long-range time-series forecasting. These models had to be adapted to perform regression and classification, as described in Section 3. Within  the family of transformer-based models, these are:
\begin{itemize}
    \item \textbf{PatchTST:} A channel-independent “patching” Transformer that tokenizes univariate time series into subseries-level patches, enabling longer receptive fields with lower attention cost \citep{Yuqietal-2023-PatchTST}.

    \item \textbf{Crossformer:} A Transformer for multivariate time series that tokenizes inputs with cross-dimensional embeddings and applies a two-stage attention layer to model both cross-time and cross-feature dependencies \citep{zhang2023crossformer}. 
    
    \item \textbf{Spacetimeformer:} A long-range Transformer that jointly learns temporal and spatial interactions by treating spatiotemporal values as tokens, combining sequence and graph-like reasoning \citep{grigsby2021longrange}.

\end{itemize}

For completeness, we include non-transformer-based models in our comparisons. These encompass classical ML architectures such as CNNs and LSTMs \citep{lecunDeepLearning2015}, which have found wide adoption in the industry \citep{leiMachineryHealthPrognostics2020,wangDeepLearningSmartManufacturing2020}, as well as statistical and tabular fit-predict baselines, and more recent models:
\begin{itemize}
\item \textbf{1D-CNN:} Convolution-based models that slide learnable kernels over temporal inputs to extract local trends, scale efficiently over long histories, and capture multi-resolution features through residual blocks.
\item \textbf{LSTM:} Recurrent networks with gated memory cells that propagate and update a hidden state step-by-step, enabling nonlinear modeling of sequences with adaptive retention of long-term information. We implement the bi-direction variant, which  integrates future covariates effectively \citep{siami-naminiPerformanceLSTMBiLSTM2019}
\item \textbf{Statistical regression baselines:} Linear, polynomial, and exponential regression are included as low-capacity reference models.
\item \textbf{MLP:} A multilayer perceptron is included as a non-sequential neural baseline. It uses fully connected layers to model nonlinear interactions among the flattened temporal features.
\item \textbf{XGBoost:} A classical tree-based baseline for tabular tasks. XGBoost uses gradient boosting with regularized decision trees and is evaluated on the same tabularized rows as the tabular foundation models \citep{chen2016xgboost}. 
\item \textbf{TiDE}, a recently introduced dense residual model, built on MLP-based encoder–decoders and quasi-linear networks for long-term forecasting 
\citep{daslong}. 

\end{itemize}

\subsection{Adapting Baseline Models to Prognostics and Diagnostics tasks in PHM}

The sequence models considered here---Bi-LSTM, 1D-CNN, TiDE, and all Transformer-based architectures---were originally designed for time-series forecasting and are applied directly to the sequential input $\vW_k$. We adapt them to PHM regression and classification tasks by replacing their forecasting heads with task-specific regression or classification heads.

In addition, the original encoder decoder forecasting formulation does not directly apply to prognostics and diagnostics tasks in PHM. The past values of the target variable, namely the remaining useful life, are unavailable at inference time because the degradation and thus the time to failure are unknown prior to asset breakdown. Moreover, no future information is considered, as the objective is to regress onto the target at the same time index $k$.
We therefore modify encoder-decoder transformer architectures by removing the encoder and directly feeding the sequential data $\vW_k$ to the decoder. This step is not necessary for PatchTST, since it is a decoder-only transformer by design. However, its channel-independent processing cannot be applied to the regression target for the same reason. Therefore, we concatenate decoder latents and train a separate regression or classification head. TiDE introduces an attention-free multilayer perceptron (MLP)-based encoder-decoder architecture. Without lookback data, we disable TiDE's encoder pathway as well and rely only on the model's dynamic covariate pathway to process $\vW_k$. Since all Transformer models produce point predictions for forecasting (auto-regressively), we use the same head for regression by simulating a prediction horizon of one. For diagnostics tasks, we replace the regression head with a classification head that outputs $n$ classes.

\section{Results and Analysis}
\label{sec:empirical_validation}

This section evaluates whether the proposed tabularization enables tabular foundation models to serve as a unified interface for PHM tasks. We compare tabular foundation models against multi-task baselines, including sequence models, transformer models, and a conventional gradient boosting framework, on prognostics and diagnostics benchmarks. We then analyze whether tabularizing PHM tasks is effective in practical scenarios by examining its behavior under limited training data, the role of temporal context retained in the tabular representation, and the structure of the predictive uncertainty produced by TabPFN. Full result tables, transformation schemas, and hyperparameter details are provided in Appendix~\ref{app:additional_results}.

\subsubsection{Overall Benchmark Performance}

We first compare tabular foundation models across twelve diagnostics and progostics scenarios.
Table~\ref{tab:results_main_combined} reports performance for all 13 models across 12 datasets. Prognostics results are evaluated using normalized MAE ($\times 100$; top floor, $\downarrow$), while diagnostics results are evaluated using F1 score ($\times 100$; bottom floor, $\uparrow$), defined according to \cref{sec:evaluation_protocol}.

The prognostics results show a heterogeneous performance landscape: no single architecture dominates all datasets, and strong dataset-specific baselines remain competitive. Nevertheless, the tabular foundation models obtain the best average ranks, with TabDPT ranking first overall (average rank 2.67) and TabPFN second (average rank 3.33). A plausible explanation for this aggregate difference is that TabPFN's synthetic pre-training priors are not designed specifically for PHM data. In contrast, TabDPT is pre-trained on real tabular datasets and constructs row-level contexts. This difference in pre-training and context construction may better match the heterogeneous row distributions induced by PHM tabularization. In addition to the best overall ranking, the tabular foundation models also perform best on certain datasets. TabPFN gives the lowest normalized MAE on PHME20 (1.95 $\pm$ 0.03) and Unibo (3.72 $\pm$ 0.06), while TabDPT gives the lowest normalized MAE on N-CMAPSS Prognostics (NC-P) (6.85 $\pm$ 0.02). Conventional sequence and transformer models remain strongest on selected tasks, with STF leading NC-DS02 (4.89 $\pm$ 0.10), TiDE leading NB14 (3.44 $\pm$ 0.17), and LSTM leading XJTU-SY (21.89 $\pm$ 0.40).

\begin{table}[t]
\centering
\scriptsize
\setlength{\tabcolsep}{4pt}
\renewcommand{\arraystretch}{0.92}
\caption{Main evaluation results. \textbf{Top block:} prognostics, measured by MAE in the normalized target space ($\times 100$) ($\downarrow$). \textbf{Bottom block:} diagnostics, measured by F1 score ($\times 100$) ($\uparrow$). Models are grouped by family: simple baselines, deep sequence models, transformers, tabular models, and tabular foundation models. \textbf{Bold}/\underline{underline} denote the best/second-best result.}
\label{tab:results_main_combined}
\begin{adjustbox}{max width=\textwidth}
\begin{tabular}{lrrrrrrr}
\toprule
Model & \multicolumn{1}{c}{NC-DS02$\downarrow$} & \multicolumn{1}{c}{NC-P$\downarrow$} & \multicolumn{1}{c}{NB14$\downarrow$} & \multicolumn{1}{c}{PHME20$\downarrow$} & \multicolumn{1}{c}{Unibo$\downarrow$} & \multicolumn{1}{c}{XJTU-SY$\downarrow$} & \multicolumn{1}{c}{Avg rank} \\
\midrule
Linear & 10.13 ± 0.14 & 16.11 ± 0.60 & 41.69 ± 12.02 & 12.19 ± 0.36 & 27.59 ± 14.36 & 76.80 ± 60.41 & 12.50 \\
Exp & 5.35 ± 0.06 & 10.96 ± 0.09 & 30.47 ± 47.76 & 8.82 ± 0.52 & 12.19 ± 0.31 & 27.22 ± 4.06 & 9.67 \\
MLP & 6.37 ± 0.23 & 13.17 ± 0.78 & 14.38 ± 9.77 & 4.62 ± 1.15 & 12.50 ± 0.76 & 30.64 ± 2.67 & 10.33 \\
LSTM & \underline{4.93 ± 0.13} & 7.56 ± 0.31 & 3.80 ± 0.22 & 3.73 ± 0.98 & 6.50 ± 0.16 & \cellcolor[gray]{0.85}\textbf{21.89 ± 0.40} & 3.67 \\
CNN-1D & 5.33 ± 0.37 & 7.53 ± 0.22 & 8.89 ± 1.70 & 5.35 ± 3.71 & 12.41 ± 1.15 & 31.02 ± 8.25 & 8.67 \\
TiDE & 5.29 ± 0.22 & 7.62 ± 0.20 & \cellcolor[gray]{0.85}\textbf{3.44 ± 0.17} & 4.20 ± 0.66 & 6.46 ± 0.78 & 25.11 ± 2.38 & 5.17 \\
TST & 5.31 ± 0.13 & \underline{7.02 ± 0.17} & 6.28 ± 0.25 & 4.11 ± 0.84 & 7.23 ± 0.39 & 33.30 ± 7.72 & 7.00 \\
STF & \cellcolor[gray]{0.85}\textbf{4.89 ± 0.10} & 7.35 ± 1.16 & 10.67 ± 3.16 & 3.91 ± 1.00 & 8.89 ± 0.81 & 28.49 ± 4.01 & 6.17 \\
CF & 5.76 ± 0.51 & 9.98 ± 0.57 & \underline{3.57 ± 0.07} & 3.87 ± 0.85 & 5.58 ± 1.08 & \underline{22.09 ± 1.06} & 5.00 \\
PTST & 16.62 ± 0.04 & 21.55 ± 0.03 & 5.22 ± 0.10 & 15.09 ± 1.13 & 11.18 ± 1.11 & 25.42 ± 1.48 & 10.33 \\
XGBoost & 8.52 ± 0.00 & 15.24 ± 0.00 & 4.48 ± 0.00 & 2.68 ± 0.00 & 4.06 ± 0.00 & 24.59 ± 0.00 & 6.50 \\
TabPFN & 4.96 ± 0.04 & 7.79 ± 0.04 & 3.91 ± 0.03 & \cellcolor[gray]{0.85}\textbf{1.95 ± 0.03} & \cellcolor[gray]{0.85}\textbf{3.72 ± 0.06} & 22.27 ± 0.35 & 3.33 \\
TabDPT & 5.07 ± 0.06 & \cellcolor[gray]{0.85}\textbf{6.85 ± 0.02} & 3.63 ± 0.04 & \underline{2.19 ± 0.01} & \underline{3.94 ± 0.05} & 23.24 ± 0.45 & 2.67 \\
\midrule
Model & \multicolumn{1}{c}{NC-D$\uparrow$} & \multicolumn{1}{c}{HSF15-A$\uparrow$} & \multicolumn{1}{c}{HSF15-C$\uparrow$} & \multicolumn{1}{c}{HSF15-P$\uparrow$} & \multicolumn{1}{c}{HSF15-V$\uparrow$} & \multicolumn{1}{c}{MZVAV$\uparrow$} & \multicolumn{1}{c}{Avg rank} \\
\midrule
Linear & 72.34 ± 3.04 & 58.75 ± 1.81 & 98.35 ± 0.82 & 54.40 ± 11.97 & 32.55 ± 2.36 & 39.89 ± 8.77 & 7.33 \\
MLP & 79.49 ± 1.80 & 91.02 ± 2.25 & \underline{99.91 ± 0.14} & 97.32 ± 0.60 & 80.99 ± 29.38 & 60.10 ± 6.39 & 5.00 \\
LSTM & \cellcolor[gray]{0.85}\textbf{88.84 ± 0.73} & 94.59 ± 0.97 & \cellcolor[gray]{0.85}\textbf{100.00 ± 0.00} & 95.94 ± 2.83 & 97.35 ± 3.32 & 51.31 ± 6.01 & 3.83 \\
CNN-1D & \underline{87.53 ± 2.83} & 94.03 ± 1.98 & \cellcolor[gray]{0.85}\textbf{100.00 ± 0.00} & 98.73 ± 0.52 & 97.92 ± 0.83 & \underline{66.11 ± 5.74} & 3.00 \\
TiDE & 32.57 ± 4.65 & 42.90 ± 5.07 & 61.57 ± 10.41 & 59.37 ± 7.68 & 42.11 ± 14.36 & 25.19 ± 5.29 & 8.17 \\
TST & 26.34 ± 3.96 & 37.37 ± 4.38 & 59.18 ± 10.29 & 46.07 ± 4.13 & 35.40 ± 5.77 & 24.93 ± 4.38 & 10.00 \\
STF & 24.55 ± 3.60 & 40.01 ± 5.75 & 65.94 ± 20.61 & 50.39 ± 11.54 & 37.34 ± 5.68 & 38.01 ± 5.56 & 8.67 \\
CF & 23.74 ± 0.93 & 25.04 ± 5.28 & 59.19 ± 10.05 & 29.46 ± 2.61 & 23.98 ± 2.95 & 17.34 ± 4.24 & 11.50 \\
PTST & 19.57 ± 0.26 & 31.56 ± 4.04 & 41.57 ± 5.18 & 41.22 ± 6.63 & 26.20 ± 2.44 & 25.81 ± 4.15 & 11.00 \\
XGBoost & 48.13 ± 0.00 & \underline{98.07 ± 0.00} & \cellcolor[gray]{0.85}\textbf{100.00 ± 0.00} & \underline{99.66 ± 0.00} & \underline{99.65 ± 0.00} & 57.08 ± 0.00 & 3.17 \\
TabPFN & 67.15 ± 1.46 & \cellcolor[gray]{0.85}\textbf{99.47 ± 0.00} & \cellcolor[gray]{0.85}\textbf{100.00 ± 0.00} & \cellcolor[gray]{0.85}\textbf{100.00 ± 0.00} & \cellcolor[gray]{0.85}\textbf{100.00 ± 0.00} & 58.32 ± 2.44 & 2.33 \\
TabDPT & 85.21 ± 0.16 & 96.66 ± 1.03 & \cellcolor[gray]{0.85}\textbf{100.00 ± 0.00} & 99.06 ± 0.25 & 98.92 ± 0.35 & \cellcolor[gray]{0.85}\textbf{71.29 ± 0.48} & 2.33 \\
\bottomrule
\end{tabular}
\end{adjustbox}
\end{table}

The diagnostic results show a similar pattern, with tabular foundation models achieving the best aggregate ranks. TabPFN and TabDPT obtain the best average rank across the six diagnostic tasks, with an average rank of 2.33 for both. TabPFN achieves near-perfect Macro-F1 on the HSF15 component-level tasks, including HSF15-A, HSF15-C, HSF15-P, and HSF15-V where, interestingly, the performance of other transformer model is significantly lacking. TabDPT achieves the highest Macro-F1 on MZVAV. Deep sequence models remain competitive on NC-D, where LSTM and CNN-1D outperform both tabular foundation models. XGBoost remains a strong conventional fit-predict baseline on several HSF15 tasks, but is generally outperformed by the tabular foundation models.  Overall, \cref{tab:results_main_combined} supports the conclusion that tabular foundation models provide the most consistent cross-task performance, expressed through aggregate robustness rather than uniform dominance on every dataset.

\subsubsection{Data Efficiency}
\begin{figure}[pos=htbp]
\centering
\begin{subfigure}[t]{0.45\textwidth}
\centering
\includegraphics[width=\textwidth]{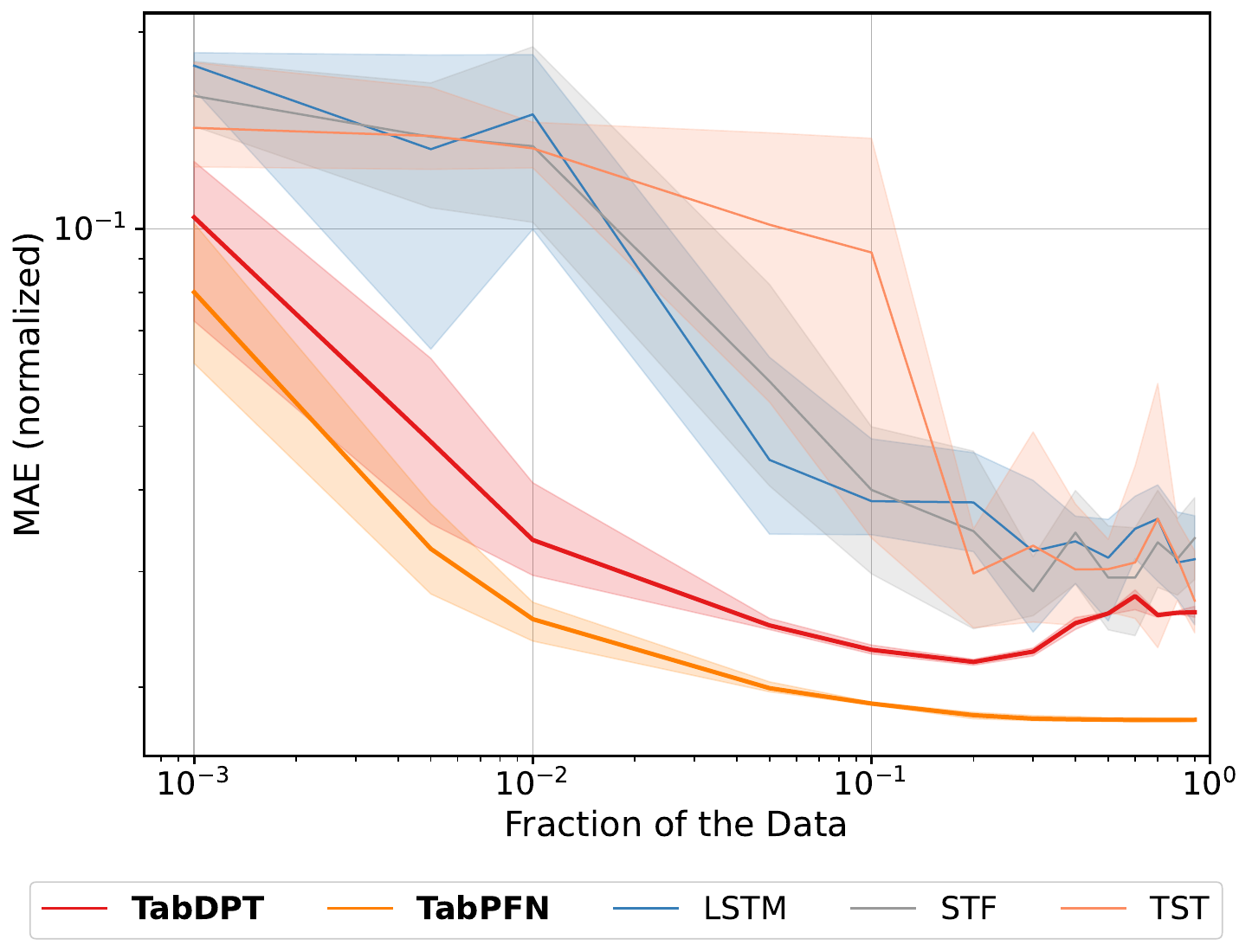}
\caption{PHME20}
\label{fig:scaling_phme20_top5}
\end{subfigure}
\hfill
\begin{subfigure}[t]{0.45\textwidth}
\centering
\includegraphics[width=\textwidth]{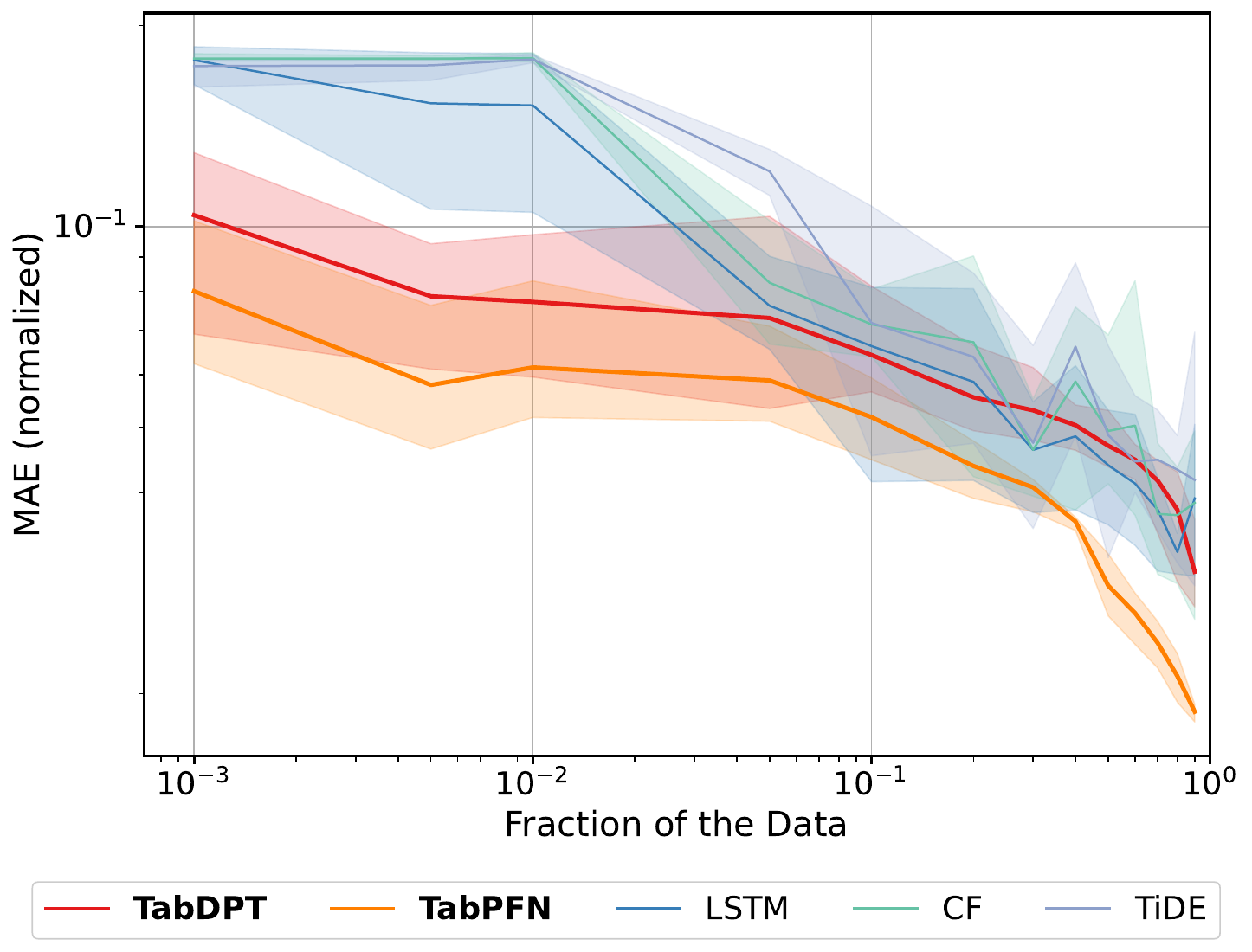}
\caption{PHME20, blockwise}
\label{fig:scaling_phme20_blockwise_top5}
\end{subfigure}

\medskip

\begin{subfigure}[t]{0.45\textwidth}
\centering
\includegraphics[width=\textwidth]{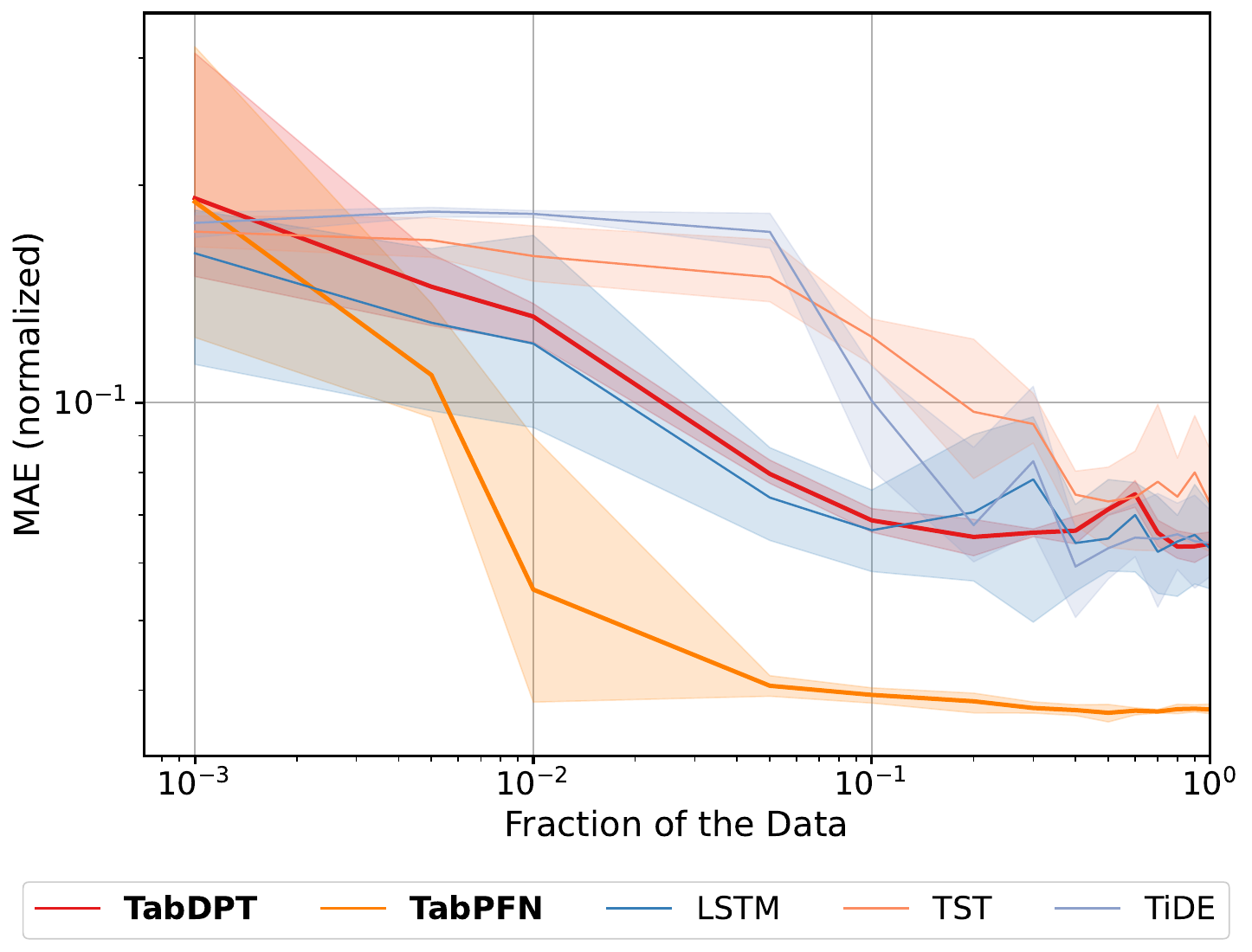}
\caption{Unibo}
\label{fig:scaling_unibo_top5}
\end{subfigure}
\hfill
\begin{subfigure}[t]{0.45\textwidth}
\centering
\includegraphics[width=\textwidth]{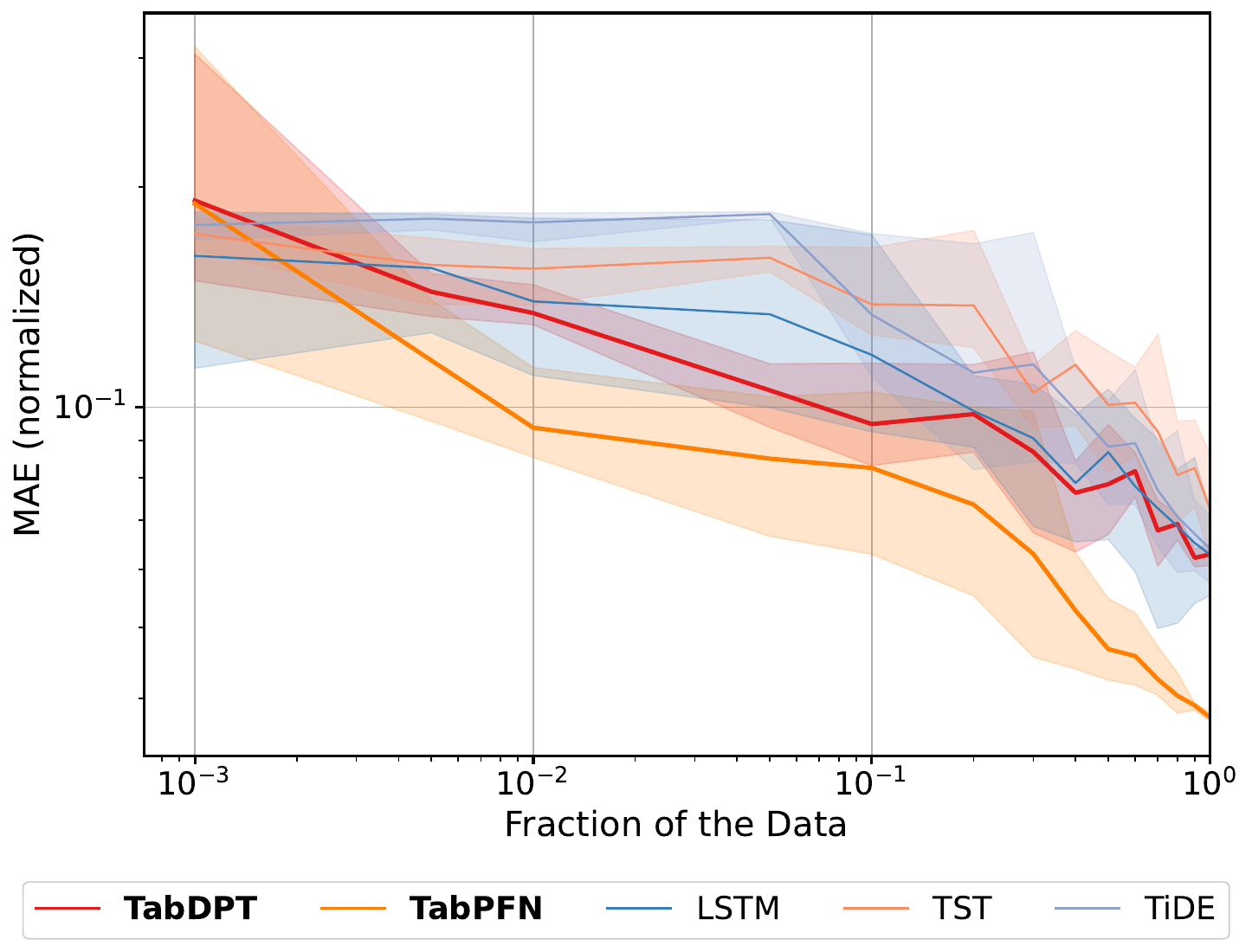}
\caption{Unibo, blockwise}
\label{fig:scaling_unibo_blockwise_top5}
\end{subfigure}

\medskip

\begin{subfigure}[t]{0.45\textwidth}
\centering
\includegraphics[width=\textwidth]{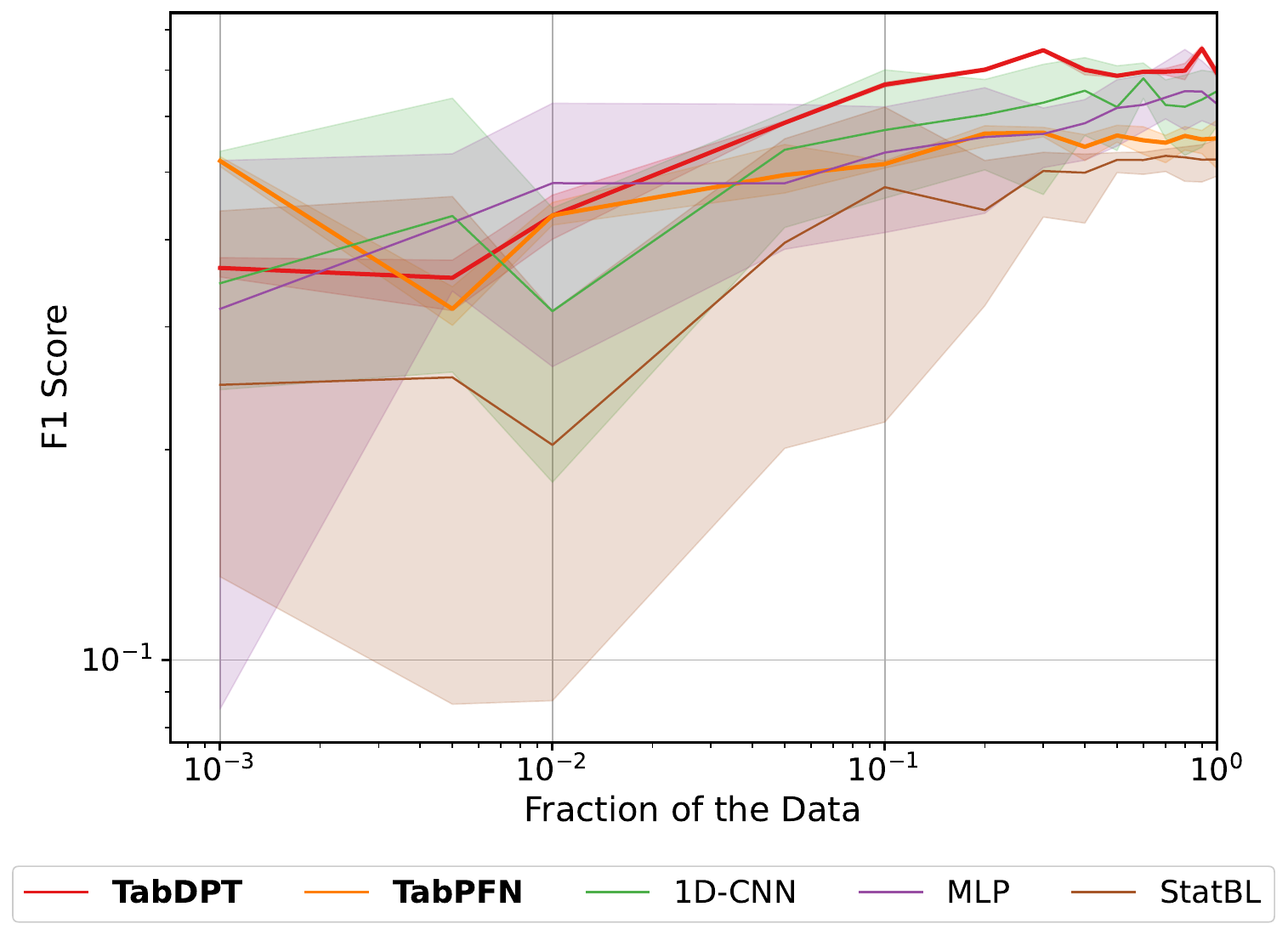}
\caption{MZVAV}
\label{fig:scaling_mzvav_top5}
\end{subfigure}
\hfill
\begin{subfigure}[t]{0.45\textwidth}
\centering
\includegraphics[width=\textwidth]{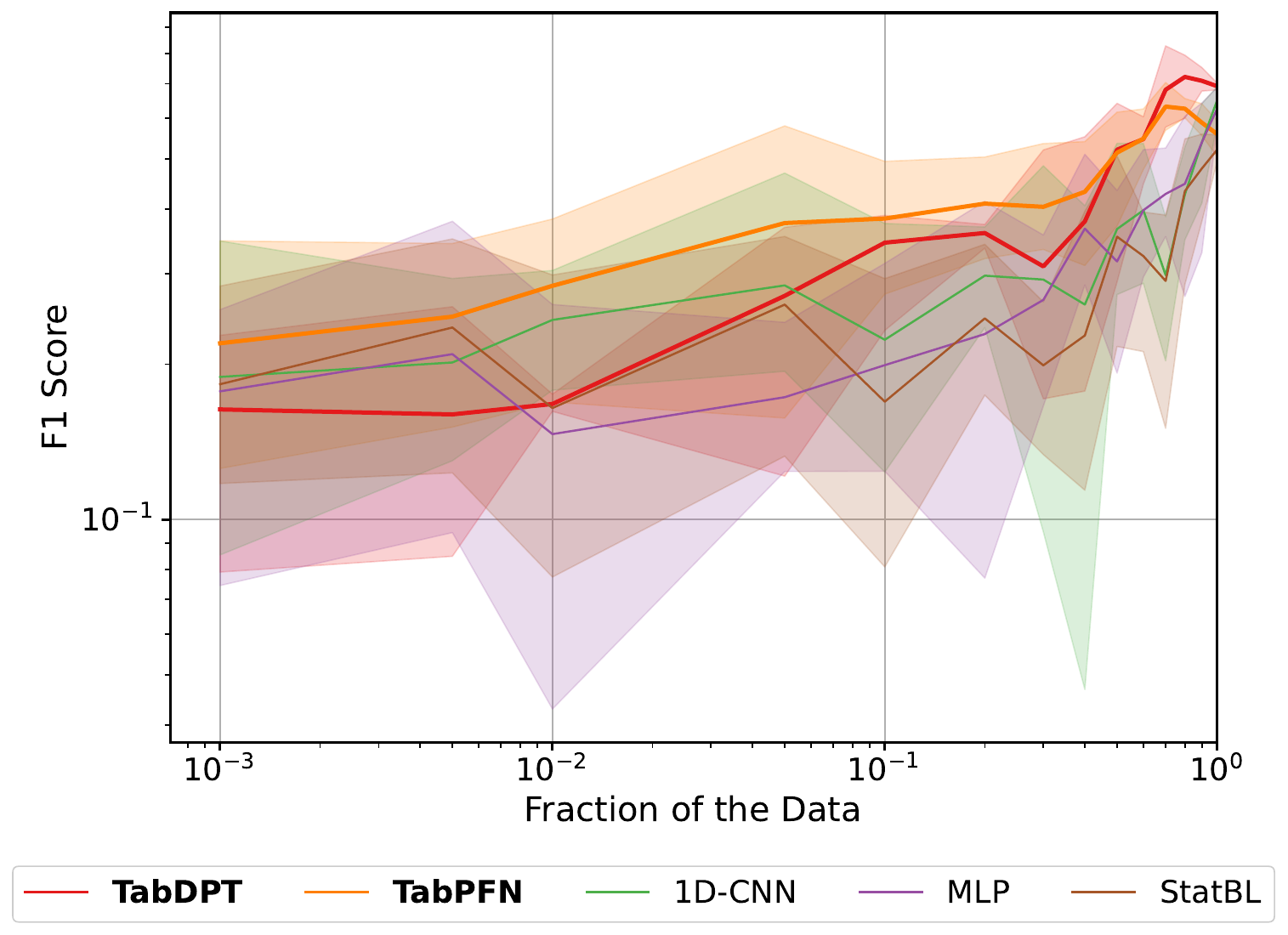}
\caption{MZVAV, blockwise}
\label{fig:scaling_mzvav_blockwise_top5}
\end{subfigure}
\caption{Scaling behavior for PHME20, Unibo, and MZVAV. The left column shows uniformly subsampled scaling laws; the right column shows blockwise subsampled scaling laws. We show the top 5 models for each dataset, which include TabPFN and TabDPT in all cases. A complete version with all models is included in the Appendix as \cref{fig:scaling_phme20_unibo_mzvav}.}
\label{fig:scaling_phme20_unibo_mzvav_top5}
\end{figure}

\begin{figure}[pos=htbp]
\centering
\begin{subfigure}[t]{0.45\textwidth}
\centering
\includegraphics[width=\textwidth]{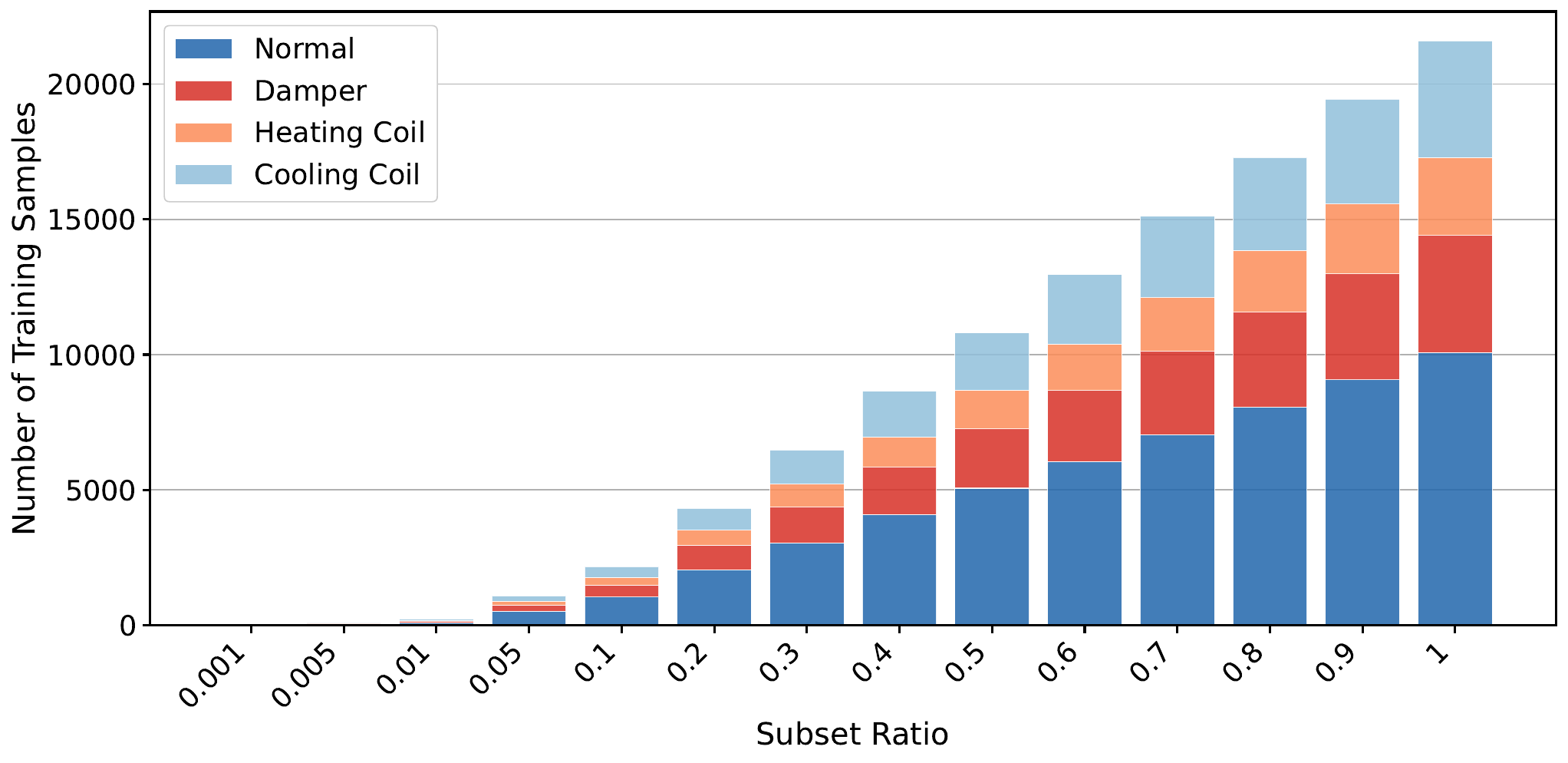}
\caption{Aggregate}
\label{fig:mzvav_class_balance_subset_ratio_aggregate}
\end{subfigure}
\hfill
\begin{subfigure}[t]{0.45\textwidth}
\centering
\includegraphics[width=\textwidth]{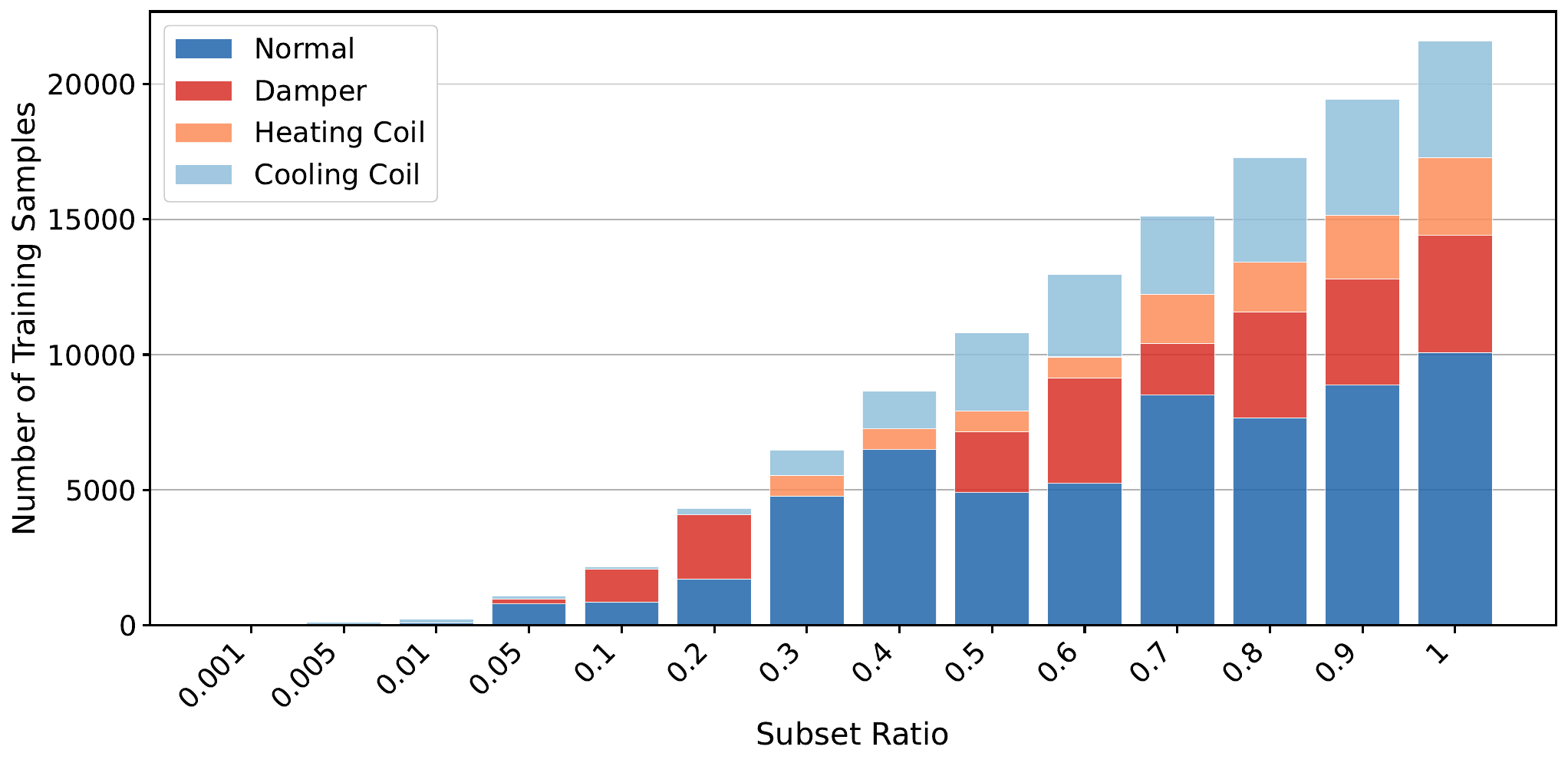}
\caption{Blockwise}
\label{fig:mzvav_class_balance_subset_ratio_blockwise}
\end{subfigure}
\caption{Class balance in the MZVAV subset-ratio experiment, comparing uniformly subsampled and blockwise subsampling (Seed: 72).}
\label{fig:mzvav_class_balance_subset_ratio}
\end{figure}

\Cref{fig:scaling_phme20_unibo_mzvav_top5} evaluates prognostic performance on PHME20 and Unibo as the fraction of available training windows is varied under the protocol described in \cref{sec:pre-trained-in-context-learning}, using both aggregate random subsampling and blockwise subsampling. The PFN-based models are already competitive at very small data fractions, and TabPFN is particularly strong in the low-data regime.
For PHME20, TabPFN and TabDPT achieve competitive performance using only 1\% of the training data, with performance saturating once 10\% of the data is available. On the Unibo data set, TabDPT requires 10\% of the data to reach competitive performance, whereas TabPFN already performs competitively with 1\%. Both PFN-based models again show a performance plateau at approximately 10\% of the training data.
Under random subsampling, the MAE decreases smoothly as additional windows are included, before reaching a saturation regime. This suggests that a relatively small but representative support set is sufficient for strong performance. In contrast, this behavior is not observed under blockwise subsampling, where the sampled data may fail to cover the test distribution. These results indicate that tabular foundation models are highly sensitive to the in-context data distribution.
This context-sensitivity is further reflected in the main benchmark results: TabPFN's performance on NC-DS02 is notably weaker than on NC-P, where the broader multi-source training distribution provides a richer and more representative context set.
The same subset-ratio protocol is evaluated on MZVAV using Macro-F1, as shown in the bottom row of \cref{fig:scaling_phme20_unibo_mzvav_top5}. Macro-F1 generally increases with larger data fractions, but the blockwise setting exhibits larger variance compared to the random subsampling for the PFN models. This indicates that blockwise subsampling changes the class coverage of the context more strongly than aggregate subsampling, which is consistent with the class-balance analysis in \cref{fig:mzvav_class_balance_subset_ratio}.

\subsubsection{The Effect of Tabularizing Time Dependencies}
\begin{figure}[pos=htbp]
\centering
\begin{subfigure}[t]{0.48\textwidth}
\centering
\includegraphics[width=\textwidth]{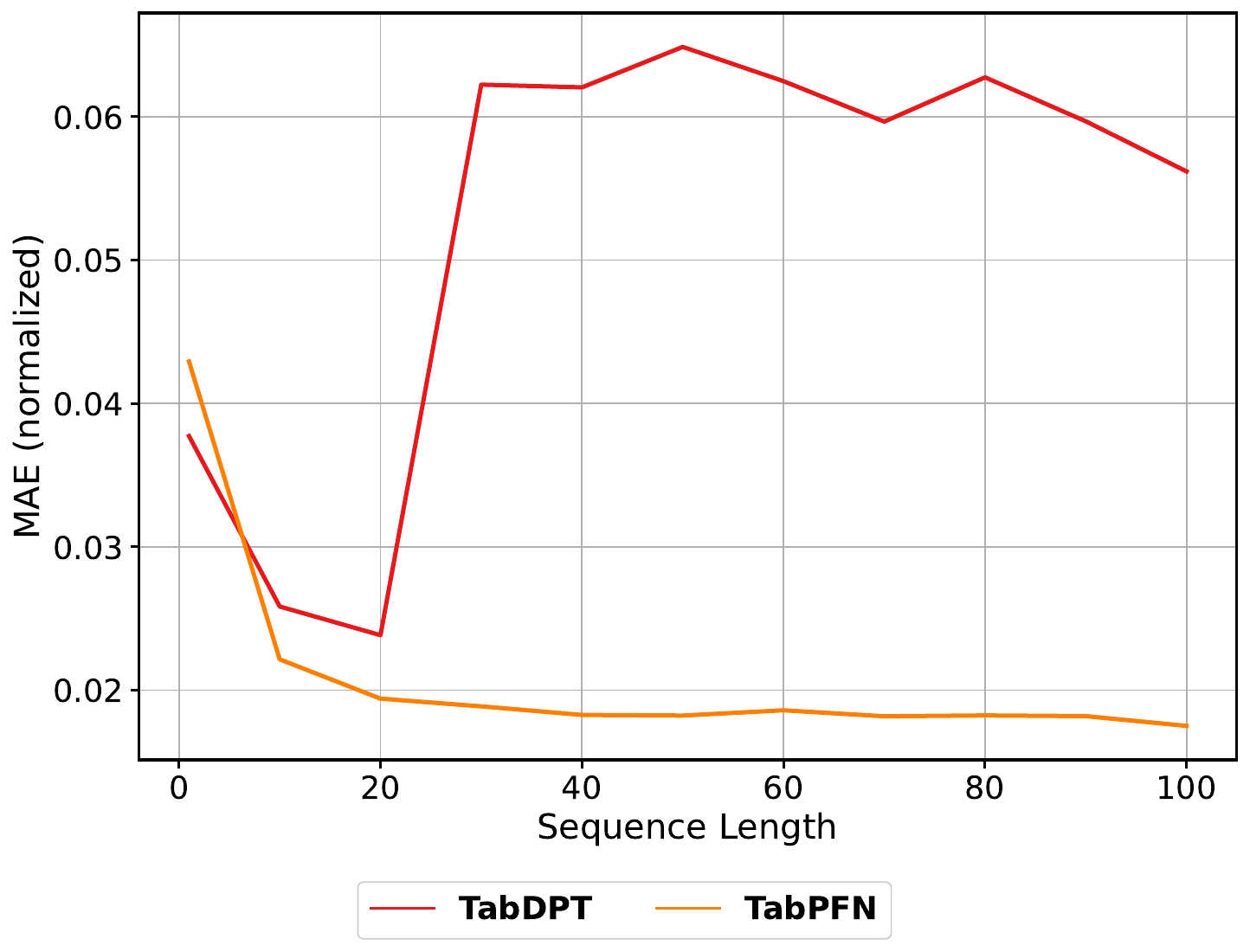}
\caption{PHME20}
\label{fig:seq_len_phme20}
\end{subfigure}
\hfill
\begin{subfigure}[t]{0.48\textwidth}
\centering
\includegraphics[width=\textwidth]{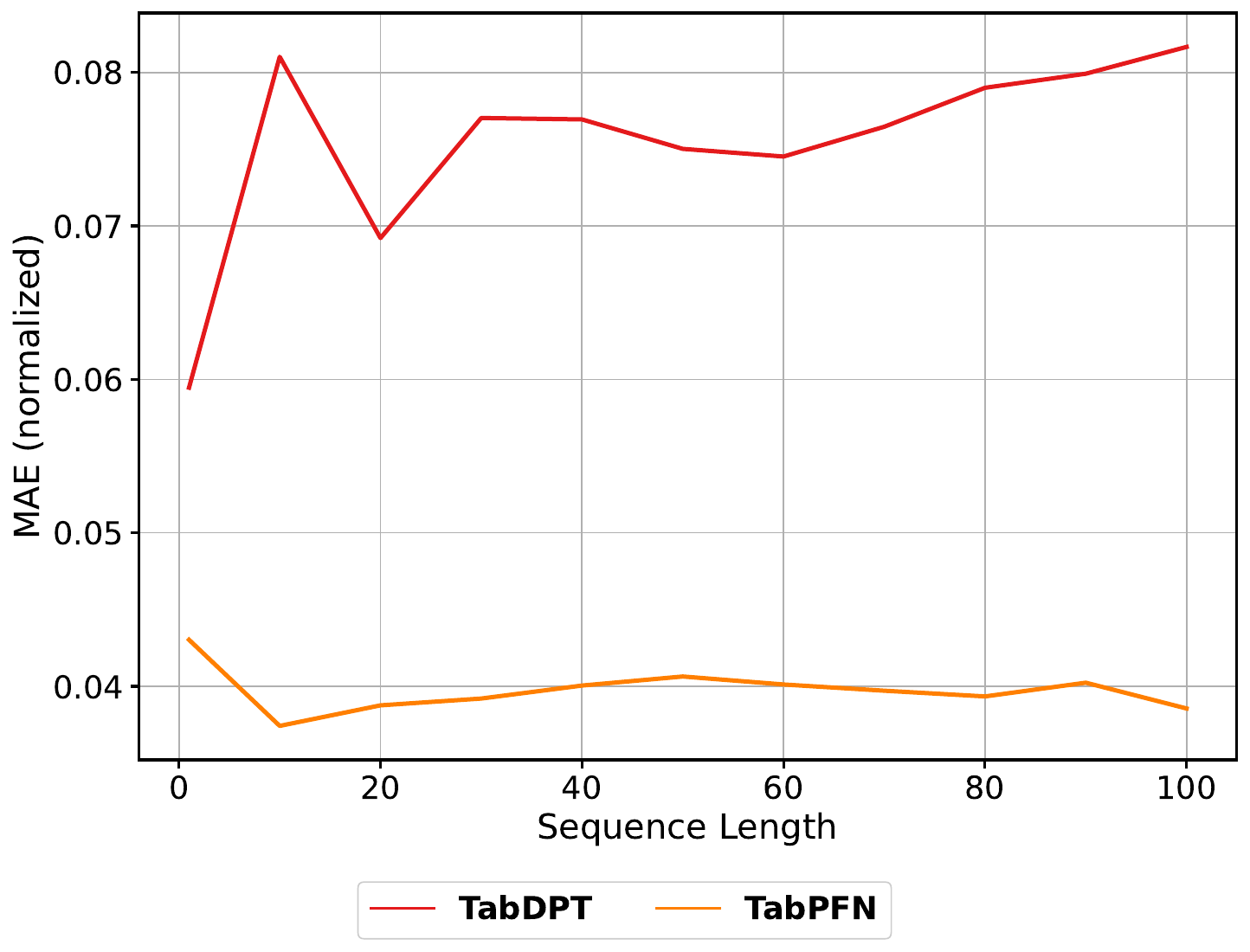}
\caption{Unibo}
\label{fig:seq_len_unibo}
\end{subfigure}

\medskip

\begin{subfigure}[t]{0.48\textwidth}
\centering
\includegraphics[width=\textwidth]{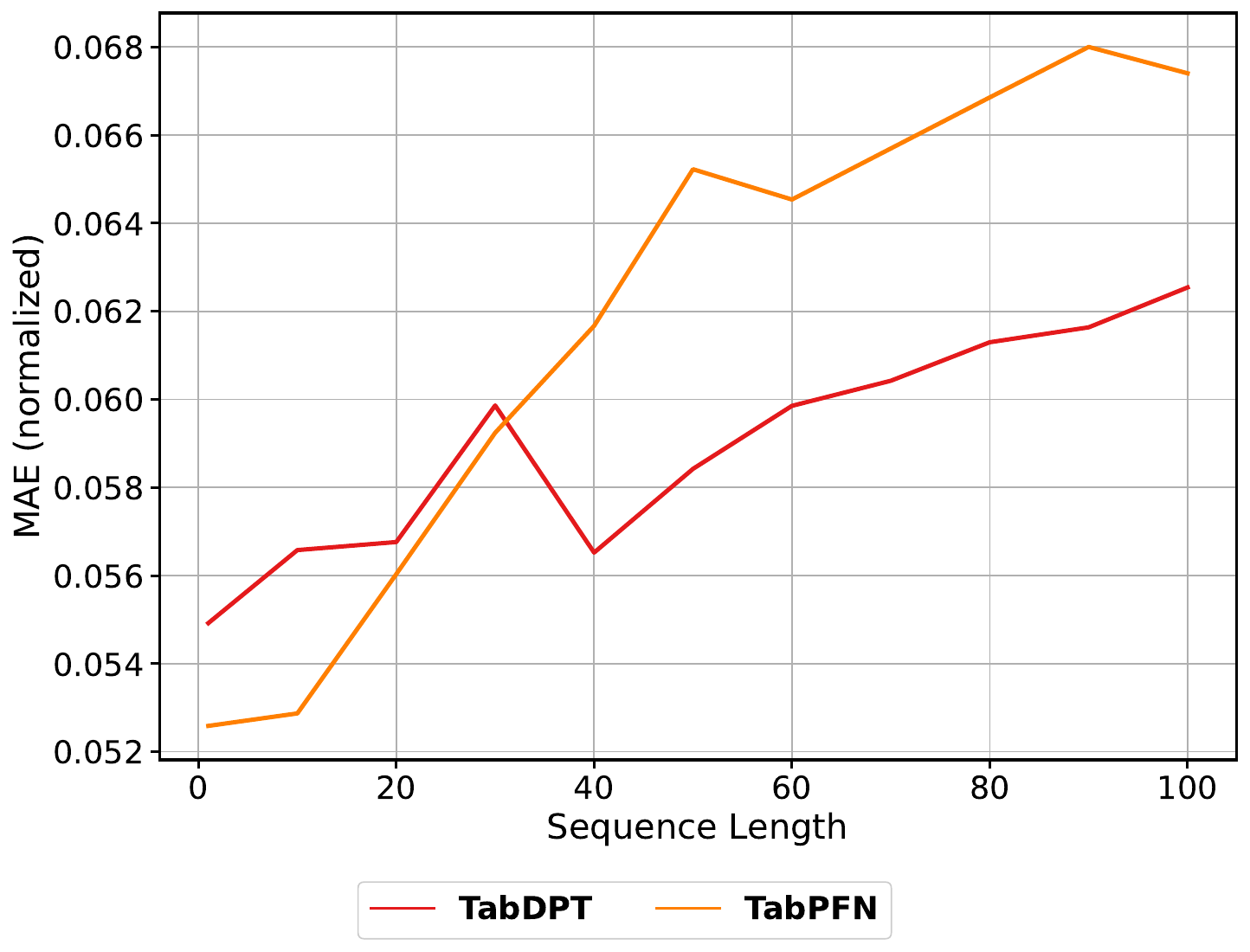}
\caption{N-CMAPSS DS02}
\label{fig:seq_len_ncmapss_ds02}
\end{subfigure}
\hfill
\begin{subfigure}[t]{0.48\textwidth}
\centering
\includegraphics[width=\textwidth]{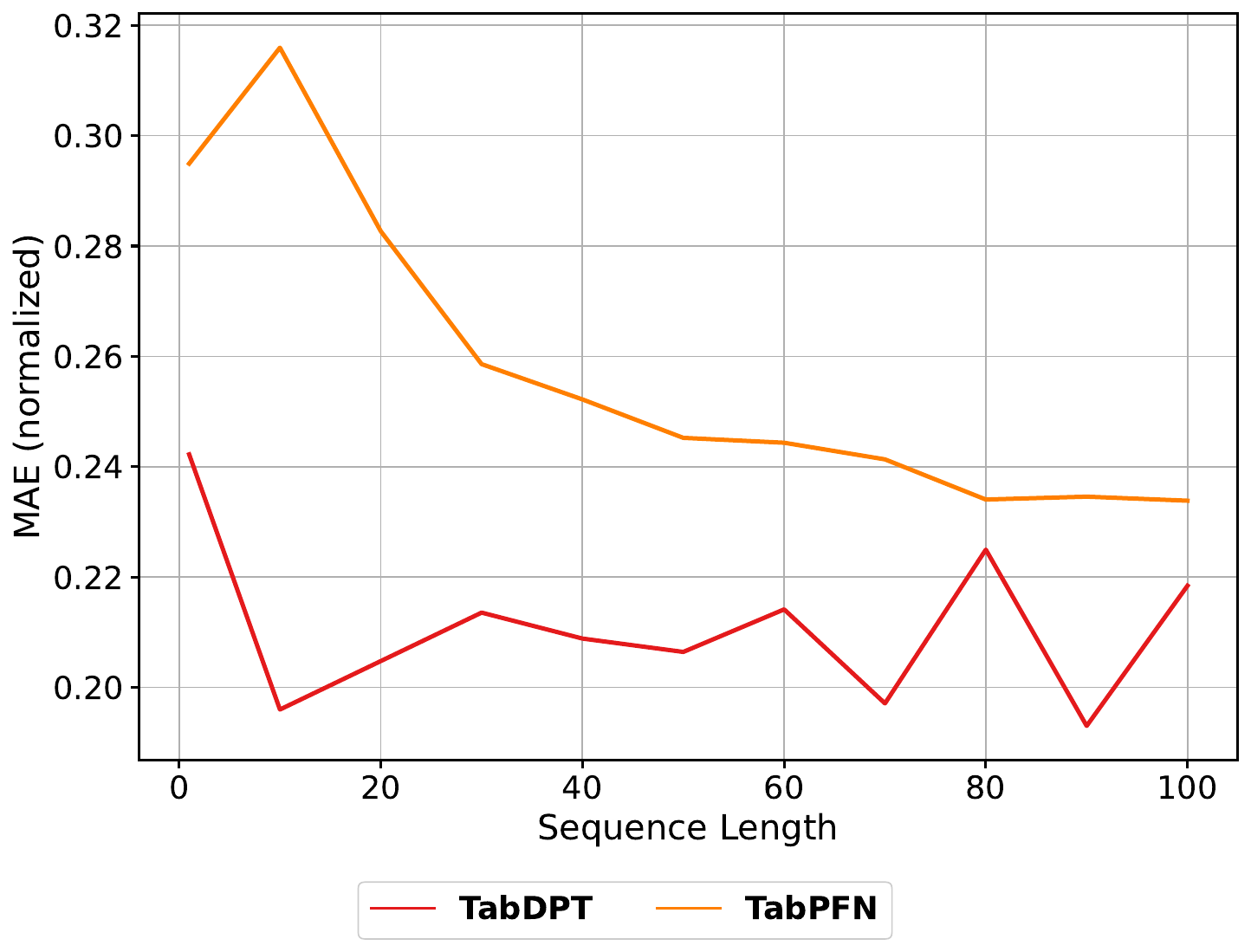}
\caption{XJTU-SY}
\label{fig:seq_len_xjtu_sy}
\end{subfigure}
\caption{Effect of sequence length on normalized MAE for selected prognostics datasets.}
\label{fig:seq_len_prognostics}
\end{figure}

Figure~\ref{fig:seq_len_prognostics} shows that increasing the sequence length improves performance on PHME20, Unibo, and XJTU-SY, confirming that the tabularized representation can preserve task-relevant temporal information. On PHME20, TabPFN improves rapidly as the sequence length increases and then saturates, whereas TabDPT deteriorates after short histories. On Unibo, TabPFN remains consistently below TabDPT across the evaluated sequence lengths, with only limited additional gains after the shortest nontrivial histories. On XJTU-SY, TabPFN benefits substantially from longer histories, while TabDPT is less monotonic but outperforms TabPFN in absolute MAE. These patterns indicate that TabPFN is better able to exploit temporal context when history carries degradation information, which is consistent with the model's cell-wise embedding of the flattened sensor-time grid, allowing TabPFN to extract meaningful causal temporal relations. TabDPT embeds complete rows and is therefore less directly aligned with temporal locality after flattening. For N-CMAPSS DS02, neither model benefits from longer histories; TabPFN becomes worse as sequence length increases, and TabDPT shows only small non-monotonic variation. 

This behavior is consistent with the structure of N-CMAPSS. Degradation is modeled primarily at the flight cycle scale, while high frequency variation within a flight is dominated by changing operating conditions, e.g. the altitude. Consequently, extending the history window over consecutive samples from the same flight may add limited degradation information. Capturing meaningful temporal degradation trends in N-CMAPSS requires context over many flight cycles rather than only adjacent samples, which is beyond the sequence lengths evaluated here.

\subsubsection{Data Imputation}

We evaluate two strategies for handling missing values. The first applies last-observation-carried-forward (LOCF) imputation to all models uniformly: missing feature entries are filled with the most recent observed value before any model receives the input. The second passes the raw NaN-valued features directly to TabPFN v2, which handles missing values internally: NaN entries are replaced by the training-partition feature mean and a per-feature binary missingness indicator is appended as extra input features. The pre-trained transformer has seen this missingness representation during training and can therefore condition its predictions on the missingness pattern. All other models receive the LOCF-imputed inputs, as they do not support NaN-valued inputs. We refer to the two TabPFN variants as \emph{TabPFN} (LOCF-imputed) and \emph{TabPFN+NaN} (TabPFN-internal handling) respectively.

\begin{figure}[pos=htbp]
\centering
\includegraphics[width=0.6\textwidth]{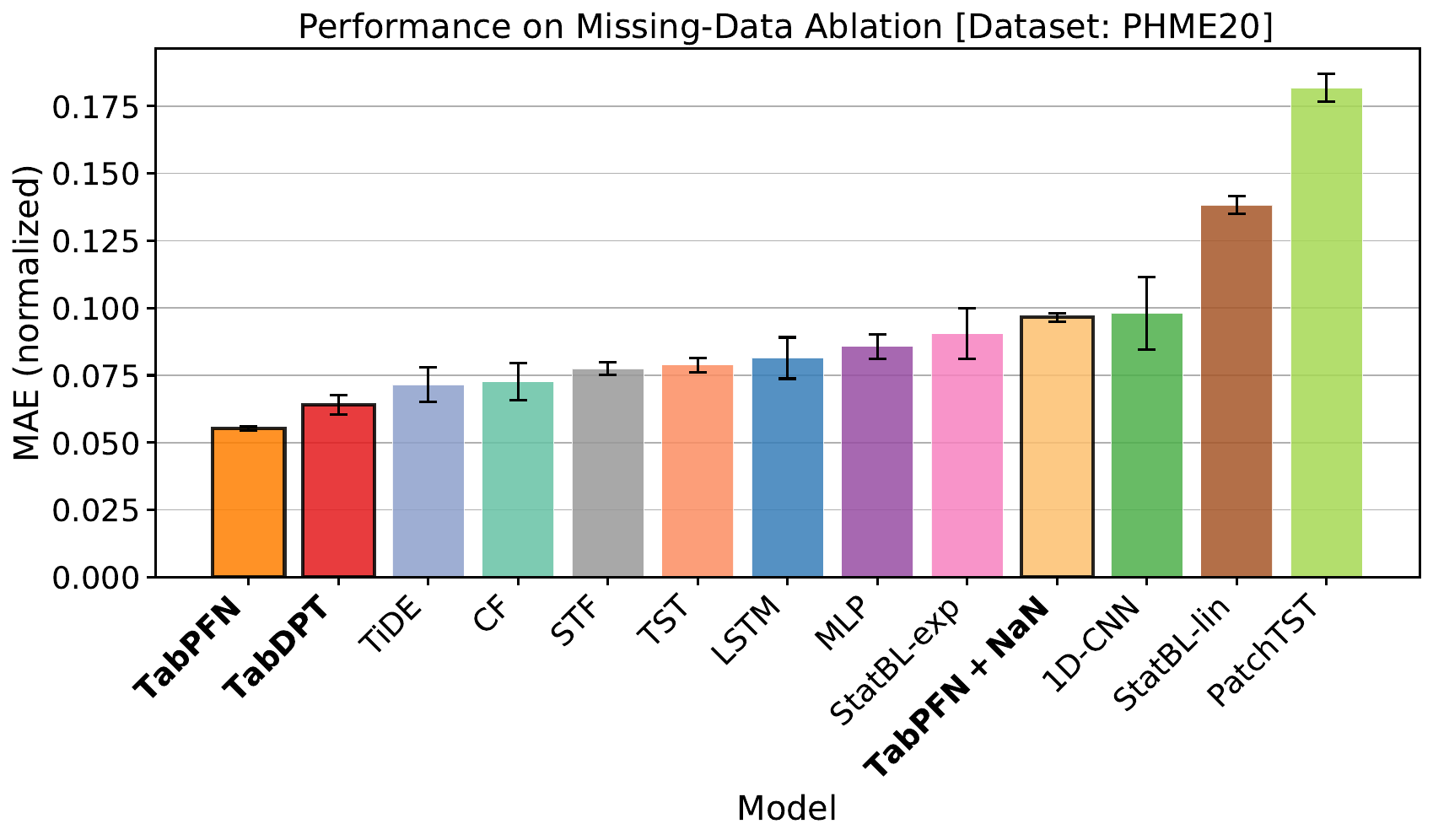}
\caption{Effect of imputation on normalized MAE for PHME20. TabPFN+Nan indicates that we use the Nan token of TabPFN to represent missing values, while TabPFN indicates that we impute missing values.}
\label{fig:missing_data_phme20}
\end{figure}

\Cref{fig:missing_data_phme20} reports the effect of missing data on normalized MAE for PHME20. TabPFN remains the strongest model under this missing-data setting, followed by TabDPT, indicating that the tabular foundation models retain their advantage when incomplete inputs are present. Unexpectedly, however, the explicit TabPFN NaN-token variant underperforms the imputed TabPFN configuration, indicating that TabPFN’s proposed missing-value handling strategy is not always effective. For this dataset and preprocessing pipeline, conventional imputation appears to provide a more informative representation than passing missingness directly to the model.

\subsubsection{Probabilistic Interpretation of the TabPFN Output}
\begin{figure}[pos=htbp]
\centering
\begin{subfigure}[t]{0.99\textwidth}
\centering
\includegraphics[width=\textwidth]{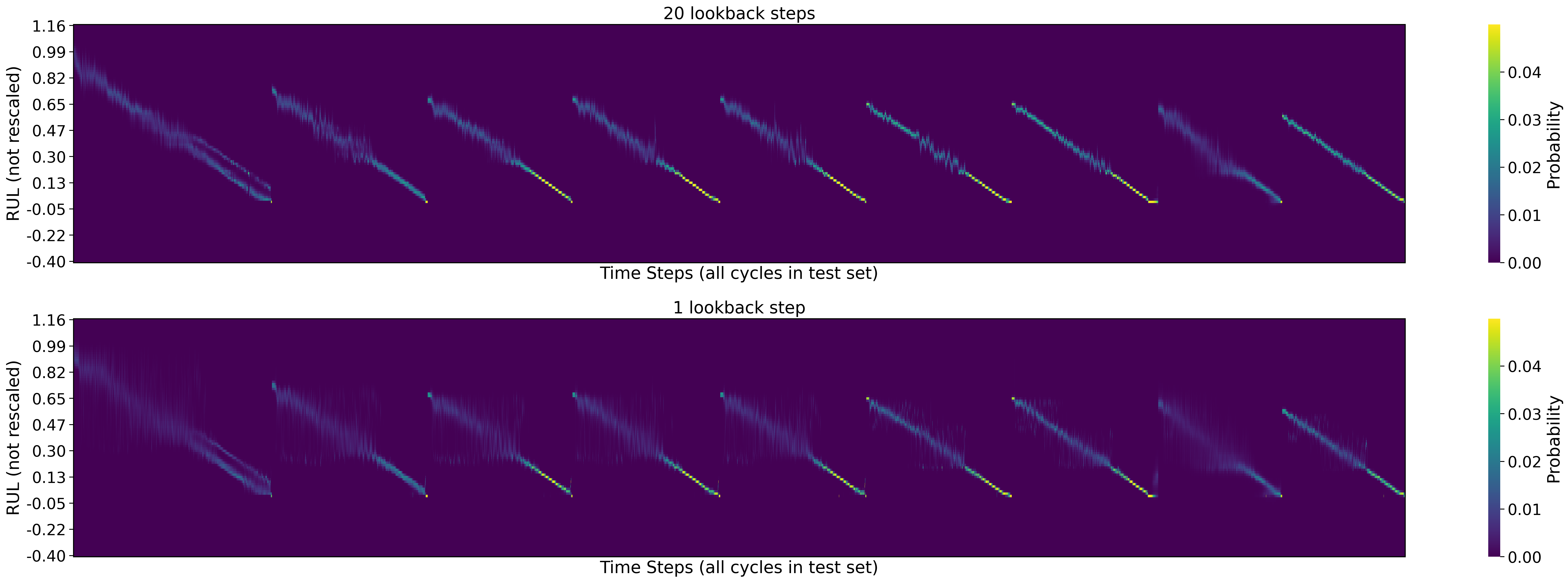}
\caption{PHME20}
\label{fig:quantile_distribution_phme20}
\end{subfigure}

\medskip

\begin{subfigure}[t]{0.99\textwidth}
\centering
\includegraphics[width=\textwidth]{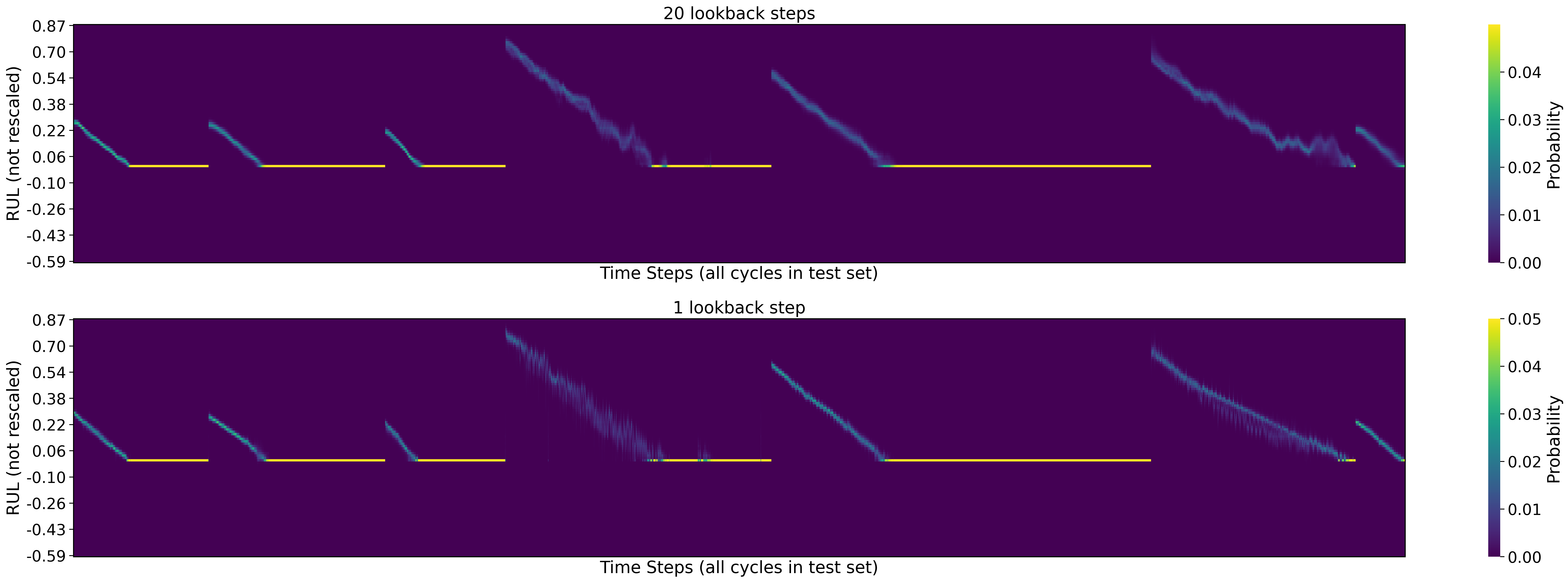}
\caption{Unibo}
\label{fig:quantile_distribution_unibo}
\end{subfigure}
\caption{Predictive quantile distribution heatmaps for PHME20 and Unibo. RUL distribution with and without tabularization of time.}
\label{fig:quantile_distribution_heatmaps}
\end{figure}

TabPFN represents continuous regression targets through a discretized predictive distribution rather than a single scalar; quantiles and point estimates are derived from this distribution.
In \cref{fig:quantile_distribution_heatmaps} we interpret this predictive distribution qualitatively, examining how probability mass concentrates across RUL values as a function of time. In both datasets, the probability mass concentrates along repeated descending trajectories, indicating that the model recovers the monotonic run-to-failure structure of individual test units. For PHME20, the 20-step lookback produces more localized probability bands, whereas the one-step setting is more diffuse with broader vertical spread, particularly in early- and mid-life regions. For Unibo, both settings show descending trajectories and pronounced probability mass near the low-RUL endpoint. Similar to PHME20, the longer lookback localizes the bands more tightly along the descending trajectory. These qualitative observations indicate that temporal context sharpens the predictive distribution, although they do not constitute a formal calibration analysis.

\FloatBarrier
\section{Conclusion \& Discussion}

Our experiments demonstrate that tabularization acts as a flexible and effective framework for PHM tasks, allowing prognostics and diagnostics to be performed under a single representation. Additionally, our experiments indicate that our tabularization scheme enables Tabular Foundation Models (TFMs) to achieve strong performance with limited hyperparameter tuning concentrated mainly on the tabular shape. The results further show that TFMs exhibit the generalization capacity and data efficiency expected of foundation models, achieving the best average ranks without task-specific training and transferring effectively to new tasks through in-context learning.

Across both prognostics and diagnostics benchmarks, TFM remain highly competitive even under severe data scarcity and in the presence of missing values — with or without explicit imputation. This combination of sample efficiency, cross-task robustness, and few-shot generalization makes them attractive for industrial settings where labeled data is scarce and failure modes are heterogeneous.

This performance, however, depends critically on the distributional richness of the in-context table. Because these models learn from the provided context at inference time, the quality and coverage of that context directly determines prediction quality. Therefore, a context that does not cover the operating regimes, degradation stages, or fault modes present in the test data will result in degraded predictions. The N-CMAPSS families provide direct evidence of this: NC-DS02 is restricted to a single N-CMAPSS dataset file, with a limited set of units and failure-mode configurations, whereas NC-P pools multiple N-CMAPSS sources and fault configurations, yielding a richer and more varied training distribution. Notably, TabDPT achieves the best normalized MAE on NC-P among all evaluated models, underlining that tabular foundation models can excel when context diversity is sufficient. This distributional perspective is also consistent with the sequence-length analysis on NC-DS02, where additional temporal context provides only minimal benefit. Once the inference-time distribution is sufficiently covered, adding more samples or variables appears to provide little additional information for the model’s posterior. Taken together with the data-efficiency results, this finding is practically encouraging: because tabular foundation models can exploit small but well-chosen context sets, providing even a modest number of targeted samples that cover rare operating regimes or fault modes may be sufficient to unlock strong generalization in those settings.

More generally, PHM performance is strongly influenced by the preprocessing pipeline, making the choice of input transformations a central component of the evaluation protocol. The transformations selected in our experiments span both time and frequency domains to provide a comprehensive representation of the input signal. Since the primary objective of this study is cross-task robustness and sample efficiency rather than exhaustive per-dataset optimization, the shared transform set is intentionally conservative. This ensures fair comparison across model families rather than maximizing performance on any individual benchmark. Identifying more general transformation sets that transfer robustly across PHM datasets, architectures, and operating regimes remains an important direction for future work.

TFM’s computational efficiency is governed by the number of in-context samples and by the dimensionality of each tabularized row. In the present experiments, subsampling is applied only at the window level, and selected context rows retain the full flattened sensor-time representation. The finite inference-time budget therefore constrains the number of context samples that can be used, which can limit attainable accuracy. The current tabularization remains restrictive, as it does not allow individual sensors or time steps to be subsampled within a window. Extending the approach with per-variate or per-timestep subsampling could reduce $D_{\mathrm{tab}}$, alleviate the computational bottleneck, and preserve more context samples by retaining only the most informative features.
Future work may also develop more advanced context-selection strategies that preserve operating-regime, degradation-stage, and class coverage while remaining within the computational limits of TFM inference.

A significant advantage of PFN-based models is the dual utility of the validation set. In traditional deep learning, the validation set is strictly partitioned for hyperparameter tuning and early stopping to mitigate overfitting, effectively isolating it from the final inference phase. In contrast, PFNs allow the validation data to be repurposed. Once the optimal configuration (e.g., tabularization parameters) is identified, the validation set can be integrated into the model's context. This enables the model to leverage a richer set of labeled points during inference on the test set, maximizing the informative value of all available data. A systematic study of how validation-set reuse interacts with context size, distributional shift, and tabularization choices remains an open and practically relevant research direction.

Finally, scalable PHM modeling may not be limited to replacing conventional methods with foundation models. Recent agentic PHM systems and benchmarks provide a complementary direction by automating parts of model configuration, tool orchestration, method reproduction, and evaluation under executable interfaces \citep{cha2025llmagentphm,feng2026phmforge,theiler2026paperbenchmark}. Such systems may make conventional methods more viable at scale by reducing the manual effort required to translate paper-specific assumptions or user-defined PHM tasks into standardized preprocessing, model, target, and evaluation configurations.

\appendix

\newcommand{\appendixinput}[1]{\input{#1}}
\newcommand{\reviewheading}[1]{#1}
\newcommand{\reviewparagraph}[1]{\par\smallskip\noindent\textbf{#1}\ }

\clearpage
\appendix
\section*{\reviewheading{Appendix}}
\phantomsection
\label{app:appendix}

\subsection*{\reviewheading{Appendix Contents}}
\label{app:contents}
\begingroup
\makeatletter
\setlength{\parindent}{0pt}
\setlength{\parskip}{0pt}
\footnotesize
\hypersetup{linkcolor=black}
\newcommand{\appendixcontentsection}[2]{%
  \@dottedtocline{1}{0em}{2.4em}{\hyperref[#1]{\textbf{#2}}}{\pageref{#1}}%
}
\newcommand{\appendixcontentsubsection}[2]{%
  \@dottedtocline{2}{2.4em}{2.4em}{\hyperref[#1]{#2}}{\pageref{#1}}%
}
\appendixcontentsection{app:dataset_protocols}{A.\hspace{0.4em}Dataset Protocols}
\appendixcontentsubsection{app:battery_datasets}{Battery datasets}
\appendixcontentsubsection{app:bearing_datasets}{Bearing datasets}
\appendixcontentsubsection{app:ncmapss_mzvav}{N-CMAPSS DS02 and multi-source families}
\appendixcontentsubsection{app:hsf15}{Hydraulic diagnostics (HSF15)}
\appendixcontentsubsection{app:phme20}{PHME20}
\appendixcontentsubsection{app:mzvav}{MZVAV}
\vspace{0.2\baselineskip}
\appendixcontentsection{app:model_details}{B.\hspace{0.4em}Implementation and Model Details}
\appendixcontentsubsection{app:model_sequence_icltab}{Sequential models and in-context tabular models}
\appendixcontentsubsection{app:model_eval_baseline}{Evaluation protocol and baseline adaptation}
\appendixcontentsubsection{app:model_tabpfn_tabdpt}{TabPFN and TabDPT details}
\vspace{0.2\baselineskip}
\appendixcontentsection{app:additional_results}{C.\hspace{0.4em}Additional Experimental Results}
\appendixcontentsubsection{app:exp_setup_full}{Experimental setup}
\appendixcontentsubsection{app:transform_schemas}{Transformation schemas}
\appendixcontentsubsection{app:hyperparameter_search}{Hyperparameter search}
\appendixcontentsubsection{app:reading_guide}{Reading the result tables}
\appendixcontentsubsection{app:full_diagnostics_results}{Diagnostics results}
\appendixcontentsubsection{app:full_prognostics_results}{Prognostics results}
\appendixcontentsubsection{app:data_efficiency_scaling}{Data-efficiency scaling}
\vspace{0.2\baselineskip}
\appendixcontentsection{app:reproducibility}{D.\hspace{0.4em}Reproducibility}
\appendixcontentsubsection{app:reproducibility_framework}{Framework-based execution}
\appendixcontentsubsection{app:reproducibility_configs}{Configuration-fixed experiments}
\appendixcontentsubsection{app:reproducibility_boundaries}{Shared preprocessing and evaluation boundaries}
\vspace{0.2\baselineskip}
\appendixcontentsection{app:data_code_access}{E.\hspace{0.4em}Data and Code Access}
\appendixcontentsubsection{app:code_access}{Code and protocol availability}
\appendixcontentsubsection{app:dataset_access}{Third-party datasets}
\hypersetup{linkcolor=blue}
\makeatother
\endgroup

\section{\reviewheading{Dataset Protocols}}
\label{app:dataset_protocols}
\label{sec:datasets}

This section records the dataset contract behind the main benchmark table. For each evaluated family, it states the source data, split logic, target definition, normalization, and metrics needed to interpret the reported prognostics or diagnostics results.

\subsection{\reviewheading{Battery Datasets}}
\label{app:battery_datasets}

The battery prognostics tasks use NB14 and UNIBO21. Both datasets are treated as run-to-failure battery degradation problems with an operation-dependent RUL target based on remaining cumulative discharge.

\reviewparagraph{NB14.}
\textbf{Description.} The NASA Randomized Battery Usage dataset \citep{bole2014adaptation} consists of 28 lithium cobalt oxide 18,650 cells organized into 7 operational groups. Battery aging is induced through repeated charge and randomized discharge cycles with loading periods of 5 minute. After 1500 loading periods, reference characterization cycles are used to assess battery health in terms of capacity.
The signals are recorded at 0.1 Hz and comprise voltage, current, and temperature recorded during the aging tests.

\textbf{Split.} Following \citet{bosello2023charge}, Group 3 is excluded, and cells RW3 and RW20 are removed because of corrupted or unusable measurements. The remaining 22 cells are split into training, validation, and test units; the exact split is shown in \cref{tab:nasa_split}.
\begin{itemize}
    \item \textbf{RW3} is excluded because of corrupted temperature measurements.
    \item \textbf{RW20} is excluded because its sensor data report zero for nearly its entire life.
\end{itemize}
\begin{table}[h!]
\centering
\caption{Train/Validation/Test Split of the NASA Randomized Battery Dataset.}
\label{tab:nasa_split}
\begin{adjustbox}{width=0.8\textwidth}
\begin{tabular}{@{}lllll@{}}
\toprule
\textbf{Group} & \textbf{Original Cells} & \textbf{Training Set} & \textbf{Validation Set} & \textbf{Test Set} \\ 
\midrule
Group 1 & RW1, RW2, RW7, RW8       & RW1, RW2                 & RW7  & RW8 \\
Group 2 & RW3, RW4, RW5, RW6       & RW4 \textit{(RW3 excl.)} & RW5  & RW6 \\
Group 3 & RW9, RW10, RW11, RW12    & \multicolumn{3}{c}{\textit{Excluded (unrealistic profile)}} \\
Group 4 & RW13, RW14, RW15, RW16   & RW13, RW14               & RW15 & RW16 \\
Group 5 & RW17, RW18, RW19, RW20   & RW17 \textit{(RW20 excl.)} & RW18 & RW19 \\
Group 6 & RW21, RW22, RW23, RW24   & RW21, RW22               & RW23 & RW24 \\
Group 7 & RW25, RW26, RW27, RW28   & RW25, RW26               & RW27 & RW28 \\ 
\midrule
\textbf{Total} & \textbf{28 Batteries} & \textbf{10 Batteries} & \textbf{6 Batteries} & \textbf{6 Batteries} \\ 
\bottomrule
\end{tabular}
\end{adjustbox}
\end{table}

\textbf{Target, normalization, and metrics.} NB14 uses the shared battery ah-RUL target defined below. Input features are min-max normalized, and performance is reported with MAE, MSE, and RMSE on $Q_{RUL}$.

\reviewparagraph{UNIBO21.}
\textbf{Description.} The UNIBO Powertools dataset \citep{10.1145/3462203.3475878} contains laboratory data from 30 lithium-ion batteries used to study batteries in cleaning-equipment conditions. The cells differ by manufacturer, type, capacity, and test regime and can be categorized in 7 groups.The experiment protocol comprises a charging with 1.8 A current to a cutoff-voltage of 4.2 V, and discharging with a 5 A current until a defined end-of-discharge voltage. The health indicators capacity and resistance were measured by performing 100 reference cycles. The collected signals are sampled with 0.1 Hz and include temperature, voltage, current, and energy measured during charging as well as discharging.

\textbf{Split.} Following \citet{bosello2023charge}, cells 019, 047, and 049 are excluded because of data corruption or incomplete end-of-life cycling. The remaining 27 cells are split into training, validation, and test units, with the exact split shown in \cref{tab:unibo_split}.
\begin{enumerate}
    \item \textbf{019}: corrupted data.
    \item \textbf{047, 049}: not cycled to end-of-life at the time of dataset construction.
\end{enumerate}
\begin{table}[h!]
\centering
\caption{Train/Validation/Test Split of the UNIBO21 Dataset.}
\label{tab:unibo_split}
\begin{adjustbox}{width=0.8\textwidth}
\begin{tabular}{@{}lllll@{}}
\toprule
\textbf{Group} & \textbf{Original Cells} & \textbf{Training Set} & \textbf{Validation Set} & \textbf{Test Set} \\ 
\midrule
DM-3.0-S  & 000, 001, 002, 003            & 000, 001                & 002                  & 003 \\
DM-3.0-H  & 009, 010, 011                 & 009                     & 010                  & 011 \\
DM-3.0-P  & 013-017, (047, 049)           & 014, 015, 017           & 016                  & 013 \\
EE-2.85-S & 006, 007, 008, 042            & 007, 008                & 042                  & 006 \\
EE-2.85-H & 043, 044                      & 043                     & ---                  & 044 \\
DP-2.00-S & 018, 036-039, 050, 051, (019) & 018, 036, 037, 038, 050 & 051                  & 039 \\
DM-4.00-S & 040, 041                      & 040                     & ---                  & 041 \\ 
\midrule
\textbf{Total} & \textbf{27 (+3 excl.)} & \textbf{15 Batteries} & \textbf{5 Batteries} & \textbf{7 Batteries} \\ 
\bottomrule
\end{tabular}
\end{adjustbox}
\end{table}

\textbf{Target, normalization, and metrics.} UNIBO21 uses the same ah-RUL formulation as NB14, with dataset-specific implementation details for end-of-life and current integration. Input features are min-max normalized, and performance is reported with MAE, MSE, and RMSE on $Q_{RUL}$.

\reviewparagraph{Target creation and metrics.}
Following \citet{bosello2023charge}, both battery datasets use normalized remaining cumulative discharge, or ah-RUL, instead of a purely time- or cycle-based target. For any cycle $n$, ah-RUL is computed as:

\begin{equation}
Q_{RUL}(n) = Q_{acc}(n_{EoL}) - Q_{acc}(n)
\end{equation}
where $Q_{RUL}(n)=0$ for cycles $n \geq n_{EoL}$. The cumulative discharge throughput is computed by integrating discharge current and normalizing by nominal capacity:
\begin{equation}
Q_{acc}(n) = \frac{1}{Q_{nom}} \sum_{i=1}^{n} \left( \int_{t_i}^{t_{i+1}} I_d(t) dt \right)
\end{equation}

The NB14 and UNIBO21 implementations differ in the definition of $n_{EoL}$ and in how discharge current is selected, so the dataset-specific target rules are stated together with the split definitions.

\subsection{\reviewheading{Bearing Datasets}}
\label{app:bearing_datasets}

The evaluated bearing prognostics task is XJTU-SY. It provides high-frequency vibration measurements from accelerated run-to-failure bearing experiments and is evaluated as a health-indicator regression task.

\reviewparagraph{XJTU-SY.}
\textbf{Description.} XJTU-SY, introduced by \citet{yaguo2019xjtu}, provides 15 run-to-failure bearings under three operating conditions. Each bearing is recorded through horizontal and vertical high-frequency vibration channels until failure with a sampling rate of 25.6 kHz. In this benchmark, each acquisition is represented as a two-channel vibration segment of shape $(32768,2)$, and the operating conditions and bearing lifetimes used for target construction are summarized below.

\textbf{Preprocessing.} The preprocessing pipeline standard-scales the raw vibration channels, computes a cumulative-sum feature, extracts time-domain and spectral descriptors, and min-max rescales the concatenated descriptor block. Fit-predict models reuse the same processed representation through a tabularized variant.

\textbf{Split.} The reported XJTU-SY experiments use the configured PHMD split. The training bearings are \texttt{1\_3}, \texttt{1\_4}, \texttt{2\_1}, \texttt{2\_4}, \texttt{2\_5}, \texttt{3\_1}, \texttt{3\_2}, and \texttt{3\_3}; the validation bearings are \texttt{1\_1}, \texttt{1\_2}, and \texttt{3\_5}; and the test bearings are \texttt{1\_5}, \texttt{2\_2}, \texttt{2\_3}, and \texttt{3\_4}.

\begin{table}[h!]
\centering
\caption{XJTU-SY Operating Conditions.}
\label{tab:xjtu_conditions}
\begin{adjustbox}{width=0.5\textwidth}
\begin{tabular}{@{}lcc@{}}
\toprule
\textbf{Condition ID} & \textbf{Engine Speed (rpm)} & \textbf{Radial Load (N)} \\
\midrule
Condition 1 & 2100 & 12,000 \\
Condition 2 & 2250 & 11,000 \\
Condition 3 & 2400 & 10,000 \\
\bottomrule
\end{tabular}
\end{adjustbox}
\end{table}
\begin{table}[h!]
\centering
\caption{XJTU-SY bearing lifetimes (min), used for regression targets and HI normalization.}
\label{tab:xjtu_lifetimes}
\begin{adjustbox}{width=0.5\textwidth}
\begin{tabular}{@{}llc@{}}
\toprule
\textbf{Condition} & \textbf{Bearing Name} & \textbf{Bearing Lifetime (min)} \\ 
\midrule
Condition 1 & Bearing 1\_1 & 123 \\
Condition 1 & Bearing 1\_2 & 161 \\
Condition 1 & Bearing 1\_3 & 158 \\
Condition 1 & Bearing 1\_4 & 122 \\
Condition 1 & Bearing 1\_5 & 52 \\ 
\midrule
Condition 2 & Bearing 2\_1 & 491 \\
Condition 2 & Bearing 2\_2 & 161 \\
Condition 2 & Bearing 2\_3 & 533 \\
Condition 2 & Bearing 2\_4 & 42 \\
Condition 2 & Bearing 2\_5 & 339 \\ 
\midrule
Condition 3 & Bearing 3\_1 & 2538 \\
Condition 3 & Bearing 3\_2 & 2496 \\
Condition 3 & Bearing 3\_3 & 371 \\
Condition 3 & Bearing 3\_4 & 1515 \\
Condition 3 & Bearing 3\_5 & 114 \\ 
\bottomrule
\end{tabular}
\end{adjustbox}
\end{table}

\textbf{Target, normalization, and metrics.} Runtime is converted into the HI target below using the bearing lifetime table. XJTU-SY is evaluated with standard regression metrics such as MAE and RMSE between the true HI and the predicted HI for each sample.

\reviewparagraph{Target creation and metrics.}
For XJTU-SY, the target variable is formulated as a normalized health indicator (HI), linearly decreasing from 1 for a healthy bearing to 0 at failure:
\begin{equation}
    HI = 1 - \frac{\text{Runtime}}{\text{Total Lifetime}}
\end{equation}
For XJTU-SY, \texttt{Total Lifetime} is the \texttt{Bearing Lifetime (min)} in \cref{tab:xjtu_lifetimes}.

\subsection{\reviewheading{N-CMAPSS DS02 and Multi-Source Families}}
\label{app:ncmapss_mzvav}

\reviewparagraph{N-CMAPSS.}
\textbf{Description.} N-CMAPSS \citep{arias2021aircraft,frederick2007user} is a NASA benchmark family derived from C-MAPSS turbofan engine simulations. It provides multivariate time series with run-to-failure trajectories, operating conditions, fault modes, and ground-truth RUL. This work uses this source family in two scopes: NC-DS02 denotes the DS02 prognostics protocol, while NC-P and NC-D denote the broader N-CMAPSS prognostics and diagnostics settings reported in the main results.

\textbf{Split.} The explicit split table below documents the DS02 protocol because it is the N-CMAPSS split currently tabulated in the manuscript. DS02 is split into five training units, one validation unit, and three test units, with the split shown in \cref{tab:cmapss_split}. The broader N-CMAPSS settings remain part of the same source family and are reported separately in the main results as NC-P and NC-D.
\begin{table}[htbp]
\centering
\caption{\protect\textcolor{black}{N-CMAPSS-02 data splits, flight classes, and fault types (notation aligned with the main text).}}
\label{tab:cmapss_split}
\begin{tabular}{@{}llc@{}}
\toprule
\textbf{Split} & \textbf{Unit number} & \textbf{Flight class} \\ 
\midrule
Train & 2 & 2 \\
Train & 5 & 2 \\
Train & 10 & 2 \\
Train & 16 & 2 \\
Train & 20 & 2 \\
\midrule
Validation & 18 & 2 \\
\midrule
Test & 11 & 2 \\
Test & 14 & 2 \\
Test & 15 & 2 \\
\bottomrule
\end{tabular}
\end{table}

\textbf{Target, normalization, and metrics.} For NC-DS02 and NC-P, the target is RUL, measured as the remaining flights or cycles before end-of-life. Features are min-max normalized, and prognostics metrics are MAE, MSE, and RMSE of estimated RUL against the true value. For NC-D, the target is a diagnostic class label and the main-results metric is F1 score.

\subsection{\reviewheading{Hydraulic Diagnostics (HSF15)}}
\label{app:hsf15}

\reviewparagraph{HSF15.}
\textbf{Description.} HSF15 \citep{hsf15_helwig} is a hydraulic-system condition-monitoring benchmark based on a laboratory test rig with multivariate sensor measurements. This work evaluates it as four component-level diagnostics tasks: accumulator (HSF15-A), cooler (HSF15-C), pump (HSF15-P), and valve (HSF15-V).

\textbf{Split.} All four HSF15 tasks use the same datasource family and split protocol, but differ in the target component and number of fault classes. The tasks are reported separately in the main benchmark table because each component defines a distinct diagnostic classification problem.

\textbf{Target, normalization, and metrics.} The target is the component-specific fault class. Features are scaled according to the shared preprocessing pipeline, and the tabular fit-predict pipeline summarizes sensor bursts through time-domain and spectral descriptors before tabularization. The main-results metric is F1 score.

\subsection{\reviewheading{PHME20}}
\label{app:phme20}

\reviewparagraph{PHME20.}
\textbf{Description.} PHME20 \citep{PHME20-GTU} is the PHM Society 2020 European Conference Data Challenge dataset. It records an experimental industrial filtration system in which a particulate-laden gas stream progressively clogs a filter, increasing differential pressure until an operational threshold is reached. Each run captures one filter lifetime under varying dust and feed conditions.

\textbf{Split.} This work follows the challenge-provided split, which assigns disjoint filter runs to training, validation, and test partitions. No additional dataset-specific filtering is applied in this protocol.

\textbf{Target, normalization, and metrics.} PHME20 is used as a prognostics task with a direct per-timestep RUL target. The preprocessing pipeline min-max scales the sensor channels and RUL target. The task logs MAE, MSE, RMSE, and \texttt{nasa\_score}.

\subsection{\reviewheading{MZVAV}}
\label{app:mzvav}

\reviewparagraph{MZVAV.}
\textbf{Description.} MZVAV is an automated fault detection and diagnostics dataset stemming from a small commerical building with multi-zone variable air volume system generated by Drexel university in the ASHRAE 1312 project. \citep{granderson_building_2020}. It is a simulated building-fault dataset featuring three air-handling units with 18 sensors collected during 26 days across summer, winter, and transition seasons with a one-minute resolution. The operation faults were manually imposed into the control system. 
The collected signals include outdoor air temperature,
supply air temperature and set-point, mixed air and return air temperature, supply air fan status and speed control, return air fan status and speed control, exhaust air damper control, outdoor air and return air damper control, cooling and heating coil valve control, supply air duct static pressure and set point,	occupancy mode indicator and fault detection ground truth. 
On this basis, we formulate a multi-class classification problem with the fault detection ground truth as label.

\textbf{Split.} Individual faults are grouped into the 4 classes "Outdoor Air Damper Stuck", "Heating Coil Valve Leaking", "Cooling Coil Valve Leaking", and "Unfaulted". The number of days per group is shown in \cref{tab:mzvav_fault_groups}. The benchmark uses a stratified train/validation/test split over days with a test size of 20\%.

\begin{table}[h!]
\centering
\begin{tabular}{@{}lc@{}}
\toprule
\textbf{Fault Category} & \textbf{Fault days} \\ \midrule
Outdoor Air Damper Stuck & 5 \\
Heating Coil Valve Leaking & 3 \\
Cooling Coil Valve Leaking & 5 \\
Unfaulted & 13 \\ \bottomrule
\end{tabular}
\caption{Number of daulty days per fault group}
\label{tab:mzvav_fault_groups}
\end{table}

\textbf{Target, normalization, and metrics.} The target is the fault detection ground truth. Features are min-max scaled to $[0,1]$. Metrics are F1-score, precision, recall, and accuracy.

\section{\reviewheading{Implementation and Model Details}}
\label{app:model_details}

\subsection{\reviewheading{Sequential Models and In-Context Tabular Models}}
\label{app:model_sequence_icltab}

This subsection collects the implementation details needed to connect the protocol above to the evaluated model families. The key distinction is the input representation: sequence models consume $\mathcal{D}_{\mathrm{seq}}$, while tabular foundation models and tabular baselines consume $\mathcal{D}_{\mathrm{tab}}$.

Conventional sequence models, such as 1D-CNNs, LSTMs, and Time-Series Transformers, are trained to minimize a task-specific loss over the training sequence dataset $\mathcal{D}_{\mathrm{seq}}^{\text{train}}$. These models do not operate on tabularized rows $\vX_m$; instead, they ingest the temporal window $\vW_m$ directly.

Let $f_\theta$ denote a model parameterized by $\theta$. The model produces a prediction $\hat{y}_m = f_\theta(\vW_m)$. The optimal parameters $\theta^*$ are obtained by minimizing empirical risk over the training sequence dataset:

\begin{equation}
    \theta^* = \operatorname*{arg\,min}_{\theta} \frac{1}{|\mathcal{D}_{\mathrm{seq}}^{\text{train}}|} \sum_{(\vW_m, y_m) \in\mathcal{D}_{\mathrm{seq}}^{\text{train}}} \ell\left(f_\theta(\vW_m), y_m\right)
\end{equation}

where $\ell(\cdot, \cdot)$ is a task-dependent loss function, such as mean squared error for prognostics or cross-entropy for diagnostics. We typically employ stochastic gradient descent variants such as AdamW to solve this optimization problem.

Tabular foundation models operate through in-context learning. In this paradigm, the model $f_\phi$, parameterized by weights $\phi$, predicts for a query row $\vX_q$ by conditioning on a context set $\mathcal{C}$ of labeled support examples:

\begin{equation}
    \hat{y}_q = f_\phi(\mathcal{C}, \vX_q)
\end{equation}

To ensure a valid evaluation and prevent leakage, the context set must be drawn strictly from the training partition of the tabular dataset:

\begin{equation}
    \mathcal{C} = \{(\vX_{j}, y_{j})\}_{j=1}^{C} \subset \mathcal{D}_{\mathrm{tab}}^{\text{train}}.
\end{equation}

XGBoost is included as a classical tree-based tabular baseline \citep{chen2016xgboost}. It uses the same tabularized rows $\vX_m$ as the tabular foundation models, but learns task-specific boosted trees from the training partition rather than using in-context prediction.

\subsection{\reviewheading{Evaluation Protocol and Baseline Adaptation}}
\label{app:model_eval_baseline}

All models are evaluated on the same held-out instances derived from $\mathcal{D}^{\text{test}}$. The only difference is the representation supplied to the model: sequence models consume $\vW_m$, while tabular models consume $\vX_m$ and, when applicable, a training-only context set $\mathcal{C}$.

For a task-specific loss $\ell(\hat{y}, y)$, the aggregate test loss is computed over the same test index set for both model families:

\begin{equation}
    \mathcal{L}_{\mathrm{test}} = \frac{1}{|\mathcal{D}^{\text{test}}|} \sum_{m \in \mathcal{D}^{\text{test}}} \ell(\hat{y}_m, y_m), \quad \text{where } \hat{y}_m =
    \begin{cases}
        f_\theta(\vW_m) & \text{if } f \text{ is a sequence model}, \\
        f_\phi(\mathcal{C}, \vX_m) & \text{if } f \text{ is an in-context tabular model}.
    \end{cases}
\end{equation}

For prognostics, this work reports regression metrics such as RMSE and MAE. For diagnostics, it reports classification metrics such as accuracy and macro-F1. For cross-dataset comparisons, models are ranked within each dataset and the average rank is reported, with rank 1 denoting the best model for a given dataset.

Baseline models trained from scratch, namely Bi-LSTM, 1D-CNN, TiDE, and Transformer-based models, are applied directly to the sequence windows $\vW_m$. Some Transformer models, including the Time-Series Transformer, Crossformer, and Spacetimeformer, use an encoder--decoder formulation in their original forecasting setting. That formulation assumes historical target values, historical covariates, and future-horizon covariates.

This formulation is not directly applicable to PHM prognostics and diagnostics because the past values of the target variable, such as remaining useful life, are unavailable at inference time, and no future information is used. We therefore adapt encoder--decoder Transformer architectures by removing the encoder and directly feeding $\vW_m$ to the decoder. This step is not needed for PatchTST, which is decoder-only by design. For PatchTST, channel-independent processing cannot be applied to the regression target for the same reason, so decoder latents are concatenated and a separate regression or classification head is trained. For TiDE, we disable the encoder pathway and use the dynamic covariate pathway to process $\vW_m$. For diagnostics, the regression head is replaced with a classification head that outputs $n$ classes.

The transformer training regimen uses a warm-up phase and reduces the learning rate upon plateau. Min-max or z-score normalization is selected depending on the dataset. All preprocessing, feature extraction, and sequence slicing, represented by $\mathcal{G}$ and $\mathcal{S}$, are shared among the models; tabular foundation models and XGBoost additionally share the tabularization schema $\mathcal{T}$.

\subsection{\reviewheading{TabPFN and TabDPT Details}}
\label{app:model_tabpfn_tabdpt}

TabPFN \citep{hollmanntabpfn, hollmann2025accurate} is a Prior-Data Fitted Network (PFN) trained on a large-scale synthetic dataset that consists of millions of small tabular instances. These tabular instances are generated from randomly sampled structural causal models. TabPFN therefore leverages the structural diversity seen during pre-training to implicitly understand and generalize to the relational structure of unseen tabular data. This enables the network to efficiently approximate the Bayesian posterior for the chosen prior:

\begin{equation}
    p(y^{\text{test}} \mid \mathcal{D}_{\mathrm{tab}}^{\text{test}}, \mathcal{Y}_{\mathrm{tab}}^{\text{train}}, \mathcal{D}_{\mathrm{tab}}^{\text{train}})
\end{equation}

Internally, TabPFN converts tabular cells into a sequence of high-dimensional feature tokens by mapping every categorical or scalar attribute to a fixed-dimensional embedding. The attribute embedding strategy is therefore similar to Spacetimeformer, which flattens spatial-temporal time sequences and embeds each signal and time coordinate in \(\vW_k \in \mathbb{R}^{L_{\mathrm{seq}} \times F}\) for a total of \(L_{\mathrm{seq}} \times F\) embeddings.
In TabPFN, the final embedding is obtained by multiplying each raw feature value with an attribute-specific offset vector that places all attributes in a shared latent space.
TabPFN is a univariate model. For multivariate outputs in PHM, we therefore parallelize the process by fitting one model per variate.
To further enhance robustness, we use an ensemble of eight TabPFN estimators.

We also evaluate TabDPT \citep{ma2024tabdpt}, a tabular foundation model that has been trained purely on real data. Due to its capability to perform classification and regression, we employ TabDPT as a retrieval-based discriminatively pretrained transformer that operates on full row representations. TabDPT supports up to $D_{\max} = 100$ features, so when $D_{\mathrm{tab}}$ exceeds this limit TabDPT applies Principal Component Analysis to obtain a projection with exactly $D_{\max}$ dimensions. TabDPT separately embeds features and labels linearly into a 768 dimensional space and combines them through element-wise addition.

During inference, TabDPT builds a local context by selecting the top $K$ most similar training rows using the FAISS library~\cite{douze2024faiss}.  The full transformer sequence is formed by concatenating the embedded training samples and the embedded evaluation sample, where the training embeddings are shifted by their label embeddings and the evaluation sample is kept unlabeled.  As for TabPFN, we use the ensemble of eight TabDPT estimators, each trained on a random subset of columns, and aggregate their outputs through a weighted average. We ensure that there is no overlap between the datasets evaluated in this work and the pre-training data of TabPFN.

\clearpage
\section{\reviewheading{Additional Experimental Results}}
\label{app:additional_results}
This appendix presents the complete experimental setup and results for the benchmark evaluation. It documents the evaluated experiment families, preprocessing schemas, and hyperparameter search spaces. For the condensed main-text summary, see Section~\ref{sec:empirical_validation}; for dataset-level protocol details, see Appendix~\ref{sec:datasets}.

\subsection{Experimental setup}
\label{app:exp_setup_full}

\paragraph{Experiment definitions.}
Each experiment family is defined by a Hydra configuration \cite{Yadan2019Hydra} that specifies the datasource, transform pipeline, model, evaluator, seed set, and hyperparameter search space. The benchmark covers gradient-trained and fit-predict models for both diagnostics and prognostics tasks. XGBoost uses a separate fit-predict configuration that reuses the same experiment definitions and transform families as the other tabular models.

\paragraph{Learning tasks and datasets.}
The evaluation covers two PHM task categories. Diagnostics comprises multiclass fault classification on MZVAV~\citep{Granderson2020}, four component-specific hydraulic diagnostics tasks on HSF15~\citep{hsf15_helwig} (accumulator, cooler, pump, and valve), and concept classification on N-CMAPSS Multi~\citep{arias2021aircraft, frederick2007user}. Prognostics comprises ah-RUL regression on NB14~\citep{bole2014adaptation} and UNIBO21~\citep{univbo_dataset}, direct RUL regression on PHME20~\citep{PHME20-GTU}, RUL prediction on the N-CMAPSS Multi and DS02 families, and bearing prognostics on XJTU-SY~\citep{yaguo2019xjtu}.

Target semantics differ by family. NB14 and UNIBO21 predict ah-RUL, i.e., remaining cumulative discharge throughput in Ampere-hours; PHME20 and the N-CMAPSS families predict remaining useful life; and the bearing family transforms runtime trajectories into a normalized degradation target through the configured health-index transform, while evaluation is reported with the per-unit metric suite. All families use disjoint train/validation/test entities according to their datasource definitions, with MZVAV as the only day-stratified diagnostics exception.

\paragraph{Models.}
The benchmark compares five model families: (i) simple baselines (Linear, Exp, MLP), (ii) deep sequence models (LSTM, CNN-1D, TiDE), (iii) transformers (TST, STF, CF, PTST), (iv) tabular models (XGBoost), and (v) tabular foundation models (TabPFN, TabDPT). Linear is regression-only and is therefore omitted from diagnostics; for diagnostics, a linear classifier serves the analogous baseline role.

\paragraph{Training and evaluation.}
Every model--dataset configuration is repeated over five random seeds. Gradient-trained models search over \mbox{$\texttt{seq\_len} \in \{1,10,50\}$} and \mbox{$\texttt{lr} \in \{0.001, 0.0005, 0.0001\}$}, with batch size 512, a maximum of 200 epochs, and early stopping. Fit-predict models search over context/stride pairs $(1,1)$, $(5,1)$, $(10,5)$, $(20,5)$, and $(50,50)$. Diagnostics selects the best configuration on \texttt{val/f1}; prognostics (generic RUL objective) selects on \texttt{val/loss}. Evaluation uses three evaluator families: \texttt{classification} (diagnostics), \texttt{rul} (direct RUL regression), and \texttt{per\_unit} (battery and bearing families).

\subsection{Transformation schemas}
\label{app:transform_schemas}

All experiment families obey the same leakage-prevention invariant: any fitted normalization or feature-extraction statistic is estimated on the training partition only and then reused unchanged for validation and test. The subsections below describe the preprocessing pipeline applied to each dataset group; Across all families, the transformed tensors are subsequently windowed according to the sequence-length and stride settings described in Section~\ref{app:hyperparameter_search}.

\subsubsection{Battery datasets (NB14 and UNIBO21)}
\label{app:schema_battery}

For both NB14 and UNIBO21, raw sensor channels and the ah-RUL target are min-max scaled (MinMaxScaler). Two descriptor branches are then extracted from the cycle traces: time-domain statistics---mean, variance, kurtosis, peak factor, and related summaries (TimeStatsTransform)---and frequency-domain statistics via FFT (SpectralStatsTransform). The resulting descriptor vectors are concatenated (ConcatenateTransform) and re-scaled with a final min-max transform. For fit-predict models, the pipeline additionally tabularizes the processed history into a single feature vector for one-shot inference (TimeseriesTabularizer). This pipeline corresponds to the \texttt{combined} / \texttt{combined\_fit\_predict} configuration group.

\subsubsection{Bearing dataset (XJTU-SY)}
\label{app:schema_bearings}

For XJTU-SY, raw vibration channels are standardized with training-partition statistics (StandardScaler). Runtime targets are converted into a normalized degradation trajectory via HealthIndexTransform. An additional cumulative-sum feature is computed and min-max scaled, and time-domain plus spectral descriptors (TimeStatsTransform, SpectralStatsTransform) are extracted from the standardized vibration signals. The resulting features are concatenated and min-max rescaled. For fit-predict models, the pipeline tabularizes the combined representation while preserving unit identifiers for per-unit evaluation. This pipeline corresponds to the \texttt{combined} / \texttt{combined\_fit\_predict} configuration group.

\subsubsection{N-CMAPSS families}
\label{app:schema_ncmapss}

Sensor features and operating descriptors are standardized with fixed N-CMAPSS scalers (N\_CMAPSSDescriptorsScaler, StandardScaler), and the RUL label is multiplied by the constant factor 0.01 (ConstantScaler). Each flight is then temporally aggregated with non-overlapping windows of 60~timesteps (WindowedAggregationTransform), and the aggregated sensor features and descriptors are concatenated into a single input representation. For the multi-source diagnostics family, the transform additionally builds unified concept classes from the source-specific concept annotations. Fit-predict models tabularize the aggregated histories after the same scaling and concatenation stages. This pipeline corresponds to the \texttt{depater2023\_default} / \texttt{depater2023\_fit\_predict\_history} configuration group.

\subsubsection{Building diagnostics (MZVAV)}
\label{app:schema_mzvav}

No additional feature engineering is applied for MZVAV. The pipeline rescales the sensor channels with a precomputed dataset-specific min-max scaler (MinMaxScalerMZVAV) and routes the scalar target directly to the fault-classification key consumed by the diagnostics models. For fit-predict models, the pipeline tabularizes the resulting history windows without introducing a separate descriptor stage. This pipeline corresponds to the \texttt{default} / \texttt{fit\_predict\_history} configuration group.

\subsubsection{Hydraulic diagnostics (HSF15)}
\label{app:schema_hsf15}

All four HSF15 component tasks share the same preprocessing pipeline. Features are min-max scaled (MinMaxScaler), the component-specific target is reduced to the last label in each window and assigned to the fault-classification key, and the sensor burst is summarized through time-domain statistics (TimeStatsTransform) and spectral statistics (SpectralStatsTransform), followed by concatenation and a final min-max rescaling. For fit-predict models, the pipeline tabularizes the statistics-based representation for XGBoost, TabPFN, and TabDPT. This pipeline corresponds to the \texttt{default} / \texttt{statistics\_fit\_predict} configuration group.

\subsubsection{PHM challenge prognostics (PHME20)}
\label{app:schema_phme20}

Both features and the direct RUL target are min-max scaled (MinMaxScaler), with no additional handcrafted feature extraction. For fit-predict models, the pipeline adds history tabularization for one-shot inference. This pipeline corresponds to the \texttt{normalize\_feature\_target} / \texttt{normalize\_feature\_target\_fit\_predict} configuration group.

\subsection{Hyperparameter search}
\label{app:hyperparameter_search}

The benchmark exposes two hyperparameter-search families: a gradient-trained family for sequence models and simple neural baselines, and a fit-predict family for tabular/foundation models including XGBoost. Table~\ref{tab:hyperparameter_search} summarizes the search spaces.

\begin{table}[H]
\centering
\footnotesize
\caption{Hyperparameter search families used in the benchmark evaluation.}
\label{tab:hyperparameter_search}
\renewcommand{\arraystretch}{1.08}
\begin{adjustbox}{max width=\linewidth}
\begin{tabular}{@{}p{2.1cm}p{4.1cm}p{4.8cm}p{2.4cm}p{2.1cm}@{}}
\toprule
\textbf{Search family} & \textbf{Models} & \textbf{Search space} & \textbf{Fixed settings} & \textbf{Selection rule} \\
\midrule
Gradient-trained & LSTM, 1D-CNN, STF, Crossformer, Timeseries Transformer, TiDE, PatchTST, MLP, linear classifier, linear regression, exponential regression & \texttt{seq\_len} $\in \{1,10,50\}$; \texttt{lr} $\in \{0.001, 0.0005, 0.0001\}$ & Batch size 512; max 200 epochs; early stopping & \texttt{val/f1} for diagnostics; \texttt{val/loss} for prognostics \\
Fit-predict & XGBoost, TabPFN, TabDPT & Context/stride pairs $(1,1)$, $(5,1)$, $(10,5)$, $(20,5)$, $(50,50)$ & Five seeds; deterministic transform reuse; one-shot fit/predict evaluation & \texttt{val/f1} for diagnostics; \texttt{val/loss} for prognostics \\
\bottomrule
\end{tabular}
\end{adjustbox}
\end{table}

All successful runs persist the resolved configuration, Hydra override trace, run metadata, and reproduction instructions alongside the predictions and evaluator outputs. The corresponding artifact protocol is described in Appendix~\ref{app:reproducibility}.

\subsection{Reading the result tables}
\label{app:reading_guide}

The result tables in Sections~\ref{app:full_diagnostics_results}--\ref{app:full_prognostics_results} share a common reading convention, summarized here once so individual captions can stay terse.

\paragraph{Dataset abbreviations.}
Column headers use short identifiers consistently across both the main-text and appendix tables. Their full meanings are:

\begin{table}[H]
\centering
\footnotesize
\renewcommand{\arraystretch}{1.05}
\begin{tabular}{@{}llp{6.0cm}@{}}
\toprule
\textbf{Abbreviation} & \textbf{Task} & \textbf{Source dataset (domain)} \\
\midrule
NC-DS02 & Prognostics & N-CMAPSS DS02 (turbofan engine RUL) \\
NC-P    & Prognostics & N-CMAPSS Multi-source (turbofan engine RUL) \\
NB14    & Prognostics & NASA Randomized Battery Usage (battery ah-RUL) \\
PHME20  & Prognostics & PHM 2020 Challenge (industrial filtration RUL) \\
Unibo   & Prognostics & UNIBO Powertools (battery ah-RUL) \\
XJTU-SY & Prognostics & XJTU-SY (bearing degradation) \\
NC-D    & Diagnostics & N-CMAPSS Multi-source (turbofan concept classification) \\
HSF15-A & Diagnostics & HSF15 (hydraulic accumulator, 4-way) \\
HSF15-C & Diagnostics & HSF15 (hydraulic cooler, 3-way) \\
HSF15-P & Diagnostics & HSF15 (hydraulic pump, 3-way) \\
HSF15-V & Diagnostics & HSF15 (hydraulic valve, 4-way) \\
MZVAV   & Diagnostics & MZVAV (multi-zone HVAC fault, 4-way) \\
\bottomrule
\end{tabular}
\end{table}%

\paragraph{Cell convention.}
Each cell reports mean $\pm$ std over five independent seeds. Bold cells are the best in their column; underlined cells are the second best. Rows are grouped by model family in the order: simple baselines, deep sequence models, transformers, tabular models, tabular foundation models.

\paragraph{Metric scaling and direction.}
Diagnostics metrics (F1, AUROC, Accuracy) are reported on a 0--100 scale; prognostics MAE/MSE are reported both in the framework's normalized target space (reported $\times 100$) and in original engineering units (denormalized) for practitioner interpretation. Arrows in column headers indicate metric direction ($\downarrow$ lower is better, $\uparrow$ higher is better). The \emph{Avg rank} column is the mean of per-task ranks across the columns of that table.

\subsection{Diagnostics results}
\label{app:full_diagnostics_results}

Diagnostics is evaluated by F1 (the headline metric in Section~\ref{sec:empirical_validation}) and complemented here by AUROC and Accuracy. The three metrics agree on the top group: TabDPT, TabPFN, CNN-1D, and XGBoost cluster within $1.0$ average-rank of each other on F1, with LSTM joining the leading tier on F1 and Accuracy. MZVAV (multi-zone HVAC fault classification under a day-stratified split) is the hardest task in every metric and the only family where the gap to chance is small. The metric-robustness check therefore supports the F1 choice in the main text.

\begin{table}[H]
\caption{F1 score on diagnostics ($\uparrow$). Same numbers as the diagnostics floor of Table~\ref{tab:results_main_combined}, reproduced here with full per-task breakdown.}
\label{tab:results_test_best_rerun_f1}
\begin{adjustbox}{max width=\textwidth}
\begin{tabular}{lrrrrrrr}
\toprule
Model & \multicolumn{1}{c}{NC-D} & \multicolumn{1}{c}{HSF15-A} & \multicolumn{1}{c}{HSF15-C} & \multicolumn{1}{c}{HSF15-P} & \multicolumn{1}{c}{HSF15-V} & \multicolumn{1}{c}{MZVAV} & \multicolumn{1}{c}{Average rank} \\
\midrule
Linear & 72.34 ± 3.04 & 58.75 ± 1.81 & 98.35 ± 0.82 & 54.40 ± 11.97 & 32.55 ± 2.36 & 39.89 ± 8.77 & 7.33 \\
MLP & 79.49 ± 1.80 & 91.02 ± 2.25 & \underline{99.91 ± 0.14} & 97.32 ± 0.60 & 80.99 ± 29.38 & 60.10 ± 6.39 & 5.00 \\
LSTM & \cellcolor[gray]{0.85}\textbf{88.84 ± 0.73} & 94.59 ± 0.97 & \cellcolor[gray]{0.85}\textbf{100.00 ± 0.00} & 95.94 ± 2.83 & 97.35 ± 3.32 & 51.31 ± 6.01 & 3.83 \\
CNN-1D & \underline{87.53 ± 2.83} & 94.03 ± 1.98 & \cellcolor[gray]{0.85}\textbf{100.00 ± 0.00} & 98.73 ± 0.52 & 97.92 ± 0.83 & \underline{66.11 ± 5.74} & 3.00 \\
TiDE & 32.57 ± 4.65 & 42.90 ± 5.07 & 61.57 ± 10.41 & 59.37 ± 7.68 & 42.11 ± 14.36 & 25.19 ± 5.29 & 8.17 \\
TST & 26.34 ± 3.96 & 37.37 ± 4.38 & 59.18 ± 10.29 & 46.07 ± 4.13 & 35.40 ± 5.77 & 24.93 ± 4.38 & 10.00 \\
STF & 24.55 ± 3.60 & 40.01 ± 5.75 & 65.94 ± 20.61 & 50.39 ± 11.54 & 37.34 ± 5.68 & 38.01 ± 5.56 & 8.67 \\
CF & 23.74 ± 0.93 & 25.04 ± 5.28 & 59.19 ± 10.05 & 29.46 ± 2.61 & 23.98 ± 2.95 & 17.34 ± 4.24 & 11.50 \\
PTST & 19.57 ± 0.26 & 31.56 ± 4.04 & 41.57 ± 5.18 & 41.22 ± 6.63 & 26.20 ± 2.44 & 25.81 ± 4.15 & 11.00 \\
XGBoost & 48.13 ± 0.00 & \underline{98.07 ± 0.00} & \cellcolor[gray]{0.85}\textbf{100.00 ± 0.00} & \underline{99.66 ± 0.00} & \underline{99.65 ± 0.00} & 57.08 ± 0.00 & 3.17 \\
TabPFN & 67.15 ± 1.46 & \cellcolor[gray]{0.85}\textbf{99.47 ± 0.00} & \cellcolor[gray]{0.85}\textbf{100.00 ± 0.00} & \cellcolor[gray]{0.85}\textbf{100.00 ± 0.00} & \cellcolor[gray]{0.85}\textbf{100.00 ± 0.00} & 58.32 ± 2.44 & 2.33 \\
TabDPT & 85.21 ± 0.16 & 96.66 ± 1.03 & \cellcolor[gray]{0.85}\textbf{100.00 ± 0.00} & 99.06 ± 0.25 & 98.92 ± 0.35 & \cellcolor[gray]{0.85}\textbf{71.29 ± 0.48} & 2.33 \\
\bottomrule
\end{tabular}
\end{adjustbox}
\end{table}

\begin{table}[H]
\caption{AUROC on diagnostics ($\uparrow$).}
\label{tab:results_test_best_rerun_auroc}
\begin{adjustbox}{max width=\textwidth}
\begin{tabular}{lrrrrrrr}
\toprule
Model & \multicolumn{1}{c}{NC-D} & \multicolumn{1}{c}{HSF15-A} & \multicolumn{1}{c}{HSF15-C} & \multicolumn{1}{c}{HSF15-P} & \multicolumn{1}{c}{HSF15-V} & \multicolumn{1}{c}{MZVAV} & \multicolumn{1}{c}{Average rank} \\
\midrule
Linear & 93.76 ± 1.29 & 85.10 ± 1.35 & \underline{99.43 ± 0.63} & 77.41 ± 12.99 & 64.74 ± 4.49 & 71.47 ± 9.58 & 7.17 \\
MLP & 96.31 ± 0.67 & 98.95 ± 0.39 & \cellcolor[gray]{0.85}\textbf{100.00 ± 0.00} & 99.84 ± 0.07 & 90.02 ± 21.11 & 85.24 ± 2.03 & 5.17 \\
LSTM & \cellcolor[gray]{0.85}\textbf{98.41 ± 0.16} & 99.54 ± 0.11 & \cellcolor[gray]{0.85}\textbf{100.00 ± 0.00} & 99.68 ± 0.27 & 99.88 ± 0.17 & \underline{86.32 ± 4.06} & 3.33 \\
CNN-1D & \underline{98.11 ± 0.62} & 99.44 ± 0.19 & \cellcolor[gray]{0.85}\textbf{100.00 ± 0.00} & 99.94 ± 0.06 & 99.94 ± 0.05 & 85.45 ± 2.29 & 3.17 \\
TiDE & 62.71 ± 3.84 & 69.00 ± 5.66 & 81.46 ± 5.14 & 78.78 ± 11.24 & 64.91 ± 11.98 & 60.77 ± 9.35 & 8.00 \\
TST & 57.03 ± 2.91 & 62.99 ± 6.37 & 79.87 ± 9.04 & 65.30 ± 6.07 & 61.60 ± 5.19 & 57.44 ± 9.01 & 9.33 \\
STF & 55.23 ± 3.04 & 58.44 ± 5.10 & 81.96 ± 14.13 & 64.97 ± 9.06 & 59.98 ± 15.81 & 62.61 ± 10.35 & 9.50 \\
CF & 52.00 ± 0.87 & 53.48 ± 5.54 & 73.59 ± 8.17 & 37.35 ± 0.53 & 51.76 ± 0.94 & 47.64 ± 6.33 & 11.67 \\
PTST & 50.01 ± 0.24 & 58.74 ± 1.64 & 59.83 ± 4.32 & 62.24 ± 5.03 & 52.84 ± 2.86 & 52.97 ± 5.01 & 11.17 \\
XGBoost & 82.56 ± 0.00 & \underline{99.94 ± 0.00} & \cellcolor[gray]{0.85}\textbf{100.00 ± 0.00} & \cellcolor[gray]{0.85}\textbf{100.00 ± 0.00} & \cellcolor[gray]{0.85}\textbf{100.00 ± 0.00} & 84.51 ± 0.00 & 3.17 \\
TabPFN & 93.97 ± 0.31 & \cellcolor[gray]{0.85}\textbf{100.00 ± 0.00} & \cellcolor[gray]{0.85}\textbf{100.00 ± 0.00} & \cellcolor[gray]{0.85}\textbf{100.00 ± 0.00} & \cellcolor[gray]{0.85}\textbf{100.00 ± 0.00} & 83.67 ± 0.77 & 2.67 \\
TabDPT & 97.38 ± 0.06 & 99.91 ± 0.05 & \cellcolor[gray]{0.85}\textbf{100.00 ± 0.00} & \underline{99.99 ± 0.00} & \underline{99.99 ± 0.00} & \cellcolor[gray]{0.85}\textbf{88.57 ± 0.73} & 2.50 \\
\bottomrule
\end{tabular}
\end{adjustbox}
\end{table}

\begin{table}[H]
\caption{Accuracy on diagnostics ($\uparrow$).}
\label{tab:results_test_best_rerun_accuracy}
\begin{adjustbox}{max width=\textwidth}
\begin{tabular}{lrrrrrrr}
\toprule
Model & \multicolumn{1}{c}{NC-D} & \multicolumn{1}{c}{HSF15-A} & \multicolumn{1}{c}{HSF15-C} & \multicolumn{1}{c}{HSF15-P} & \multicolumn{1}{c}{HSF15-V} & \multicolumn{1}{c}{MZVAV} & \multicolumn{1}{c}{Average rank} \\
\midrule
Linear & 73.79 ± 2.35 & 61.47 ± 1.91 & 98.36 ± 0.82 & 63.08 ± 14.59 & 45.93 ± 2.01 & 61.56 ± 4.94 & 7.33 \\
MLP & 79.72 ± 1.82 & 92.39 ± 2.06 & \underline{99.91 ± 0.14} & 98.05 ± 0.41 & 85.28 ± 22.19 & 71.54 ± 4.90 & 5.00 \\
LSTM & \cellcolor[gray]{0.85}\textbf{88.42 ± 0.72} & 95.42 ± 0.82 & \cellcolor[gray]{0.85}\textbf{100.00 ± 0.00} & 96.96 ± 1.94 & 97.64 ± 3.10 & 70.71 ± 4.77 & 3.50 \\
CNN-1D & \underline{86.91 ± 2.91} & 95.00 ± 1.57 & \cellcolor[gray]{0.85}\textbf{100.00 ± 0.00} & 99.09 ± 0.36 & 98.38 ± 0.67 & \underline{77.81 ± 3.10} & 3.00 \\
TiDE & 35.30 ± 3.73 & 49.24 ± 10.75 & 65.39 ± 8.98 & 70.93 ± 3.78 & 57.08 ± 10.30 & 30.15 ± 7.47 & 8.50 \\
TST & 26.80 ± 3.37 & 43.36 ± 8.23 & 62.53 ± 10.12 & 52.24 ± 6.62 & 45.19 ± 6.74 & 45.19 ± 9.25 & 9.83 \\
STF & 24.82 ± 3.12 & 47.02 ± 9.33 & 72.30 ± 15.62 & 64.90 ± 9.85 & 57.59 ± 4.35 & 50.34 ± 9.76 & 8.50 \\
CF & 28.90 ± 3.18 & 30.71 ± 5.80 & 61.83 ± 10.46 & 35.15 ± 4.49 & 33.06 ± 3.25 & 27.02 ± 10.45 & 11.17 \\
PTST & 21.56 ± 0.93 & 34.41 ± 4.03 & 42.44 ± 4.66 & 43.36 ± 7.01 & 31.85 ± 5.87 & 39.70 ± 5.78 & 11.33 \\
XGBoost & 54.75 ± 0.00 & \underline{98.53 ± 0.00} & \cellcolor[gray]{0.85}\textbf{100.00 ± 0.00} & \underline{99.77 ± 0.00} & \underline{99.77 ± 0.00} & 62.62 ± 0.00 & 3.33 \\
TabPFN & 70.61 ± 1.07 & \cellcolor[gray]{0.85}\textbf{99.58 ± 0.00} & \cellcolor[gray]{0.85}\textbf{100.00 ± 0.00} & \cellcolor[gray]{0.85}\textbf{100.00 ± 0.00} & \cellcolor[gray]{0.85}\textbf{100.00 ± 0.00} & 65.55 ± 1.96 & 2.50 \\
TabDPT & 84.74 ± 0.16 & 97.27 ± 0.83 & \cellcolor[gray]{0.85}\textbf{100.00 ± 0.00} & 99.27 ± 0.19 & 99.17 ± 0.26 & \cellcolor[gray]{0.85}\textbf{80.10 ± 0.36} & 2.33 \\
\bottomrule
\end{tabular}
\end{adjustbox}
\end{table}

\paragraph{Cross-family observations.}
Across the three diagnostics metrics, the model families separate clearly. Tabular foundation models (TabDPT, TabPFN) and XGBoost rank in the top tier even though they consume tabularized windows rather than raw sequences; this is most visible on HSF15 (where many models reach 100 F1), but it also holds on harder tasks such as NC-D and MZVAV. Deep sequence models split into a strong pair (LSTM, CNN-1D) and a weaker TiDE variant under this training budget. By contrast, the transformer family (TST, STF, CF, PTST) performs near chance on most diagnostics tasks, despite consuming identical inputs under the same protocol. Simple baselines (MLP, Linear) remain competitive on several HSF15 components, reinforcing that hydraulic component diagnostics is comparatively easy relative to MZVAV and NC-D. Finally, the robustness check is consistent: F1, AUROC, and Accuracy induce nearly identical rankings, so the headline conclusions do not depend on the specific metric choice.

\subsection{Prognostics results}
\label{app:full_prognostics_results}

The prognostics metric surface is reported in three families. Aggregate MAE and MSE (normalized and denormalized) are computed by pooling all predictions and all units before averaging. Per-unit aggregation, defined only on battery and bearing tasks where the framework's \texttt{per\_unit} evaluator is active, computes one error per monitored unit before averaging across units; this tightens rankings among leading models and reduces the influence of long-trajectory units that dominate window-level pools. Domain-specific scores (NASA, PHM) are reported with the per-task scoping enforced by the framework's metric registry.

\subsubsection{Aggregate errors}
\label{app:prog_aggregate}

Tables~\ref{tab:results_app_mae} and~\ref{tab:results_app_mse} report MAE and MSE in a two-block format. The top block reports errors in the normalized target space (reported $\times 100$), which is used to compute cross-task \emph{Avg rank}; the bottom block reports the same predictions in original engineering units for practitioner interpretation. MSE emphasizes occasional large errors, but it preserves the headline ordering among the leading models. Because MSE penalizes large residuals quadratically, it is sensitive to rare catastrophic predictions; we report it alongside MAE to make these failure modes visible.

\begin{table}[H]
\centering
\footnotesize
\caption{MAE on prognostics. Top block: normalized target space ($\times 100$) ($\downarrow$); bottom block: original engineering units ($\downarrow$). Same numbers as the prognostics floor of Table~\ref{tab:results_main_combined} (top), reproduced with full per-task breakdown.}
\label{tab:results_app_mae}
\begin{adjustbox}{max width=\textwidth}
\begin{tabular}{lrrrrrrr}
\toprule
Model & \multicolumn{1}{c}{NC-DS02} & \multicolumn{1}{c}{NC-P} & \multicolumn{1}{c}{NB14} & \multicolumn{1}{c}{PHME20} & \multicolumn{1}{c}{Unibo} & \multicolumn{1}{c}{XJTU-SY} & \multicolumn{1}{c}{Average rank} \\
\midrule
\multicolumn{8}{l}{\textit{Normalized target space ($\downarrow$)}} \\
\midrule
Linear & 10.13 ± 0.14 & 16.11 ± 0.60 & 41.69 ± 12.02 & 12.19 ± 0.36 & 27.59 ± 14.36 & 76.80 ± 60.41 & 12.50 \\
Exp & 5.35 ± 0.06 & 10.96 ± 0.09 & 30.47 ± 47.76 & 8.82 ± 0.52 & 12.19 ± 0.31 & 27.22 ± 4.06 & 9.67 \\
MLP & 6.37 ± 0.23 & 13.17 ± 0.78 & 14.38 ± 9.77 & 4.62 ± 1.15 & 12.50 ± 0.76 & 30.64 ± 2.67 & 10.33 \\
LSTM & \underline{4.93 ± 0.13} & 7.56 ± 0.31 & 3.80 ± 0.22 & 3.73 ± 0.98 & 6.50 ± 0.16 & \cellcolor[gray]{0.85}\textbf{21.89 ± 0.40} & 3.67 \\
CNN-1D & 5.33 ± 0.37 & 7.53 ± 0.22 & 8.89 ± 1.70 & 5.35 ± 3.71 & 12.41 ± 1.15 & 31.02 ± 8.25 & 8.67 \\
TiDE & 5.29 ± 0.22 & 7.62 ± 0.20 & \cellcolor[gray]{0.85}\textbf{3.44 ± 0.17} & 4.20 ± 0.66 & 6.46 ± 0.78 & 25.11 ± 2.38 & 5.17 \\
TST & 5.31 ± 0.13 & \underline{7.02 ± 0.17} & 6.28 ± 0.25 & 4.11 ± 0.84 & 7.23 ± 0.39 & 33.30 ± 7.72 & 7.00 \\
STF & \cellcolor[gray]{0.85}\textbf{4.89 ± 0.10} & 7.35 ± 1.16 & 10.67 ± 3.16 & 3.91 ± 1.00 & 8.89 ± 0.81 & 28.49 ± 4.01 & 6.17 \\
CF & 5.76 ± 0.51 & 9.98 ± 0.57 & \underline{3.57 ± 0.07} & 3.87 ± 0.85 & 5.58 ± 1.08 & \underline{22.09 ± 1.06} & 5.00 \\
PTST & 16.62 ± 0.04 & 21.55 ± 0.03 & 5.22 ± 0.10 & 15.09 ± 1.13 & 11.18 ± 1.11 & 25.42 ± 1.48 & 10.33 \\
XGBoost & 8.52 ± 0.00 & 15.24 ± 0.00 & 4.48 ± 0.00 & 2.68 ± 0.00 & 4.06 ± 0.00 & 24.59 ± 0.00 & 6.50 \\
TabPFN & 4.96 ± 0.04 & 7.79 ± 0.04 & 3.91 ± 0.03 & \cellcolor[gray]{0.85}\textbf{1.95 ± 0.03} & \cellcolor[gray]{0.85}\textbf{3.72 ± 0.06} & 22.27 ± 0.35 & 3.33 \\
TabDPT & 5.07 ± 0.06 & \cellcolor[gray]{0.85}\textbf{6.85 ± 0.02} & 3.63 ± 0.04 & \underline{2.19 ± 0.01} & \underline{3.94 ± 0.05} & 23.24 ± 0.45 & 2.67 \\
\midrule
\multicolumn{8}{l}{\textit{Original engineering units ($\downarrow$)}} \\
\midrule
Linear & 10.13 ± 0.14 & 16.11 ± 0.60 & 451.64 ± 130.15 & 43.13 ± 1.27 & 135.80 ± 70.68 & 907.40 ± 774.36 & 12.50 \\
Exp & 5.35 ± 0.06 & 10.96 ± 0.09 & 330.02 ± 517.36 & 31.21 ± 1.85 & 60.02 ± 1.51 & 349.71 ± 60.98 & 9.67 \\
MLP & 6.37 ± 0.23 & 13.17 ± 0.78 & 155.77 ± 105.84 & 16.35 ± 4.06 & 61.51 ± 3.72 & 384.95 ± 40.56 & 10.50 \\
LSTM & \underline{4.93 ± 0.13} & 7.56 ± 0.31 & 41.13 ± 2.34 & 13.19 ± 3.47 & 32.00 ± 0.77 & \cellcolor[gray]{0.85}\textbf{271.33 ± 3.45} & 3.67 \\
CNN-1D & 5.33 ± 0.37 & 7.53 ± 0.22 & 96.25 ± 18.38 & 18.95 ± 13.12 & 61.10 ± 5.66 & 380.05 ± 96.92 & 8.50 \\
TiDE & 5.29 ± 0.22 & 7.62 ± 0.20 & \cellcolor[gray]{0.85}\textbf{37.22 ± 1.80} & 14.86 ± 2.35 & 31.78 ± 3.86 & 320.41 ± 24.80 & 5.33 \\
TST & 5.31 ± 0.13 & \underline{7.02 ± 0.17} & 68.00 ± 2.71 & 14.56 ± 2.96 & 35.60 ± 1.94 & 424.88 ± 100.25 & 7.00 \\
STF & \cellcolor[gray]{0.85}\textbf{4.89 ± 0.10} & 7.35 ± 1.16 & 115.61 ± 34.23 & 13.83 ± 3.53 & 43.76 ± 4.01 & 361.41 ± 54.88 & 6.17 \\
CF & 5.76 ± 0.51 & 9.98 ± 0.57 & \underline{38.70 ± 0.80} & 13.71 ± 3.02 & 27.48 ± 5.31 & \underline{282.20 ± 19.10} & 5.00 \\
PTST & 16.62 ± 0.04 & 21.55 ± 0.03 & 56.53 ± 1.09 & 53.40 ± 3.99 & 55.02 ± 5.47 & 312.68 ± 23.46 & 10.17 \\
XGBoost & 8.52 ± 0.00 & 15.24 ± 0.00 & 48.54 ± 0.00 & 9.48 ± 0.00 & 19.98 ± 0.00 & 311.19 ± 0.00 & 6.50 \\
TabPFN & 4.96 ± 0.04 & 7.79 ± 0.04 & 42.32 ± 0.28 & \cellcolor[gray]{0.85}\textbf{6.89 ± 0.09} & \cellcolor[gray]{0.85}\textbf{18.33 ± 0.30} & 292.43 ± 5.26 & 3.50 \\
TabDPT & 5.07 ± 0.06 & \cellcolor[gray]{0.85}\textbf{6.85 ± 0.02} & 39.35 ± 0.47 & \underline{7.75 ± 0.04} & \underline{19.42 ± 0.24} & 282.83 ± 5.84 & 2.50 \\
\bottomrule
\end{tabular}
\end{adjustbox}
\end{table}

\begin{table}[H]
\centering
\footnotesize
\caption{MSE on prognostics. Top block: normalized target space ($\times 100$) ($\downarrow$); bottom block: original engineering units ($\downarrow$). MSE redistributes weight onto large per-window errors but preserves the leading-model ranking from MAE.}
\label{tab:results_app_mse}
\begin{adjustbox}{max width=\textwidth}
\begin{tabular}{lrrrrrrr}
\toprule
Model & \multicolumn{1}{c}{NC-DS02} & \multicolumn{1}{c}{NC-P} & \multicolumn{1}{c}{NB14} & \multicolumn{1}{c}{PHME20} & \multicolumn{1}{c}{Unibo} & \multicolumn{1}{c}{XJTU-SY} & \multicolumn{1}{c}{Average rank} \\
\midrule
\multicolumn{8}{l}{\textit{Normalized target space ($\downarrow$)}} \\
\midrule
Linear & 1.37 ± 0.04 & 3.95 ± 0.38 & 32.16 ± 16.14 & 2.13 ± 0.10 & 17.83 ± 17.83 & 107.37 ± 128.29 & 12.50 \\
Exp & 0.47 ± 0.01 & 1.92 ± 0.03 & 28.60 ± 61.10 & 1.18 ± 0.13 & 2.59 ± 0.08 & 11.21 ± 3.67 & 9.50 \\
MLP & 0.76 ± 0.05 & 2.77 ± 0.35 & 5.09 ± 6.91 & 0.35 ± 0.15 & 2.91 ± 0.51 & 14.80 ± 3.26 & 10.50 \\
LSTM & \underline{0.43 ± 0.02} & 1.04 ± 0.09 & 0.27 ± 0.03 & 0.23 ± 0.11 & 1.38 ± 0.07 & \cellcolor[gray]{0.85}\textbf{6.80 ± 0.22} & 3.83 \\
CNN-1D & 0.48 ± 0.05 & 1.01 ± 0.06 & 1.31 ± 0.60 & 0.60 ± 0.75 & 2.54 ± 0.48 & 14.34 ± 8.16 & 8.33 \\
TiDE & 0.47 ± 0.04 & 1.09 ± 0.04 & \cellcolor[gray]{0.85}\textbf{0.22 ± 0.01} & 0.28 ± 0.10 & 1.34 ± 0.21 & 9.68 ± 1.58 & 5.00 \\
TST & 0.46 ± 0.02 & \underline{0.91 ± 0.05} & 0.71 ± 0.05 & 0.29 ± 0.11 & 1.47 ± 0.20 & 16.52 ± 6.55 & 6.83 \\
STF & \cellcolor[gray]{0.85}\textbf{0.41 ± 0.02} & 0.99 ± 0.29 & 1.92 ± 1.15 & 0.26 ± 0.16 & 1.92 ± 0.34 & 13.79 ± 4.66 & 6.17 \\
CF & 0.57 ± 0.06 & 1.83 ± 0.18 & \underline{0.24 ± 0.02} & 0.26 ± 0.10 & 0.82 ± 0.34 & \underline{7.50 ± 1.02} & 5.00 \\
PTST & 3.71 ± 0.02 & 6.34 ± 0.02 & 0.45 ± 0.02 & 3.58 ± 0.46 & 2.06 ± 0.41 & 9.80 ± 1.67 & 10.33 \\
XGBoost & 1.02 ± 0.00 & 3.55 ± 0.00 & 0.33 ± 0.00 & 0.13 ± 0.00 & \cellcolor[gray]{0.85}\textbf{0.61 ± 0.00} & 9.04 ± 0.00 & 6.17 \\
TabPFN & 0.44 ± 0.01 & 1.14 ± 0.01 & 0.27 ± 0.00 & \cellcolor[gray]{0.85}\textbf{0.06 ± 0.00} & 0.72 ± 0.03 & 7.55 ± 0.22 & 3.50 \\
TabDPT & 0.50 ± 0.01 & \cellcolor[gray]{0.85}\textbf{0.90 ± 0.01} & 0.25 ± 0.01 & \underline{0.10 ± 0.00} & \underline{0.69 ± 0.01} & 8.63 ± 0.53 & 3.33 \\
\midrule
\multicolumn{8}{l}{\textit{Original engineering units ($\downarrow$)}} \\
\midrule
Linear & 136.91 ± 4.13 & 394.76 ± 38.18 & 377375.42 ± 189407.02 & 2662.38 ± 123.32 & 43203.12 ± 43216.43 & 1877895.28 ± 2475205.90 & 12.50 \\
Exp & 47.10 ± 1.17 & 191.83 ± 2.71 & 335612.34 ± 716896.98 & 1482.75 ± 160.37 & 6278.27 ± 188.56 & 228864.76 ± 83234.84 & 9.50 \\
MLP & 75.94 ± 5.37 & 276.90 ± 34.56 & 59680.85 ± 81102.09 & 444.59 ± 194.00 & 7060.46 ± 1246.54 & 287760.52 ± 74993.08 & 10.50 \\
LSTM & \underline{43.44 ± 2.21} & 104.08 ± 8.97 & 3208.61 ± 369.69 & 293.94 ± 136.27 & 3344.16 ± 173.37 & \cellcolor[gray]{0.85}\textbf{123688.33 ± 2403.85} & 3.83 \\
CNN-1D & 48.09 ± 4.82 & 101.08 ± 5.83 & 15328.28 ± 7016.36 & 749.70 ± 938.67 & 6162.49 ± 1160.94 & 254182.17 ± 130965.65 & 8.17 \\
TiDE & 46.60 ± 3.56 & 109.02 ± 4.13 & \cellcolor[gray]{0.85}\textbf{2633.01 ± 107.72} & 347.09 ± 128.23 & 3238.60 ± 518.75 & 194915.73 ± 28818.57 & 5.17 \\
TST & 46.15 ± 2.03 & \underline{91.22 ± 4.96} & 8293.91 ± 557.49 & 358.65 ± 139.75 & 3565.53 ± 491.39 & 331603.13 ± 135837.29 & 6.83 \\
STF & \cellcolor[gray]{0.85}\textbf{41.31 ± 2.25} & 99.50 ± 29.06 & 22492.01 ± 13489.49 & 329.53 ± 200.78 & 4641.65 ± 816.21 & 275295.87 ± 99592.97 & 6.33 \\
CF & 56.62 ± 6.20 & 182.98 ± 17.52 & \underline{2873.02 ± 197.96} & 320.71 ± 120.56 & 1986.17 ± 825.15 & \underline{149462.60 ± 25698.73} & 5.00 \\
PTST & 370.79 ± 2.11 & 634.40 ± 2.30 & 5338.39 ± 225.45 & 4483.61 ± 577.87 & 4999.49 ± 987.11 & 182128.52 ± 38677.98 & 10.17 \\
XGBoost & 102.50 ± 0.00 & 355.02 ± 0.00 & 3823.81 ± 0.00 & 161.62 ± 0.00 & \cellcolor[gray]{0.85}\textbf{1484.77 ± 0.00} & 176032.70 ± 0.00 & 6.17 \\
TabPFN & 44.17 ± 0.59 & 114.50 ± 1.47 & 3121.41 ± 41.35 & \cellcolor[gray]{0.85}\textbf{79.84 ± 1.56} & 1744.71 ± 73.12 & 158581.64 ± 4950.91 & 3.50 \\
TabDPT & 50.30 ± 0.90 & \cellcolor[gray]{0.85}\textbf{90.43 ± 0.62} & 2875.42 ± 67.81 & \underline{121.85 ± 2.64} & \underline{1675.44 ± 32.22} & 163786.30 ± 10020.04 & 3.33 \\
\bottomrule
\end{tabular}
\end{adjustbox}
\end{table}

\paragraph{Cross-family observations.}
Prognostics rankings are more compressed than diagnostics. TabDPT, TabPFN, and LSTM lead on combined rank, but CF, TiDE, and STF remain within a few rank points and each takes at least one column-best on normalized MAE. Two patterns stand out. First, the transformer family that collapses on diagnostics is competitive on prognostics (e.g., STF is best on NC-DS02 and TST is near the top on NC-P), suggesting a task-specific failure mode rather than an architecture-wide limitation. Second, simple baselines (Linear, Exp) degrade more on prognostics than on diagnostics: Exp is consistently far from the leaders, while Linear remains mid-tier. Switching from MAE to MSE tightens the leading group and penalizes rare catastrophic errors, but it does not change the top of the leaderboard.

\subsubsection{Per-unit aggregation (battery and bearing families)}
\label{app:prog_per_unit}

Battery (NB14, UNIBO21) and bearing (XJTU-SY) prognostics are evaluated trajectory-level rather than window-level: the framework's \texttt{per\_unit} evaluator computes one error per monitored unit and then aggregates. Table~\ref{tab:results_app_mae_mean} reports the per-unit-mean MAE on these three families in the same two-floor form (normalized top, denormalized bottom). The values are produced from the same predictions as the aggregate tables above; the difference is purely in the aggregation order (per-unit-then-mean vs.\ pooled). The corresponding MSE per-unit-mean variants are omitted from the appendix as they preserve the same ranking with no additional insight.

\begin{table}[H]
\centering
\footnotesize
\caption{Per-unit-mean MAE on battery and bearing prognostics. Top block: normalized target space ($\times 100$) ($\downarrow$); bottom block: original engineering units ($\downarrow$). Computed by the \texttt{per\_unit} evaluator: one error per monitored unit, then averaged across units.}
\label{tab:results_app_mae_mean}
\begin{adjustbox}{max width=\textwidth}
\begin{tabular}{lrrrr}
\toprule
Model & \multicolumn{1}{c}{NB14} & \multicolumn{1}{c}{Unibo} & \multicolumn{1}{c}{XJTU-SY} & \multicolumn{1}{c}{Average rank} \\
\midrule
\multicolumn{5}{l}{\textit{Normalized target space ($\downarrow$)}} \\
\midrule
Linear & 38.58 ± 9.96 & 31.66 ± 21.46 & 75.37 ± 47.26 & 13.00 \\
Exp & 30.74 ± 48.55 & 10.93 ± 0.18 & 21.38 ± 1.44 & 10.00 \\
MLP & 14.83 ± 9.66 & 13.88 ± 1.78 & 25.21 ± 2.73 & 11.33 \\
LSTM & 4.49 ± 0.32 & 7.02 ± 0.71 & \underline{18.57 ± 0.50} & 4.33 \\
CNN-1D & 9.12 ± 1.43 & 11.74 ± 1.15 & 26.13 ± 6.63 & 10.67 \\
TiDE & \cellcolor[gray]{0.85}\textbf{4.02 ± 0.12} & 6.30 ± 0.64 & 21.28 ± 2.55 & 4.33 \\
TST & 7.20 ± 0.26 & 7.39 ± 1.18 & 25.08 ± 6.06 & 8.33 \\
STF & 10.16 ± 3.03 & 8.10 ± 0.72 & 21.44 ± 2.55 & 9.00 \\
CF & 4.14 ± 0.22 & 5.43 ± 1.01 & 18.87 ± 0.43 & 3.67 \\
PTST & 5.57 ± 0.15 & 10.56 ± 0.97 & 21.13 ± 1.16 & 7.33 \\
XGBoost & 4.82 ± 0.00 & 3.59 ± 0.00 & 19.20 ± 0.00 & 4.33 \\
TabPFN & \underline{4.07 ± 0.02} & \cellcolor[gray]{0.85}\textbf{3.25 ± 0.07} & \cellcolor[gray]{0.85}\textbf{17.83 ± 0.13} & 1.33 \\
TabDPT & 4.09 ± 0.08 & \underline{3.57 ± 0.15} & 20.09 ± 0.30 & 3.33 \\
\midrule
\multicolumn{5}{l}{\textit{Original engineering units ($\downarrow$)}} \\
\midrule
Linear & 417.94 ± 107.84 & 155.85 ± 105.63 & 434.14 ± 341.49 & 13.00 \\
Exp & 333.02 ± 525.87 & 53.79 ± 0.91 & 153.84 ± 22.97 & 10.00 \\
MLP & 160.68 ± 104.69 & 68.31 ± 8.74 & 173.19 ± 15.11 & 11.00 \\
LSTM & 48.67 ± 3.42 & 34.55 ± 3.51 & \cellcolor[gray]{0.85}\textbf{123.73 ± 2.27} & 4.00 \\
CNN-1D & 98.76 ± 15.52 & 57.78 ± 5.68 & 175.36 ± 46.62 & 10.33 \\
TiDE & \cellcolor[gray]{0.85}\textbf{43.52 ± 1.29} & 31.02 ± 3.17 & 141.91 ± 13.43 & 4.00 \\
TST & 77.96 ± 2.77 & 36.37 ± 5.79 & 188.24 ± 43.65 & 9.00 \\
STF & 110.11 ± 32.79 & 39.87 ± 3.52 & 161.01 ± 22.65 & 9.00 \\
CF & 44.84 ± 2.43 & 26.75 ± 4.98 & \underline{124.85 ± 6.01} & 3.33 \\
PTST & 60.29 ± 1.65 & 51.99 ± 4.78 & 143.67 ± 8.34 & 7.67 \\
XGBoost & 52.24 ± 0.00 & 17.68 ± 0.00 & 138.98 ± 0.00 & 4.67 \\
TabPFN & \underline{44.10 ± 0.25} & \cellcolor[gray]{0.85}\textbf{16.00 ± 0.32} & 125.89 ± 1.97 & 2.00 \\
TabDPT & 44.26 ± 0.83 & \underline{17.58 ± 0.72} & 131.36 ± 2.54 & 3.00 \\
\bottomrule
\end{tabular}
\end{adjustbox}
\end{table}

Per-unit aggregation tightens the gap among the top three (TabDPT, TabPFN, LSTM) on NB14 and UNIBO21, where pooled-MAE differences were inflated by long-trajectory units, and reorders the middle of the table on XJTU-SY. The headline that tabular foundation models lead on average rank is unchanged.

\subsubsection{Domain-specific prognostic scores}
\label{app:prog_domain_scores}

Two community-standard scores are reported on the families they apply to. The \emph{NASA score} is defined for direct-RUL targets and is reported on the N-CMAPSS families (NC-DS02, NC-P) and PHME20; it asymmetrically penalizes late predictions. The \emph{PHM score} is the bearing/battery-prognostics convention and is reported on NB14, UNIBO21, and XJTU-SY. Per-task scoping is enforced by the framework's metric registry (Section~\ref{sec:evaluation_protocol}); the two scores are presented together in Table~\ref{tab:results_app_domain_scores} as two blocks of one table because they apply to disjoint family sets.

\begin{table}[H]
\caption{Domain-specific prognostic scores. Top block: NASA score on direct-RUL families (NC-DS02, NC-P, PHME20; $\downarrow$); bottom block: PHM score ($\times 100$) on battery and bearing prognostics (NB14, UNIBO21, XJTU-SY; $\uparrow$). Per-task scoping is enforced by the framework's metric registry. Note that the two scores apply to disjoint family sets and use opposite directions.}
\label{tab:results_app_domain_scores}
\begin{adjustbox}{max width=\textwidth}
\begin{tabular}{lrrrr}
\toprule
Model & \multicolumn{1}{c}{NC-DS02} & \multicolumn{1}{c}{NC-P} & \multicolumn{1}{c}{PHME20} & \multicolumn{1}{c}{Average rank} \\
\midrule
\multicolumn{5}{l}{\textit{NASA score on direct-RUL families ($\downarrow$)}} \\
\midrule
Linear & 2.03 ± 0.07 & 7.90 ± 2.31 & 3044.09 ± 2526.30 & 11.00 \\
Exp & 0.85 ± 0.02 & 2.69 ± 0.04 & 229.49 ± 39.40 & 6.67 \\
MLP & 2.53 ± 1.62 & 4910.51 ± 10876.24 & 481046.65 ± 1027047.25 & 12.33 \\
LSTM & 0.81 ± 0.04 & 1.46 ± 0.13 & 8.50 ± 5.84 & 4.00 \\
CNN-1D & 0.87 ± 0.09 & 1.44 ± 0.09 & 2709.03 ± 5992.51 & 7.00 \\
TiDE & 0.86 ± 0.06 & 1.69 ± 0.06 & 11.82 ± 9.43 & 6.00 \\
TST & 0.85 ± 0.03 & \cellcolor[gray]{0.85}\textbf{1.26 ± 0.09} & 12.64 ± 7.24 & 4.00 \\
STF & \cellcolor[gray]{0.85}\textbf{0.78 ± 0.03} & \underline{1.30 ± 0.33} & 21.05 ± 30.49 & 3.33 \\
CF & 1.01 ± 0.11 & 2.99 ± 0.18 & 2215.75 ± 4653.64 & 9.00 \\
PTST & 6.11 ± 0.12 & 10.69 ± 0.04 & 19528004.00 ± 11187183.05 & 12.67 \\
XGBoost & 1.62 ± 0.00 & 6.10 ± 0.00 & \underline{3.95 ± 0.00} & 7.33 \\
TabPFN & \underline{0.80 ± 0.01} & 1.49 ± 0.03 & \cellcolor[gray]{0.85}\textbf{1.21 ± 0.02} & 3.00 \\
TabDPT & 0.91 ± 0.02 & 1.31 ± 0.01 & 5.94 ± 2.04 & 4.67 \\
\midrule
Model & \multicolumn{1}{c}{NB14} & \multicolumn{1}{c}{Unibo} & \multicolumn{1}{c}{XJTU-SY} & \multicolumn{1}{c}{Average rank} \\
\midrule
\multicolumn{5}{l}{\textit{PHM score on battery and bearing prognostics ($\uparrow$)}} \\
\midrule
Linear & 5.38 ± 1.51 & 5.85 ± 2.55 & 11.40 ± 10.24 & 13.00 \\
Exp & 14.45 ± 8.13 & 9.91 ± 0.30 & 19.15 ± 4.44 & 10.67 \\
MLP & 15.46 ± 6.30 & 10.85 ± 1.11 & 18.27 ± 2.20 & 10.00 \\
LSTM & 31.28 ± 2.21 & 14.74 ± 0.46 & \cellcolor[gray]{0.85}\textbf{24.83 ± 0.78} & 3.00 \\
CNN-1D & 18.29 ± 2.08 & 10.70 ± 1.32 & 18.36 ± 4.77 & 10.00 \\
TiDE & \cellcolor[gray]{0.85}\textbf{36.28 ± 1.72} & 14.63 ± 2.39 & 22.46 ± 2.77 & 4.33 \\
TST & 24.31 ± 0.92 & 13.71 ± 1.41 & 14.89 ± 7.81 & 8.67 \\
STF & 14.34 ± 6.88 & 10.79 ± 1.27 & 21.58 ± 3.29 & 9.67 \\
CF & \underline{33.45 ± 1.31} & 17.21 ± 1.78 & \underline{24.72 ± 1.78} & 2.67 \\
PTST & 24.17 ± 0.71 & 12.57 ± 1.66 & 23.41 ± 1.61 & 7.00 \\
XGBoost & 24.49 ± 0.00 & 20.24 ± 0.00 & 20.14 ± 0.00 & 5.67 \\
TabPFN & 26.03 ± 0.28 & \cellcolor[gray]{0.85}\textbf{21.59 ± 0.41} & 24.40 ± 0.63 & 3.33 \\
TabDPT & 29.16 ± 0.40 & \underline{20.35 ± 0.73} & 24.42 ± 1.01 & 3.00 \\
\bottomrule
\end{tabular}
\end{adjustbox}
\end{table}

Domain-specific scores mostly track MAE/MSE, but they surface two important failure modes. On the NASA score, a small number of catastrophically late predictions on PHME20 causes MLP and PTST to degrade by orders of magnitude under the asymmetric penalty---a pattern that MAE/MSE can hide. On the PHM score, CF achieves the best score on XJTU-SY despite being mid-tier on MAE, reflecting the score's emphasis on end-of-life behavior. Reporting symmetric errors alongside domain scores therefore highlights behaviors that any single metric would miss.

\subsection{\reviewheading{Data-Efficiency Scaling}}
\label{app:data_efficiency_scaling}
This section supports the main-paper claim that data efficiency depends not only on the number of available training windows, but also on whether the context covers the relevant operating regimes, degradation stages, and diagnostic classes.
The additional scaling figure compares aggregate and blockwise subsampling for PHME20, Unibo, and MZVAV. It extends the main-results discussion by showing when a small context remains representative and when blockwise subsampling removes important trajectory segments or class coverage.

\begin{figure}[H]
\centering
\begin{subfigure}[t]{0.44\textwidth}
\centering
\includegraphics[width=\textwidth]{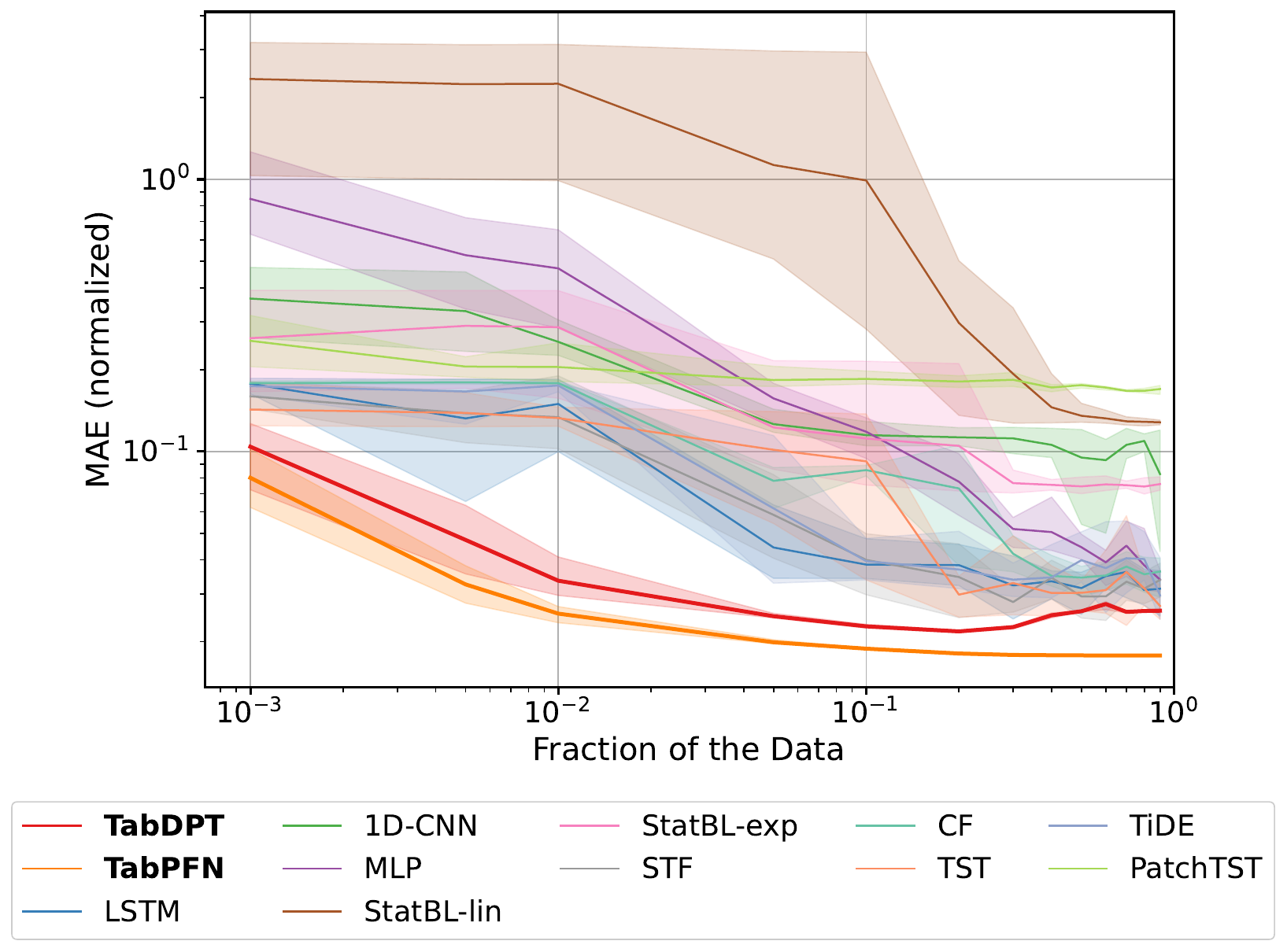}
\caption{PHME20}
\label{fig:scaling_phme20}
\end{subfigure}
\hfill
\begin{subfigure}[t]{0.44\textwidth}
\centering
\includegraphics[width=\textwidth]{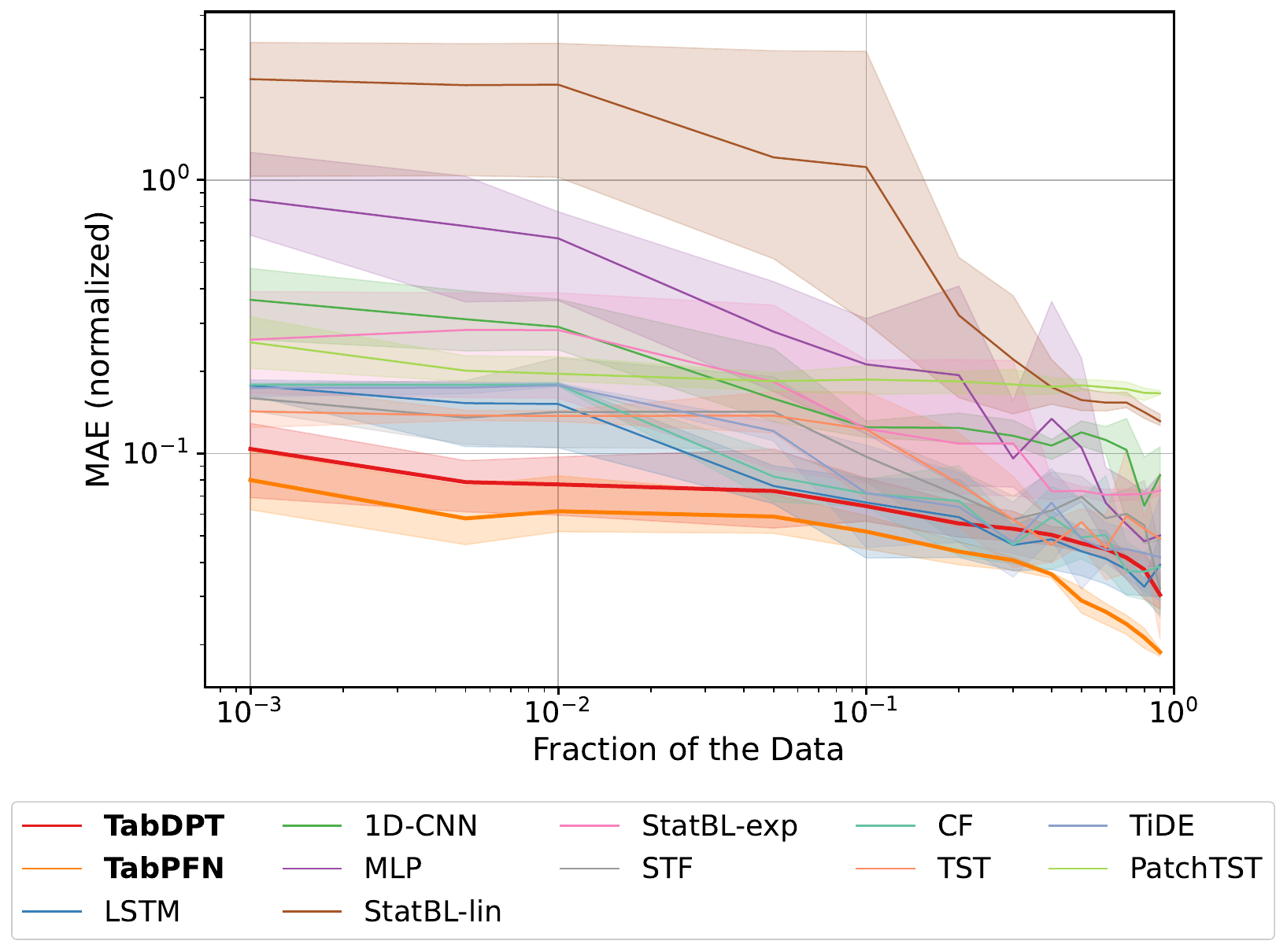}
\caption{PHME20, blockwise}
\label{fig:scaling_phme20_blockwise}
\end{subfigure}

\smallskip

\begin{subfigure}[t]{0.44\textwidth}
\centering
\includegraphics[width=\textwidth]{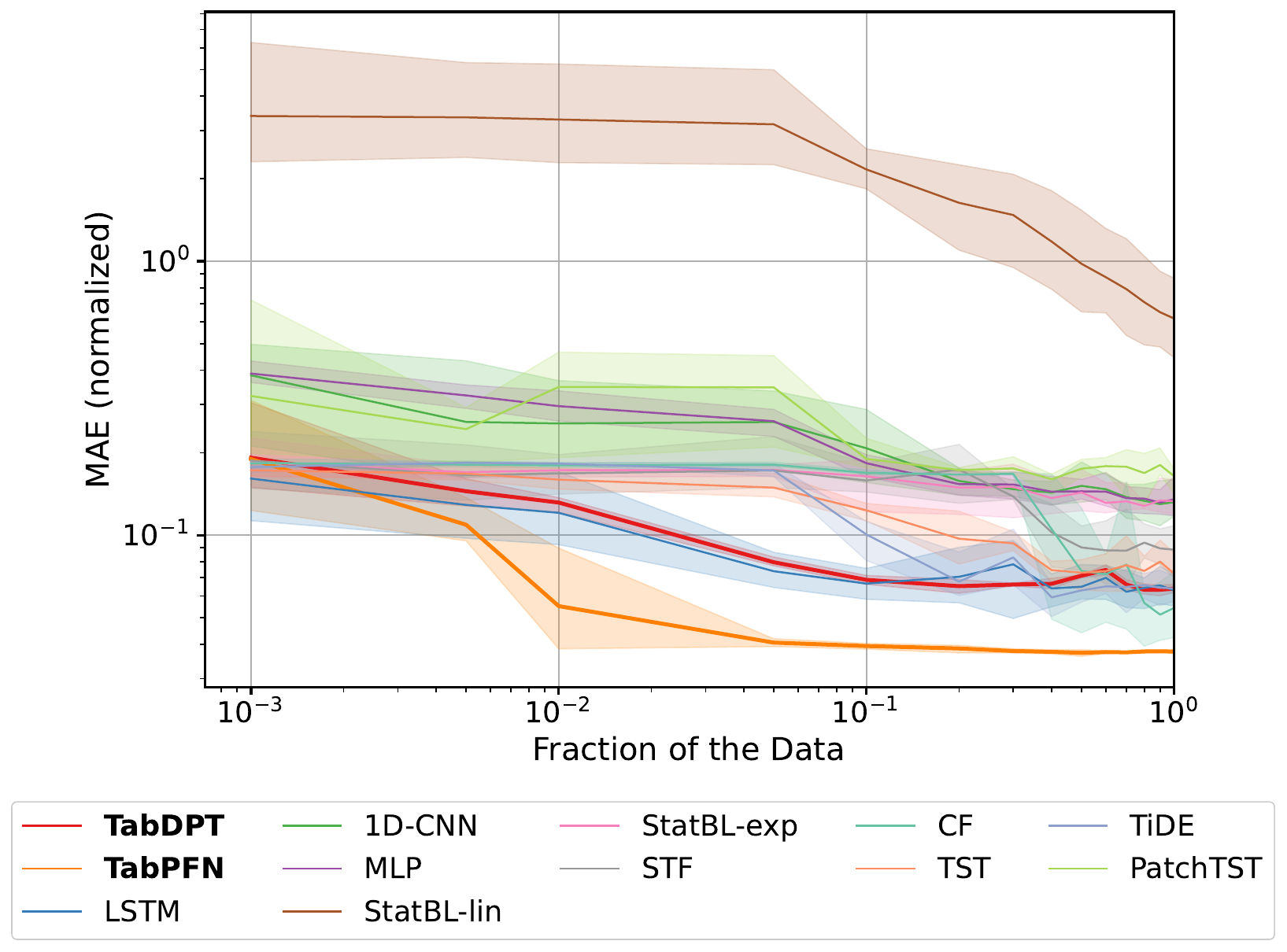}
\caption{Unibo}
\label{fig:scaling_unibo}
\end{subfigure}
\hfill
\begin{subfigure}[t]{0.44\textwidth}
\centering
\includegraphics[width=\textwidth]{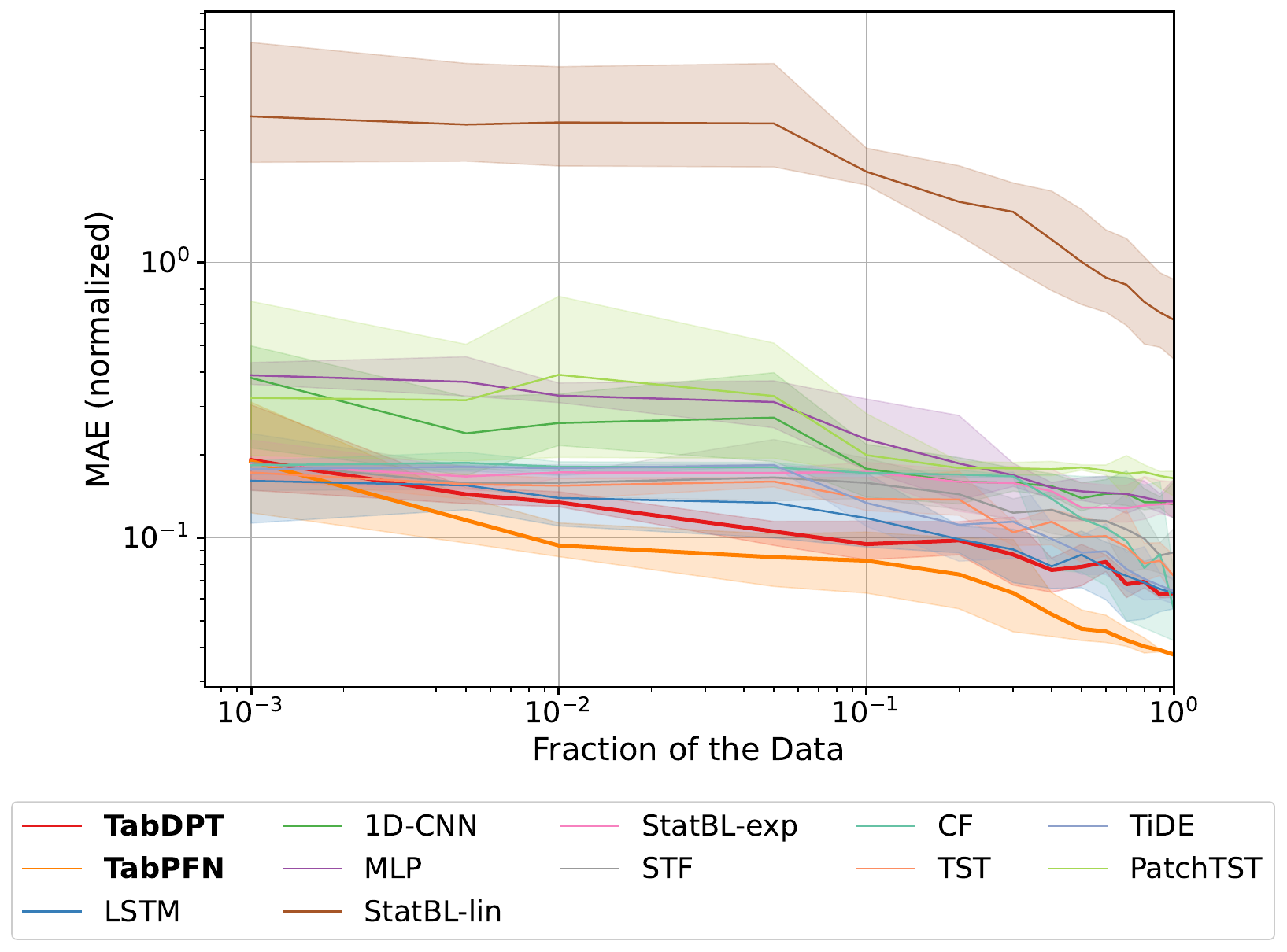}
\caption{Unibo, blockwise}
\label{fig:scaling_unibo_blockwise}
\end{subfigure}

\smallskip

\begin{subfigure}[t]{0.44\textwidth}
\centering
\includegraphics[width=\textwidth]{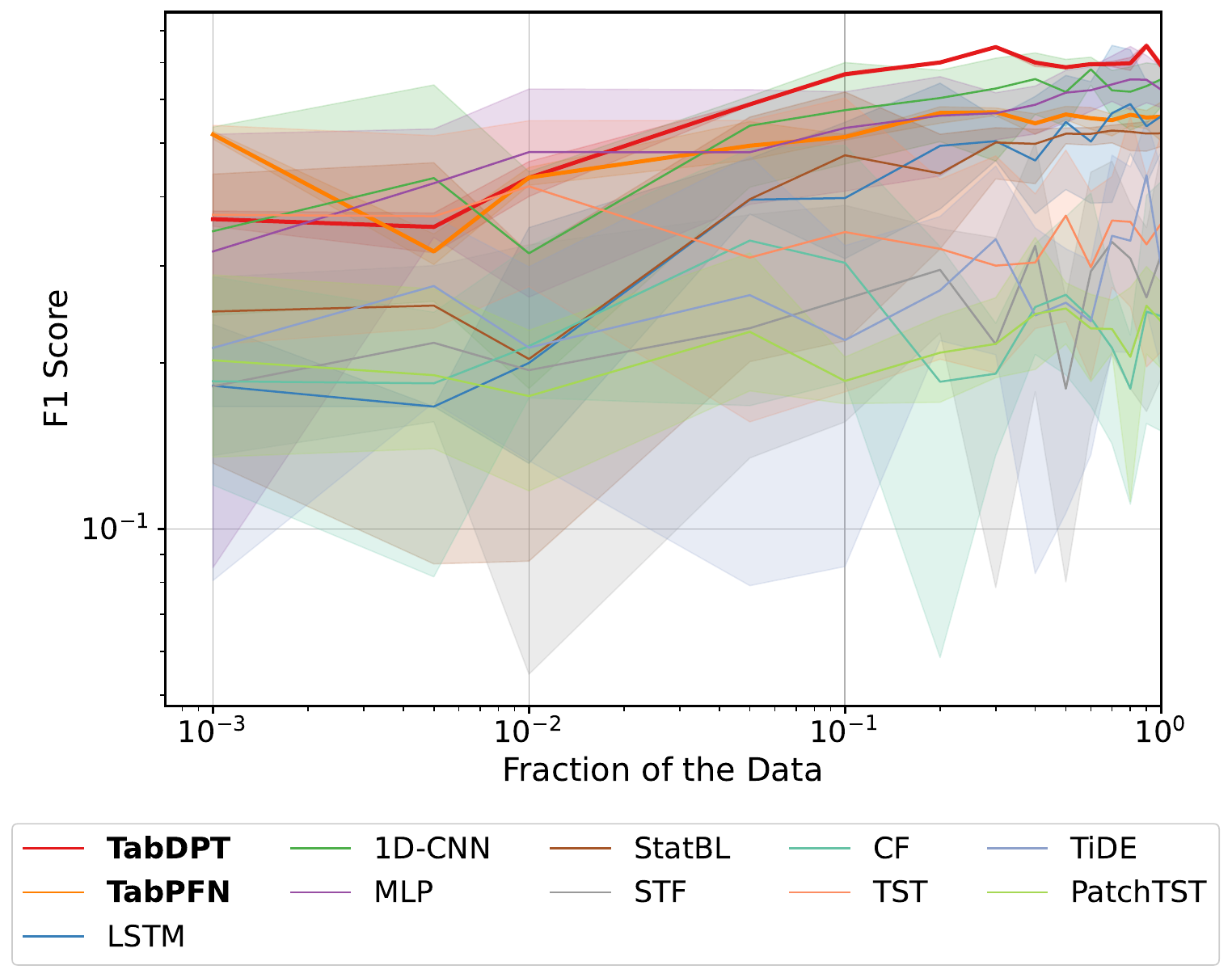}
\caption{MZVAV}
\label{fig:scaling_mzvav}
\end{subfigure}
\hfill
\begin{subfigure}[t]{0.44\textwidth}
\centering
\includegraphics[width=\textwidth]{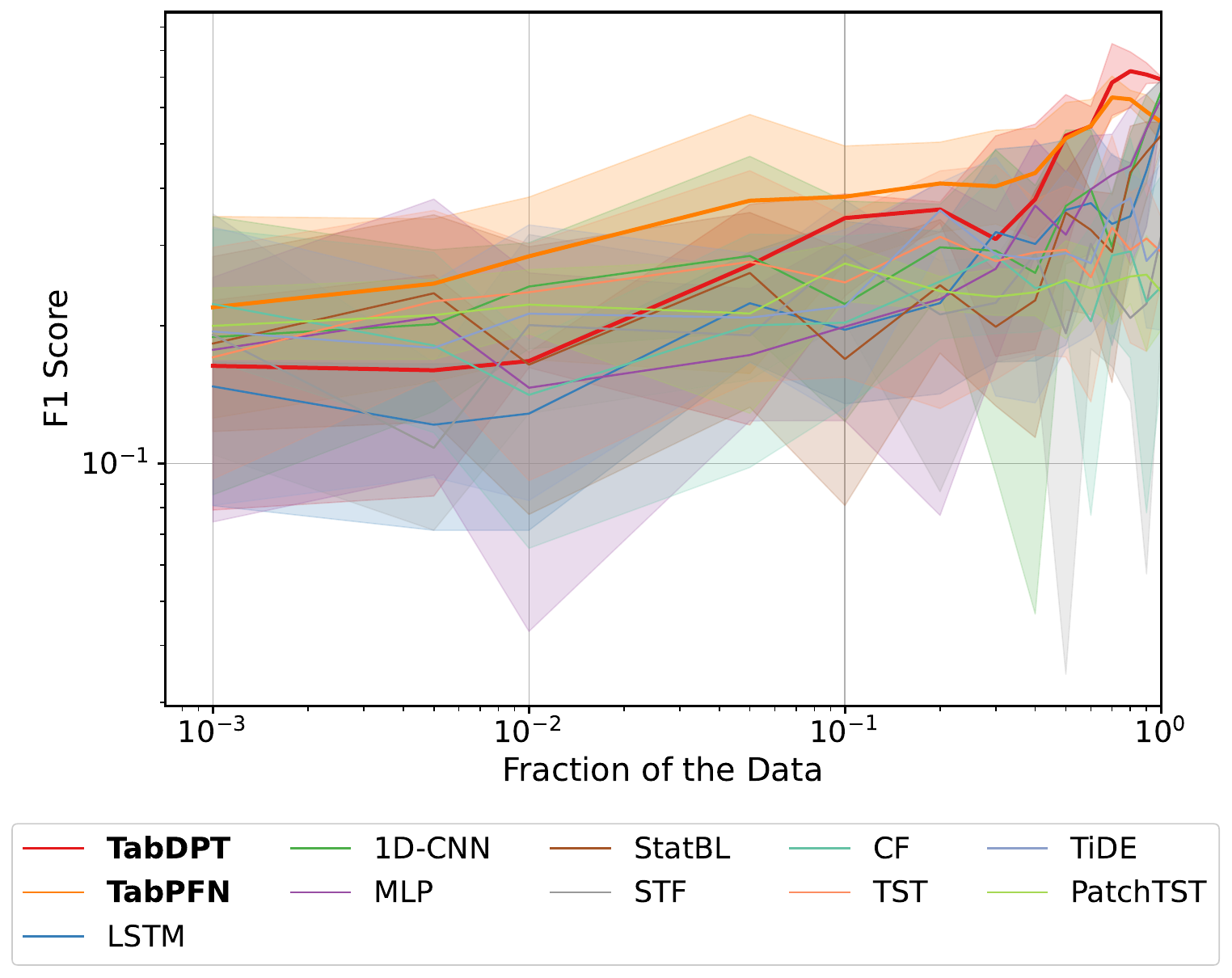}
\caption{MZVAV, blockwise}
\label{fig:scaling_mzvav_blockwise}
\end{subfigure}
\caption{Data-efficiency scaling under aggregate and blockwise context subsampling for PHME20, Unibo, and MZVAV. Each row compares aggregate random subsampling with contiguous blockwise subsampling, highlighting when small contexts preserve or lose coverage of trajectories or diagnostic classes.}
\label{fig:scaling_phme20_unibo_mzvav}
\end{figure}

\section{\reviewheading{Reproducibility}}
\label{app:reproducibility}

\subsection{\reviewheading{Framework-Based Execution}}
\label{app:reproducibility_framework}

All reported experiments are executed through the PICID evaluation framework \citep{telyatnikov2026picid}. In this paper, reproducibility therefore means that the benchmark choices described in the method and dataset appendices are fixed as executable experiment configurations rather than reconstructed manually from prose.

\subsection{\reviewheading{Configuration-Fixed Experiments}}
\label{app:reproducibility_configs}

Each experiment is wrapped in a resolved configuration that fixes the datasource, split, transform pipeline, target construction, sequence-slicing parameters, tabularization rule, model family, hyperparameters, evaluator, seed, and validation-selection rule. These configuration choices define the experiment from datasource construction to metric computation, so a reported number is tied to a concrete protocol instance.

The run record should identify:
\begin{itemize}
    \item the resolved configuration, including command-line overrides;
    \item the code revision and dependency environment;
    \item the dataset source or local path configuration used for execution;
    \item the saved evaluation summaries, predictions, and plots used in the manuscript.
\end{itemize}

\subsection{\reviewheading{Shared Preprocessing and Evaluation Boundaries}}
\label{app:reproducibility_boundaries}

The same configured preprocessing and evaluation boundaries are used across the model families compared in the paper. Dataset splits are reported in the dataset-protocol section above, model selection uses validation data, and final metrics are computed on the held-out test partition.

The leakage constraints are those defined in the formal protocol. Parameters fitted by feature transformations $\mathcal{G}$, target transformations $\mathcal{H}$, target-alignment choices inside $\widetilde{\mathcal{H}}$, normalization statistics, and tabularization choices are determined only from the training partition. Once fitted, these choices are frozen and applied unchanged to validation and test partitions.

For tabular in-context models, labeled context examples are drawn only from the training partition. For intra-unit temporal splits, context selection must also respect temporal order so that no future-derived sample is used to predict an earlier query. Machine-specific paths are kept separate from the experimental protocol through path configurations, so local storage layout does not change the scientific comparison.

\section{\reviewheading{Data and Code Access}}
\label{app:data_code_access}

The public repository link is recorded as \href{https://github.com/EPFL-IMOS/picid}{github.com/EPFL-IMOS/picid}. An archival artifact identifier can be added before submission if one is created.

\subsection{\reviewheading{Code and Protocol Availability}}
\label{app:code_access}

The implementation is distributed as a Python project with version-controlled experiment configurations. These configurations define the datasource, preprocessing pipeline, task definition, model, and evaluator used for each reported benchmark setting, while machine-specific dataset and artifact locations are supplied separately through path configurations.

The code repository and experiment configurations are linked as \href{https://github.com/EPFL-IMOS/picid}{\texttt{github.com/EPFL-IMOS/picid}}.

\subsection{\reviewheading{Third-Party Datasets}}
\label{app:dataset_access}

The evaluated datasets are third-party benchmark datasets and should be obtained according to their respective source terms rather than redistributed as part of the paper. The evaluated datasets are NB14 \citep{bole2014adaptation}, UNIBO21 \citep{10.1145/3462203.3475878}, XJTU-SY \citep{yaguo2019xjtu}, N-CMAPSS \citep{arias2021aircraft,frederick2007user}, HSF15 \citep{hsf15_helwig}, PHME20 \citep{PHME20-GTU}, and MZVAV \citep{granderson_building_2020}.

The implementation supports local dataset paths through the selected path configuration. Dataset access therefore remains separate from the benchmark protocol: users obtain the third-party data from the appropriate source, place or cache them locally, and then run the same configured preprocessing and evaluation pipeline.

\bibliographystyle{cas-model2-names} 
\bibliography{references}

\end{document}